\documentclass[mnsc,sglanonrev]{informs4}

 \makeatletter
 \def\theARTICLETOPLEFT{}
 \def\theARTICLETOPRIGHT{}
 \def\theJOURNAL{}%
 \def\theJOURNALshort{}%
 \RRHFirstLine{}
 \LRHFirstLine{}
 \RRHSecondLine{\bf\theRUNAUTHOR:\enskip {\it\theRUNTITLE}}
 \LRHSecondLine{\bf\theRUNAUTHOR:\enskip {\it\theRUNTITLE}}
 \makeatother

\usepackage{eqndefns-left} 
\Equationvalidatefalse
\RequirePackage{tgtermes}
\RequirePackage{newtxtext}
\RequirePackage{newtxmath}
\RequirePackage{bm}
\RequirePackage{endnotes}

\OneAndAHalfSpacedXII 

\usepackage{algorithm}
\usepackage[noend]{algpseudocode}
\usepackage{tikz}

\usepackage{subcaption}
\usepackage{booktabs}
\usepackage{graphicx}
\usepackage{multirow}
\usepackage{caption}

\newcommand{\Ind}[1]{\mathbb{I}\{#1\}}
\newcommand{\qedsymbol}{\hfill \ensuremath{\square}}

\usepackage[size=tiny]{todonotes}

\usepackage{natbib}
 \bibpunct[, ]{(}{)}{,}{a}{}{,}%
 \def\bibfont{\small}%
\EquationsNumberedThrough    

\TheoremsNumberedThrough     
\ECRepeatTheorems  %

\MANUSCRIPTNO{MNSC-0001-2024.00}

\begin{document}


\RUNAUTHOR{Digalakis and Pérignon and Saurin and Sentenac}

\RUNTITLE{The Challenger}


\TITLE{The Challenger: When Do New Data Sources Justify Switching Machine Learning Models?} 
\ARTICLEAUTHORS{%
\AUTHOR{Vassilis Digalakis Jr}
\AFF{
Boston University}

\AUTHOR{Christophe Pérignon, Sébastien Saurin, Flore Sentenac}
\AFF{HEC Paris}
} 

\ABSTRACT{%
\textbf{Problem definition:}
Organizations often have an incumbent predictive model in production when new data sources become available. Because historical training data lack the new features, a challenger model must be trained on a small but growing full-feature dataset. We study whether, and when, the organization should switch to the challenger. The decision is statistical and economic: the challenger's predictive performance improves as full-feature data accumulate, but repeated retraining is costly and delays benefits from deployment.
\textbf{Methodology/results:}
We develop a framework linking learning-curve dynamics to model-switching economics. 
Under a standard power-law learning curve and finite data-collection horizon $T$, the optimal time to train and evaluate the challenger scales as $T^{1/(1+\alpha)}$: learning-curve shape (through its learning speed $\alpha$) is the primary theoretical determinant of when to stop experimenting; costs determine switching profitability. 
Even without knowing the learning curve, the operational problem is tractable: we show that any algorithm stopping on the $T^{2/3}$ scale and making reliable switch/discard decisions achieves $O(T^{2/3}\sqrt{\log T})$ regret relative to a full-foresight oracle. We propose a sequential evaluation algorithm that uses local learning-curve trends to anticipate improvement, and test it in a real-world credit-scoring study. Even with this local approximation, the algorithm theoretically and empirically achieves near-oracle performance. It is also more stable than greedy sequential evaluation algorithms, where noisy early estimates trigger premature discarding, or simple one-shot evaluation algorithms, which work only when their fixed evaluation time matches the (unknown in practice) theoretical timing scale.
\textbf{Managerial implications:}
Our framework offers a step toward principled model governance when new data sources require costly collection, validation, and deployment.
We show that evaluation cadence, decision confidence, and optimism about future learning interact with the cost structure in ways that standard one-shot or greedy evaluation practices overlook: a slowly-improving challenger may not justify continued evaluation, even if it would later outperform the incumbent.
}





\maketitle

\section{Introduction} \label{sec:intro}
Organizations increasingly rely on predictive models to inform decision-making in various domains. The effectiveness of these predictive tasks hinges on two fundamental elements: the choice and optimization of the {predictive model} and the availability and quality of the {data} used to train this model. In this regard, the rise of alternative data has opened new frontiers in predictive modeling. These novel data sources become available either through technological advancements (e.g., satellite imagery, \mbox{geolocation} data, transaction data, web-traffic metrics, social media data, online reviews, sensor data, environmental and weather data, supply chain and shipping data, patent/IP filings) or regulations granting individuals the right to share their personal data with third parties (e.g., the Consumer Financial Protection Bureau's open banking rules, the EU's Payment Services Directive~2, or the EU Data Act for IoT-generated data).

The emergence of new data sources is both an opportunity and a disruption in the lifecycle of a predictive model. The opportunity arises because alternative data can offer novel signals and improve forecasting performance beyond what is achievable with the original features. However, newly available features cannot be immediately integrated into existing models. They must first be incorporated into an augmented model, which requires training, validation, and comparison with the existing model before deployment. Despite its practical importance, the question of how to effectively incorporate new features into an established forecasting process has been overlooked in the literature, and this study seeks to address this gap.
 
We formalize this problem by considering an organization that has deployed a model in production, referred to as the {incumbent model} $f_I$, which was fully trained on an initial set of features. At a given point in time, some new features become available, which may enhance prediction accuracy. In response, the organization develops a new model, called the {challenger model} $f_C$, that incorporates an expanded set of features including the new ones. However, as the large dataset used to train $f_I$ lacks these new features, it cannot be used to train $f_C$. Thus, $f_C$ must be trained on a new dataset containing a limited number of samples, which typically hinders its initial performance. Over time, as more data including the new features are collected, the performance of the challenger model $f_C$ is expected to improve and may eventually exceed that of the incumbent model $f_I$.

Transitioning from a fully trained incumbent model to a newly developed challenger involves several important trade-offs. If $f_C$ delivers a meaningful improvement over $f_I$, adopting it earlier can reduce forecasting errors and generate real operational value. These performance gains, however, must be weighed against several types of costs. Accessing new features may require payments to external data providers or incur internal expenses related to data extraction and processing. Training and validating $f_C$ also entails feature-engineering efforts, computational expenditures—including cloud infrastructure and electricity—and human labor for model development, debugging, and testing. Once $f_C$ is deemed superior, additional deployment and switching costs arise from integrating the model into production. Feature-access and training/validation costs may include both fixed components and variable components that scale with the number of samples incorporating the new features, whereas deployment and switching costs are fixed and incurred only once.

The central research question we address is: 
\emph{when does the emergence of new data sources justify switching from the incumbent model $f_I$ to the challenger model $f_C$?} 
We develop a sequential decision-making framework that guides the decision to either
\emph{(i) switch} -- deploying $f_C$ and replacing $f_I$;
\emph{(ii) discard} -- permanently rejecting $f_C$ and stopping data collection; 
or \emph{(iii) continue} -- collecting more data and deferring the decision until a later time.
Our approach combines statistical estimation to quantify the challenger’s performance gains, with the economics of training, data collection, and delayed deployment, while explicitly accounting for statistical uncertainty in the decision.

\subsection{Contributions and Outline}
We make three contributions to the literature on data-driven decision making and model lifecycle management. \textit{(i)} 
First, we formalize a model-switching problem that arises when new data sources become available after an incumbent predictive model is already in production. The challenger can use the expanded feature set, but must be trained on a small and growing full-feature dataset. Our framework links the challenger’s learning curve, the costs of collecting and validating new data, retraining and switching costs, and the timing of future gains. This yields an economically grounded decision problem with three actions: switch, discard, or continue.
\textit{(ii)}
Second, we characterize the structure of the optimal switching decision. In a finite-horizon benchmark with a power-law learning curve, the oracle training and evaluation time scales as $T^{1/(1+\alpha)}$. Thus, learning speed $\alpha$ (i.e., the learning-curve exponent) determines when experimentation should stop, whereas costs determine whether switching is profitable. We then show that full knowledge of the learning curve is not necessary for good performance: any procedure that stops on a sufficient (statistical learning-derived) time scale and makes reliable switch/discard decisions achieves $O(T^{2/3}\sqrt{\log T})$ regret relative to a full-foresight oracle.
\textit{(iii)}
Third, we translate these insights into practical evaluation algorithms/rules. We study a one-shot rule, a greedy sequential rule, and a look-ahead sequential rule that uses local learning-curve trends to anticipate future improvement. In a real-world credit-scoring application, the look-ahead rule is the most stable and approaches oracle value across early- and late-switching regimes. In contrast, one-shot rules perform well only when their fixed evaluation time happens to match the right scale, and greedy rules can discard promising challengers prematurely when early estimates are noisy. 
The study illustrates how learning curves and cost parameters interact and demonstrates the practical relevance of our framework for data-driven decision making in financial institutions.

The remainder of the paper is organized as follows. Section~\ref{sec:problem-formulation} presents the problem and model formulation. Section~\ref{sec:theory} characterizes the oracle benchmark and develops regret guarantees for implementable procedures. Section~\ref{sec:algs} introduces practical evaluation algorithms. Section~\ref{sec:numerical} reports the credit-scoring case study. Section~\ref{sec:conclusion} discusses managerial implications and concludes. All proofs and additional analyses are deferred to the Electronic Companion (EC). To keep the exposition focused on the central learning-versus-deployment trade-off, the main text analyzes a simplified setting that isolates the key primitives of the problem; we use remarks throughout the paper to preview richer variants and develop the corresponding formal extensions in the EC.

\subsection{Related Literature} \label{ssec:lit}
Our work sits at the intersection of four research streams that have evolved largely in parallel. 
The first is the literature on machine learning (ML) deployment and operationalization, which studies how predictive models are selected, deployed, and governed as components of operational decision systems.
The second is the literature on model retraining, which studies when predictive models should be updated as new data become available. 
The third is the learning-curve literature, which characterizes how model performance evolves with additional samples and how such curves can guide data acquisition and training decisions.  
The fourth is the literature on alternative data sources in finance, where newly accessible signals arrive gradually and often require careful economic evaluation before integration. 
Our framework bridges these streams by modeling performance gains, data and retraining costs, and the timing of information arrival within a unified formulation, thereby formalizing the decision of when new data justify switching to a challenger model.
Our setting also relates conceptually to sequential decision-making and optimal stopping, as the choice to switch, discard, or continue is made repeatedly under uncertainty. However, the stopping problem we study does not match standard formulations, as gains and costs depend endogenously on sample size rather than exogenous stochastic rewards.
Relatedly, \citet{kaps2025choose} develop a parallel economic framework for exploring, continuing, or switching among competing technologies; our context is predictive-model governance rather than technology investment.

\paragraph{ML deployment and operationalization.}
A growing stream in operations and management science treats ML as part of an operational decision system: recent review papers document the rise of data analytics and ML in operations management and emphasize that value is created when predictions are embedded in decisions rather than evaluated only by statistical accuracy \citep{misic2020data,davis2024best}. Methodologically, predictive-to-prescriptive analytics and decision-focused learning study how predictive models should be trained or used when downstream operational objectives matter \citep{bertsimas2020predictive,elmachtoub2022smart}, with applications including inventory and pricing \citep{ban2019big,ferreira2016analytics}. Closest to the deployment perspective of our paper, \citet{digalakis2025ml} develop a framework for selecting among AI models by jointly accounting for capability, deployment costs, and compliance constraints, thereby formalizing the gap between model performance and deployability. \citet{chan2026deployment} similarly show that selecting AI-assisted interventions based only on predictive performance can be suboptimal under capacity constraints and noisy compliance. Other work studies the organizational and behavioral conditions under which deployed analytics tools are actually used, including managerial adherence to pricing recommendations and worker adoption of algorithmic advice \citep{caro2023believing,kawaguchi2021workers}. \citet{feng2025tests} develop statistical tests for replacing human decision makers with algorithms. Our work expands this literature by focusing on a distinct model-lifecycle decision: an organization already has a predictive model in production, a challenger enabled by newly available features must be trained on a small but growing full-feature dataset, and the decision maker must jointly choose when to keep collecting features, when to discard the challenger, and when to switch models.


\paragraph{Model retraining.}
This stream studies when a deployed predictive model should be retrained as new data and modeling options become available. The literature on model retraining has developed along two main directions.
The first examines how to select among competing models using only partial training data: progressive sampling procedures \citep{john1996static,10.1145/312129.312188} enlarge the training set in stages and stop when additional data no longer increases expected utility, trading off performance gains against data and computational costs. 
The second studies retraining deployed models when the data distribution evolves, focusing on detecting distribution shifts \citep{bifet2007learning,pesaranghader2016fast} and adapting models to mitigate their effects \citep{schwinn2022improving,kabra2024limitations,bertsimas2024towards}. Recent work also develops tools for diagnosing which components of distribution shift drive model-performance degradation \citep{cai2026diagnosing}. Cost-aware retraining rules include staleness-based policies \citep{MAHADEVAN2024} and return-on-investment frameworks for deciding when updates are worthwhile \citep{ZLIOBAITE2015}.
Our work is complementary but tackles a different problem: rather than determining when to update a fixed model under partial training data or distribution drifts, we study when the arrival of new features and new data justifies switching from an incumbent model to a challenger. We integrate learning-curve dynamics, data-acquisition and retraining costs, and future economic gains into a unified optimization problem that formalizes the transition decision.


\paragraph{Learning curves.}
This stream studies how model performance evolves with sample size and how such learning curves can inform data acquisition and training decisions. \citet{viering2022shapelearningcurvesreview} document that learning curves frequently exhibit smooth, monotone improvement and are often well-approximated by power laws, while also highlighting irregular behaviors—temporary degradation, plateaus, or non-monotonicity—that complicate prediction and planning. 
For example, \cite{pmlr-vR2-frey99a} and \cite{4476671} show that learning curves of decision trees follow a power law, allowing early prediction of error rates as data grow. 
Closer to our setting, \citet{Mohr_2024} review {utility curves}, which combine performance gains with data and computational costs. Such curves peak at the optimal training size and enable the development of data-acquisition policies based on expected utility. 
\cite{10.1007/s10618-007-0082-x} and \cite{10.1145/1557019.1557076} fit explicit functional forms to partial learning curves to project future performance or optimal training sizes, whereas \cite{sabharwal2016selectingnearoptimallearnersincremental} use local linear approximations to estimate near-optimal learners from limited data. 
Our setting builds on this foundation but differs in the following way: rather than determining how much data to use to train a given model, we study how learning-curve dynamics interact with economic costs to determine {whether, and when,} an organization should transition from an incumbent to a challenger model as new data accumulate. 


\paragraph{Alternative data and ML in finance.}
Finance provides a natural empirical setting because alternative data can improve screening, forecasting, and investment decisions. 
While the challenge of integrating new data sources into forecasting models is universal, it has been studied extensively in financial applications (see \cite{Foucault2025} for a survey). 
In credit markets, alternative data improve forecasting in credit-scoring models used by lenders and buy-now-pay-later providers (while also broadening credit access for borrowers with limited credit histories), using applicants' digital footprints \citep{Berg2020,Gambacorta2024}, borrowers' pictures \citep{Ravina2025}, loan descriptions \citep{Iyer2016}, open-banking data \citep{He2023,Babina2025}, merchants' transaction volumes and payment histories \citep{Gambacorta2023}, mobile-phone data \citep{Oskarsdottir_Baesens2019}, email usage \citep{Djeundje2021}, and psychometric and cognitive assessments \citep{Bryan2024}. 
Beyond lending, alternative data have been shown to predict financial outcomes in asset management and venture capital \citep{Green2019,Bonelli2025,Chi2025,Dessaint2024,Cao2024}. 
In these studies, the value of alternative data is measured by improvements in forecasting accuracy under the simplifying assumption that alternative features are fully available for all observations. 
More broadly, despite the growing use of alternative data and increasingly complex ML models in finance \citep{GuKellyXiu2020,kelly2022financial,ChenMS2024,Kelly2024}, this literature largely abstracts from the data-acquisition, training, validation, and deployment costs that determine when switching to a new model is economically justified.
In contrast, we adopt a more operational perspective: alternative data accumulate gradually, their gains are uncertain, and their collection, validation, and deployment are costly
\section{Problem and Model Formulation} \label{sec:problem-formulation}
This section formalizes the challenger-switching decision problem.
We introduce the data-arrival process, the economic performance gap between the challenger and the incumbent, the cost structure, and the value functions that govern the decision. 
To isolate the central trade-off around the switching decision (between learning and deployment), we present a stylized setting featuring a finite, non-discounted horizon, constant sample flow, and a simple cost structure. We theoretically analyze this setting in Sections~\ref{sec:theory} and \ref{sec:algs}, and provide extensions to richer settings in the EC. The algorithms and empirical analyses in Sections~\ref{sec:algs} and \ref{sec:numerical} consider a practical setting by allowing arbitrary data-arrival processes, sparse evaluation schedules, and empirical estimation of the economic performance gap.

\subsection{Defining the problem setting} \label{ssec:decision-problem}
We consider a supervised learning setting over a feature space $\mathcal{X}$ and target space $\mathcal{Y}$. The available features are partitioned into two sets: an {initial feature set} $\mathcal I$ and an {expanded feature set} $\mathcal C \supset \mathcal I$, which becomes available only later. Initially, only the features in $\mathcal{I}$ were observable; we write $\mathcal{X}_{\mathcal{I}}$ for the projection of $\mathcal{X}$ onto the features in $\mathcal{I}$. 
An {incumbent} predictive model $f_I : \mathcal{X}_{\mathcal{I}} \to \mathcal{Y}$ was trained on a large historical dataset of size $N_0$, where each sample contains only the features in $\mathcal{I}$. 
This model belongs to a hypothesis class $\mathcal{F}_I$ and is assumed to have been trained to (or near) its optimal expected performance under the distribution of $(X_{\mathcal{I}},Y)$.

At a later point in time, the features in $\mathcal{C}\setminus\mathcal{I}$ become observable. 
Their availability enables the training of a new {challenger} model $f_C : \mathcal{X}_{\mathcal{C}} \to \mathcal{Y},$ drawn from a (possibly richer) hypothesis class $\mathcal{F}_C \supseteq \mathcal{F}_I$. 
Because the historical dataset does not contain the new features, the challenger can initially be trained only on a much smaller dataset of size $n \ll N_0$ for which all features in $\mathcal{C}$ are observed. 
As the number of full-feature samples increases over time, the challenger’s expected performance---which is uncertain at early stages---may eventually surpass that of the incumbent.

We study a sequential process with finite horizon $T<\infty$ in which data arrive over discrete time steps $t \in \mathcal{T} = \{1,\dots,T \} := [T]$, each representing a natural unit of the process (e.g., a new sample or batch). These time steps are inherent to the problem and determine when new data become available. At each time step $t$, $n$ new samples containing all features in $\mathcal{C}$ are observed, so that the cumulative number of full-feature samples after $t$ is $N_t = t \cdot n$, corresponding to a fixed batch size and constant sample flow. 
The decision maker does not necessarily act at every time step. Instead, decisions are made at a subset of these time steps, denoted by decision epochs $\mathcal{T}_D = \{t_1, \dots, t_K\} \subseteq \mathcal{T}$. At each decision epoch $t \in \mathcal{T}_D$, the decision maker:
\begin{itemize}
    \item Retrains the challenger on all data collected so far ($N_t$ samples), producing an updated $f_C^{(t)}$.
    \item Evaluates whether to: {(i) switch} -- deploying $f_C^{(t)}$ and replacing $f_I$; {(ii) discard} -- permanently rejecting $f_C$ and stopping data collection; or {(iii) continue} -- collecting more full-feature samples and deferring the decision until a later epoch.
\end{itemize}
Between decision epochs, data continue to arrive, but the challenger is not retrained or re-evaluated. This distinction between time steps and decision epochs captures the practical reality that organizations may acquire data continuously but reassess deployment decisions only at scheduled intervals.
We study the impact of epoch design on different algorithms in Sections~\ref{sec:algs} and \ref{ec:sec4-epoch-design}.

\subsection{Measuring the economic performance gap} \label{ssec:perf-gains}
To quantify the benefit of transitioning models, we define how a model's monetary gains are measured. Given any predictive model $f$, let $\ell:\mathcal{Y}\times\mathcal{Y}\to\mathbb{R}$ be a bounded gain function representing the monetary return $\ell(f(x),y)$ obtained when $f$ predicts $y$ from $x$. For any \(t \ge t_1\), let
\(
k(t):=\max\{k:t_k\le t,\ t_k\in\mathcal T_D\}
\)
denote the most recent decision epoch. Then, the expected {economic performance gap}  of model $f_C$ over model $f_I$ at time step $t$ is defined as:
\begin{equation} \label{eq:gains}
G(t):=
\mathbb E\!\left[\ell\!\left(f_C^{(t_{k(t)})}(X_C),Y\right)\right]
-
\mathbb E\!\left[\ell\!\left(f_I(X_I),Y\right)\right],
\end{equation}
where expectations are taken with respect to the distribution of $(X,Y)$ and $X_\mathcal{C}$ (resp. $X_\mathcal{I}$) is the projection of random variable $X$ onto the features in $\mathcal{C}$ (resp. $\mathcal{I}$). 
The first expectation in \eqref{eq:gains} evaluates the challenger. Since the challenger is retrained {only} at decision epochs $t_k\in\mathcal{T}_D$, the model used at a general time step $t \in \mathcal{T}$ is the one from the most recent epoch $k(t)$, trained using all $N_{t_{k(t)}}$ full-feature samples available at time~$t_{k(t)}$. 
When \(\mathcal T_D=\mathcal T\), this reduces to the gap of the challenger trained exactly at time \(t\).
The second expectation in \eqref{eq:gains} is the fixed expected performance of the incumbent $f_I$, which is trained once on historical data and does not change over time.  
Thus $G(t)$ captures the expected incremental value of switching at time~$t$, based on the last trained challenger.  
\begin{remark}
Our formulation accommodates many practical performance measures through the choice of $\ell$.
Suppose correct decisions yield monetary gains $g_1$ (true positives) and $g_2$ (true negatives), while errors incur costs $c_1$ (false positives) and $c_2$ (false negatives).  A function capturing this structure is
\(
\ell(f(x),y)
=
g_1\,\Ind{f(x)=1,y=1}
+
g_2\,\Ind{f(x)=0,y=0}
-
c_1\,\Ind{f(x)=1,y=0}
-
c_2\,\Ind{f(x)=0,y=1},
\)
where $G(t)$ measures the expected economic performance gap from improved classification under these weights. In this case, 
\(
G(t) = g_1\,\Delta\!P(\mathrm{TP})
     + g_2\,\Delta\!P(\mathrm{TN})
     - c_1\,\Delta\!P(\mathrm{FP})
     - c_2\,\Delta\!P(\mathrm{FN}),
\)
where $\Delta\!P(\cdot)$ is the challenger--incumbent difference in the corresponding error component. 
The definition generalizes to performance measures that cannot be expressed as the expectation of loss functions, such as the area-under-the-curve (AUC): $G(t) = g\,\big(\mathrm{AUC}_C(t) - \mathrm{AUC}_I\big),$ where we assume that an increase in AUC produces an expected monetary improvement of $g$ per unit.
\end{remark}
\begin{remark}
The gap process $G(t)$ is central in the switching problem. If $G(t)$ were known at all times, the decision maker could compute an oracle policy that exactly anticipates the future value of continued learning and chooses the best switching time. Section~\ref{ssec:theory-optimal} studies this oracle problem by imposing a standard learning-curve structure on $G(t)$, motivated by the extensive literature on how predictive performance improves with sample size. 
In practice, however, $G(t)$ is unknown. The decision maker observes only finite data and must work with an empirical estimate $\widehat{G}(t)$, which introduces statistical uncertainty. One of our main findings is that exact knowledge of the full gap process is not necessary: algorithms that collect enough data and make sufficiently reliable terminal switch/discard decisions can still achieve near-oracle value (see Section~\ref{ssec:theory-regretbound}).
\end{remark}

\subsection{Measuring costs} \label{ssec:costs}

We distinguish costs into pre-decision, decision-time, and post-decision: some costs are incurred before the switching decision (while evaluating the challenger), some are incurred only if the challenger is adopted, and some continue after adoption.
For pre- and post-decision, we use a constant per-sample cost structure: the same per-sample cost $c\ge0$ is paid to evaluate the challenger before deployment and to operate it after deployment.
Here, $c$ captures the cost of acquiring, processing, and maintaining the expanded feature set while the challenger is being evaluated or used in production (if it is adopted). 
Since \(n\) new full-feature samples arrive in each period, the cost at time \(t\) is $C_{\rm pre}(t)=C_{\rm post}(t)=c\cdot n$ (constant in $t$).
For decision-time, we use a one-time switching cost \(c_s\ge 0\). This captures the fixed cost of moving the challenger into production, including integration with existing systems, deployment, and any operational disruption from replacing the incumbent.
This simplification isolates the central learning-versus-deployment trade-off: waiting improves the challenger, but reduces the remaining time over which its gains can be collected.
As we show, under this cost structure, costs determine whether switching is profitable, while the optimal switching time is driven by the evolution of the economic performance gap \(G(t)\).

\begin{remark} \label{rem:costs}
Examples of richer cost structures include: 
(i) $C_{\rm pre}(t)=c_{\rm pre}n,\ C_{\rm post}(t)=c_{\rm post}n,$ with \(c_{\rm pre}\neq c_{\rm post}\): 
This captures settings in which evaluating the challenger is more or less expensive than operating it after deployment. For example, pre-decision costs may include additional validation and monitoring effort, while post-decision costs may include ongoing data-licensing or infrastructure costs. 
(ii) $ C_{\rm pre}(t) = c_{\rm pre}n + \mathbb I\{t\in\mathcal T_D\}\,c_{\rm train}N_t^q, \ q\ge 0$: 
The additional term involving the cumulative number of full-feature samples \(N_t\) captures the cost of retraining or validating the challenger at a decision epoch. When \(q=0\), this represents a fixed labor or monitoring cost per retraining run. When \(q>0\), it captures computational costs that grow with the amount of available data. We study this in our theoretical results concerning epoch design in Section~\ref{ec:sec4-epoch-design}.
(iii) More broadly, the implementable algorithms of Sections~\ref{sec:algs} and \ref{sec:numerical} work with general cost structures $C_{\rm pre}(t),\ C_{\rm post}(t)$.
\end{remark}

\subsection{Defining the value functions and switching rule} \label{ssec:objective}
At each decision epoch $t \in \mathcal{T}_D$, the decision maker chooses one of three actions: {switch} to the challenger, permanently {discard} it and stop data collection, or {continue} data collection. 
Switching to the challenger entails 
(i) the cumulative pre-decision costs incurred up to $t$, $cnt$ (sunk at $t$),
(ii) a one-time switching cost $ c_s$ incurred when switching to the challenger,
and (iii) the cumulative post-decision net gain in future time steps, $\sum_{\tau=t+1}^{T}  \left[ n \cdot ( G(t) - c ) \right].$ 
By using $G(t)$ in the post-decision net gain, we are explicitly assuming that a switch at epoch $t$ deploys the challenger model trained at that same epoch.
We do not consider the option of switching at time $t$ to a challenger trained earlier at some $t' < t$: as we explain in Section~\ref{ssec:theory-oracle}, under mild assumptions, such delayed-switch choices are always suboptimal.
Put together, the value of switching to the challenger at $t$ is:
\begin{equation} \label{eq:v-switch}
V_{\mathrm{switch}}(t)
=  \underbrace{-\,cnt}_{\text{pre-decision}}
  \underbrace{-\,c_s}_{\text{decision-time}}
  \underbrace{+ \sum_{\tau=t+1}^{T} \,\big[ n \cdot (G(t) - c) \big]}_{\text{post-decision}}
  =
-cnT-c_s+n(T-t)G(t).
\end{equation}
Discarding the challenger yields no future gains or costs:
\(
V_{\mathrm{discard}}(t) \;=\; - cnt.
\)
Since the pre-decision costs are common to both branches, the incremental value of switching is
\(
\Delta V(t)
:= V_{\mathrm{switch}}(t) - V_{\mathrm{discard}}(t)
= -\,\,c_s
  + \sum_{\tau=t+1}^{T} \,\big[ n \cdot( G(t) - c) \big].
\)
We switch at decision epoch $t$ only if $\Delta V(t) > 0$.

\begin{remark} \label{rem:disc}
Discounting can be incorporated by evaluating all cash flows at time zero. With discount
factor \(\beta\in(0,1]\), the benchmark value of switching becomes
$V_{\rm switch}^{\beta}(t)
=
-\sum_{\tau=1}^{t}\beta^\tau cn
-\beta^t c_s
+
\sum_{\tau=t+1}^{T}\beta^\tau n\{G(t)-c\},
$
while
$
V_{\rm discard}^{\beta}(t)
=
-\sum_{\tau=1}^{t}\beta^\tau cn.
$
Section~\ref{ec:sec3-extended-setting} extends our theoretical analysis, and Sections~\ref{sec:algs} and \ref{sec:numerical} implement our algorithms in the general discounted setting. We find that discounting does not qualitatively alter the structure of the results. 
\end{remark}

For any algorithm \(\mathrm{ALG}\), let \(t_{\mathrm{ALG}}\) be its terminal decision
time and \(a_{\mathrm{ALG}}\in\{\mathrm{switch},\mathrm{discard}\}\) its terminal decision.
We define its realized value as
\begin{equation} \label{eq:alg-value}
V_{\mathrm{ALG}}(T)=
\begin{cases}
V_{\rm switch}(t_{\mathrm{ALG}}), & a_{\mathrm{ALG}}=\mathrm{switch},\\
V_{\rm discard}(t_{\mathrm{ALG}}), & a_{\mathrm{ALG}}=\mathrm{discard}.
\end{cases}    
\end{equation}
We set \(V_{\rm discard}(0)=0\), corresponding to rejecting the challenger before
any new data are collected.


\section{Theoretical Developments} \label{sec:theory}
We now study a tractable version of the switching problem. We first define an oracle
benchmark that knows the true performance gaps \(G(t)\). We then impose a standard
power-law learning curve and characterize the oracle stopping time in the main
finite-horizon setting (we extend our results to a more general setting in Section~\ref{ec:sec3-extended-setting}). The results demonstrate that the optimal stopping time depends directly on the shape of the learning curve.
Finally, we show that implementable algorithms need not recover the entire learning curve: under statistical regularity conditions, it is enough to stop
on the right time scale and make a reliable terminal switch/discard decision.

\subsection{What is the oracle solution?}\label{ssec:theory-oracle}
The oracle solution is an ex-ante benchmark. With perfect knowledge of \(\{G(t)\}_{t\in\mathcal T}\),
it solves 
\(
V_{\rm ORACLE}(T)
=
\max\left\{0,\ \max_{t\in\mathcal T} V_{\rm switch}(t)\right\}.
\)
If the maximum is nonpositive, the oracle discards immediately at \(t=0\). Otherwise,
it switches at any
\(
t^*\in\arg\max_{t\in\mathcal T}V_{\rm switch}(t).
\)
Thus, the oracle’s problem reduces to a single-variable optimization over $t$ together with a final profitability check. This benchmark abstracts from the informational value of intermediate retraining and
therefore provides an upper bound for implementable procedures.
Under full foresight, two structural properties typically hold: 
(i) Since $t^*$ is known, training only one challenger is optimal (training any model other than the one deployed at $t^*$ adds pre-decision cost without increasing value). 
(ii) Delaying deployment after training is suboptimal under mild conditions, e.g., when pre-deployment costs exceed post-deployment costs or when per-sample post-deployment costs are constant; delaying can be optimal in pathological cases, e.g., if post-deployment costs drop sharply at some timestep. 

\subsection{What is the optimal stopping rule under a power-law learning curve?} \label{ssec:theory-optimal}
Having defined the oracle benchmark, we now characterize its stopping rule. In this subsection, we work at the granularity of individual time steps: the decision maker can, in principle, act after each time step, so that \(\mathcal T_D=\mathcal T\). This removes the algorithmic choice of decision epochs and isolates the economic trade-off at the heart of the problem: waiting improves the challenger by increasing the training sample size, but it also reduces the remaining horizon over which the improved model can generate value. The action is irreversible: once the challenger is adopted or discarded, the process stops.
To highlight the impact of the learning dynamics, we impose a common learning-curve structure on the challenger’s expected performance gap:
\begin{assumption}[Economic performance gap structure]\label{assume:learning-curve}
The expected economic performance gap satisfies
\(
G(t)
=
g^* - g_0 N_t^{-\alpha},
\ 
g^*>0,\ g_0>0,\ \alpha>0.
\)
\end{assumption}
This specification captures the usual pattern that the challenger improves as more full-feature samples accumulate, while marginal improvements decrease over time. The parameter \(g^*\) is the long-run per-sample value of the challenger relative to the incumbent, \(g_0\) measures the initial learning deficit, and \(\alpha\) controls the speed at which the challenger approaches its long-run performance. Thus, \(\alpha\) is the parameter that governs the shape of the learning curve.

In the finite-horizon setting under consideration,
\(
V_{\rm switch}(t)
=
-cnT-c_s+n(T-t)G(t).
\)
Thus, conditional on switching, the oracle maximizes
\(
\Phi(t):=(T-t)G(t).
\)
The final switch/discard decision is then obtained by comparing the best switching value to the value of discarding immediately.
\begin{theorem}[Exact finite-horizon stopping rule]
\label{thm:finite-stopping-main}
Consider the finite-horizon setting with \(\mathcal T_D=\mathcal T=[T]\),
\(N_t=nt\), and \(C_{\rm pre}(t)=C_{\rm post}(t)=cn\).
Let Assumption~\ref{assume:learning-curve} hold. 
Then:

\begin{enumerate}
    \item \textbf{Feasibility.}
    Switching is optimal if and only if
    $\max_{t\in\mathcal T}\Phi(t)\ge cT+\frac{c_s}{n}.$

    \item \textbf{Existence and uniqueness.}
    If \(G(T)>0\), then the continuous problem $\max_{t\in(0,T)}\Phi(t)$
    admits a unique maximizer \(t^\dagger\in(0,T)\).
    Moreover, any integer maximizer belongs to
    $
    \{\lfloor t^\dagger\rfloor,\lceil t^\dagger\rceil\}\cap\mathcal T.
    $

    \item \textbf{Stopping rule.}
    The candidate is 
    \(
    t^*_{\rm int}
    =
    \min\Bigl\{
    t\in\mathcal T:
    (T-t)G(t)\ge (T-t-1)G(t+1)
    \ \text{or}\ 
    t=T
    \Bigr\}.
    \)
\end{enumerate}
\end{theorem}

Theorem~\ref{thm:finite-stopping-main} shows that the oracle stops at the first time when the gain from waiting one more period is no longer sufficient to compensate for the loss of one period of deployment. The stopping rule therefore depends not only on the current level of the challenger’s performance, $G(t)$, but also on how quickly $G(t)$ is still improving. This is the first indication that the shape of the learning curve is a central primitive of the switching problem.
The following asymptotic characterization makes this dependence explicit:
\begin{proposition}[Asymptotic stopping rule]\label{prop:finite}
Under the assumptions of Theorem~\ref{thm:finite-stopping-main}, the optimal continuous stopping time satisfies
\(
t^\dagger
\sim_{T\to\infty}
\left(
\frac{g_0 \alpha T}{g^* n^{\alpha}}
\right)^{\frac{1}{1+\alpha}}.
\)
\end{proposition}
Here, \(t^\dagger\sim_{T\to\infty}h(T)\) means that
\(
\lim_{T\to\infty}\frac{t^\dagger}{h(T)}=1.
\)
Proposition~\ref{prop:finite} highlights that the optimal stopping time grows with the horizon \(T\) and with the initial learning deficit \(g_0\). When there is more time available, or when the challenger initially has more to learn, the oracle waits longer. Conversely, it decreases with the data arrival rate \(n\) and with the long-run gain \(g^*\): faster data accumulation or a larger eventual performance advantage makes earlier switching more attractive.
Most importantly, the timing scale is governed by the learning-curve exponent \(\alpha\): the oracle stopping time grows as
\(
T^{1/(1+\alpha)}.
\)
Thus, different learning-curve shapes lead to different optimal timing regimes. When learning is slow, the oracle waits longer because additional data continue to have substantial value. When learning is fast, the challenger approaches its long-run performance quickly, and the opportunity cost of delaying deployment dominates sooner. The exponent \(\alpha\) therefore controls how the decision maker trades off statistical improvement against remaining opportunity.
The costs $c$ and $c_s$ do not affect the candidate stopping time in this finite-horizon benchmark, because they enter the switching value as terms that are constant in t. They do, however, determine whether switching is profitable at all through the feasibility condition in Theorem~\ref{thm:finite-stopping-main}. Thus, in this setting, the learning curve determines {when} the oracle would switch, while the cost parameters determine {whether} switching is ultimately worthwhile.

\begin{remark}
Section~\ref{ec:sec3-extended-setting} extends the stopping rule to an infinite-horizon discounted setting (as per Remark~\ref{rem:disc}) with unequal pre- and post-deployment costs (as per Remark~\ref{rem:costs}). In that setting, the key object is the effective per-sample hurdle
\(
c_{\rm diff}
=
c_{\rm post}-c_{\rm pre}
+
\frac{(1-\beta)c_s}{\beta n}.
\)
Theorem~\ref{thm:threshold} shows that switching can be optimal only if
\(
g^*>c_{\rm diff},
\)
and that the stopping rule becomes
\(
G(t)-\beta G(t+1)\ge (1-\beta)c_{\rm diff}.
\)
Proposition~\ref{prop:infinite} gives the asymptotic stopping time
\(
t^*_{\beta \uparrow 1}
\sim
\left(
\frac{\alpha g_0}
{n^\alpha (g^*-c_{\rm diff})(1-\beta)}
\right)^{1/(1+\alpha)}.
\)
Thus discounting replaces the finite horizon \(T\) by the effective horizon
\((1-\beta)^{-1}\), matching the known connection between discounted and finite-horizon Markov Decision Processes.
When \(c_{\rm pre}=c_{\rm post}\), \(c_{\rm diff}\to0\) as \(\beta\uparrow1\), recovering the scale in
Proposition~\ref{prop:finite}.
Most importantly, the main insight carries over: the shape of the learning curve remains a primary determinant of optimal switching timing.
\end{remark}

\subsection{Can we approach the optimal stopping rule in practice?}  \label{ssec:theory-regretbound}

The oracle’s optimal stopping rule is, in general, unattainable: computing it requires knowledge of both the challenger’s eventual performance and the shape of its learning curve. Nonetheless, the oracle provides a useful benchmark for evaluating implementable procedures. For a fixed algorithm $\textsc{alg}$, we define its expected regret with respect to the oracle as:
\(
R_\textsc{ALG}(T):= \mathbb{E}\left[V_{\textsc{ORACLE}}(T)-V_{\textsc{ALG}}(T)\right],
\)
where the expectation is taken over both algorithmic and sampling randomness.
We now show that, although computing the oracle is unrealistic, approaching its value is not. The key point is that the algorithm does not need to recover the full learning curve (i.e., we do not impose Assumption~\ref{assume:learning-curve}), nor does it need to identify the exact oracle stopping time. It is enough to stop on the correct time scale and to make a reliable switch/discard decision whenever the value gap is sufficiently separated from zero. 
The theorem below formalizes this observation as a meta-result: any algorithm satisfying two structural conditions achieves sublinear regret.
\begin{theorem}[Meta-regret bound]\label{thm:regretbound}
    Assume that 
    the hypothesis class $\mathcal{F}_C$ of the challenger has finite VC-dimension $d$, 
    training is performed through ERM, 
    and the gain function $\ell$ takes values in $\{-1,1\}$. 
    Consider any algorithm $\textsc{alg}$, and denote by $t_{\textsc{alg}}$ 
    its terminal decision epoch (hence time step). 
    Suppose that there exist constants 
    $w_1, w_2, w_3, w_4 > 0$ such that, for all sufficiently large $T$, the 
    following conditions hold with probability at least $1 - w_1 / T$:
    
    \begin{itemize}
        \item[(C1)] The decision epoch $t_{\textsc{alg}}$ satisfies
        \(
        w_2\, T^{2/3} \leq t_{\textsc{alg}}  \leq w_3\, T^{2/3}\sqrt{\log T}.
        \)
        
        \item[(C2)] $\textsc{alg}$ uses a consistent switching rule:
        if 
        \(
            \Delta V(t_{\textsc{alg}}) + w_4\, T^{2/3} \sqrt{\log T} < 0,
        \)
        then $\textsc{alg}$ discards, and if 
        \(
            \Delta V(t_{\textsc{alg}}) - w_4\, T^{2/3} \sqrt{\log T} > 0,
        \)
        then $\textsc{alg}$ switches.
    \end{itemize}
    Then,
    \(
        R_{\textsc{alg}}(T) = O\big(T^{2/3}\sqrt{\log T}\big).
    \)
\end{theorem}

In Section~\ref{sec:algs}, we describe two natural procedures designed to satisfy conditions (C1) and (C2). More generally, Theorem~\ref{thm:regretbound} suggests that the class of valid procedures is broad: sublinear regret relative to the oracle can be viewed as a minimal requirement for any reasonable algorithm for this problem.
The proof relies on three main arguments. First, the lower bound on the decision epoch, together with finite VC dimension and ERM training, implies that the challenger trained at $t_{\textsc{alg}}$ is within $O(T^{-1/3}\sqrt{\log T})$ of the best attainable challenger in terms of per-period performance gap. Second, condition~(C2) ensures that the algorithm’s terminal switch/discard decision is consistent with the optimal decision whenever the value gap is sufficiently large. Third, the upper bound on the decision epoch controls the cost of delay: if discarding is optimal, the algorithm has incurred only $O(T^{2/3}\sqrt{\log(T)})$ sunk pre-decision costs before stopping. Combining these three bounds yields regret of order $O(T^{2/3}\sqrt{\log T})$.
Note that condition~(C2) is not intended as a directly implementable switching rule: the quantity $\Delta V(t_{\textsc{alg}})$ is not observable and must, in practice, be estimated from data. Rather, (C2) formalizes a sufficient condition under which the regret guarantee holds: the algorithm must behave as if it had an estimate of $\Delta V(t_{\textsc{alg}})$ whose error is of order $O(T^{2/3}\sqrt{\log T})$.

\section{Algorithm Design} \label{sec:algs}
We now turn to the design of practical stopping rules. 
At each decision epoch, the decision maker observes only an empirical estimate of the current challenger’s performance and must decide whether to switch, discard, or continue. An effective procedure must therefore balance statistical uncertainty, which favors waiting for more data, and the value of timely adoption, which favors stopping early once the challenger is sufficiently attractive. 
We present three algorithms:
\begin{itemize}
    \item The first is a conservative sequential evaluation rule that uses confidence bounds to account for estimation error. It serves as a benchmark for sequential procedures: it reacts to the current estimated value of switching, but does not attempt to anticipate future improvements in the challenger.
    \item The second is a one-shot evaluation rule. It trains a single challenger at a pre-specified epoch and then decides whether to switch or discard based on its empirical performance. This algorithm is intentionally simple: it is meant to illustrate a minimal procedure satisfying the timing and decision-consistency requirements of Theorem~\ref{thm:regretbound}.
    \item The third is a look-ahead sequential rule that uses a local estimate of the learning curve to project future performance. Although this extrapolation is heuristic and need not correctly describe the global shape of the learning curve, we show that the resulting confidence-adjusted procedure still satisfies the structural conditions of Theorem~\ref{thm:regretbound}. The experiments in Section~\ref{sec:numerical} show that this look-ahead rule is also the best-performing procedure empirically.
\end{itemize}   
Throughout this section, the theoretical statements and displayed value equations use the main finite-horizon setting. 
For any candidate per-sample gap \(g\) at time \(t\), define
\(
V_{\rm switch}(t;g)
=
-cnt-c_s+n(T-t)(g-c)
\),
\(
V_{\rm discard}(t)=-cnt,
\)
and
\(
\Delta V(t;g)
=
V_{\rm switch}(t;g)-V_{\rm discard}(t)
=
-c_s+n(T-t)(g-c).
\)
However, the algorithm boxes are written more abstractly in terms of \(G\), \(V_{\rm switch}\), \(V_{\rm discard}\), and \(\Delta V\). This keeps the algorithms compatible with the empirical implementation in Section~\ref{sec:numerical}, where the same objects are computed using general cost and discounting specifications.

\paragraph{Designing decision epochs.}

In Section \ref{sec:theory}, we assumed that the decision maker can in principle act after each step (i.e., $\mathcal{T}_D = \mathcal{T}$).
In practice, however, it is rarely optimal or realistic to retrain this frequently; instead, the choice of decision epochs effectively coarsens the time scale and introduces a trade-off between responsiveness and cost. Decision epochs are algorithmic design choices, determined by how frequently the organization elects to revisit the deployment question. 
As we now design practical algorithms, we re-introduce this distinction (i.e., $\mathcal{T}_D \subseteq \mathcal{T}$), shown in Figure \ref{fig:timesteps-vs-epochs}. We index decision epochs by $k \in [|\mathcal{T}_D|]$ and write $t_k \in\mathcal{T}_D$ for the actual time step at which epoch $k$ occurs, so that the $k$th evaluation uses the $N_{t_k}$ full-feature samples collected up to time~$t_k$.
\begin{figure}[h!]
\centering
\begin{tikzpicture}[xscale=1.2, yscale=1.0, >=stealth]
    \draw[->] (0,0) -- (10,0) node[right] {};

    \foreach \x [count=\i from 1] in {0.5,1.5,...,9.5} {
        \draw[->, thick, gray!70] (\x,-0.1) -- (\x,-0.8);
        \node[below, gray!70] at (\x,-0.8) {\scriptsize $t=\i$};
    }
    \node[below, gray!70] at (5,-1.1) {\small Time steps (Data/batch arrivals: problem parameter)};

    \foreach \x [count=\j from 1] in {0.5,1.5,3.5,6.5} {
        \draw[->, thick, blue!70] (\x,0.1) -- (\x,0.8);
        \node[above, blue!70] at (\x,0.8) {\scriptsize $t_{\j}$};
    }
    \node[above, blue!70] at (5,1.1) {\small Decision epochs (Evaluation points: algorithm design)};

    \node[below] at (10,-0.2) {\scriptsize $t$};
\end{tikzpicture}
\caption{Time steps vs. decision epochs}
\medskip \footnotesize \parbox{0.95\linewidth}{\textit{Note.} 
Time steps (gray, data arrivals) are inherent to the problem and exogenous. Decision epochs (blue, evaluation points), here geometrically spaced, are chosen by the algorithm designer and determine when deployment decisions are revisited.}
\label{fig:timesteps-vs-epochs}
\end{figure}

Decision-epoch design creates a cost-responsiveness trade-off: frequent epochs reduce delay but increase retraining and validation costs.
Section~\ref{ec:sec4-epoch-design} formalizes this trade-off under \(C_{\rm train}(t)=c_{\rm train}N_t^q\) (as per Remark~\ref{rem:costs}). Proposition~EC.2 shows that
geometric schedules are training-cost efficient: for \(q>0\), their cumulative retraining cost up to
the oracle time \(t^*\) is \(\Theta((t^*)^q)\), the same order as training once at \(t^*\),
whereas uniform schedules cost \(\Theta((t^*)^{q+1}/\Lambda)\). For \(q=0\), geometric schedules
use only \(\Theta(\log t^*)\) retraining rounds instead of \(\Theta(t^*/\Lambda)\). Proposition~EC.3
shows that the resulting responsiveness loss is small under the power-law learning curve. We
therefore use geometric schedules in the theoretical and empirical analyses below.

\subsection{Greedy Sequential Evaluation (GSE) Algorithm} \label{ssec:alg-gse}
We start with a {sequential} evaluation procedure that incorporates {statistical uncertainty} into the switching decision. To do so, at each decision epoch, we compute conservative lower and upper bounds (LB, UB) on the estimated value of switching, using confidence intervals that shrink as more data accumulate. We switch to (or discard) the challenger when there is sufficient statistical evidence that the value of switching is positive (or negative).
\begin{algorithm}[h!]
\caption{Greedy Sequential Evaluation (GSE)} \label{alg:seq-conf}
\begin{algorithmic}[1]
\State \textbf{Input:} decision epochs $\mathcal{T}_D$, split ratio $\rho$, confidence parameter $\gamma$
\For{$t_k \in \mathcal{T}_D$}
    \State Train $f_C^{(t_k)}$ using $(1-\rho)N_{t_k}$ training samples
    \State Compute $\widehat{G}(t_k), \widehat{V}_{\text{switch}}^{\mathrm{LB}}(t_k; \gamma), \widehat{V}_{\text{switch}}^{\mathrm{UB}}(t_k; \gamma), \widehat{\Delta V}^{\mathrm{LB}}(t_k; \gamma)$ using $\rho N_{t_k}$ validation samples
    \State \textbf{if} $\widehat{V}_{\text{switch}}^{\mathrm{LB}}(t_k; \gamma) > 0$ \textbf{then} SWITCH
    \State \textbf{if} $\widehat{V}_{\text{switch}}^{\mathrm{UB}}(t_k; \gamma) < 0$ \textbf{then} $\Big\{$
        \textbf{if} $\widehat{\Delta V}^{\mathrm{LB}}(t_k; \gamma) > 0$ \textbf{then} SWITCH; \textbf{else} DISCARD $\Big\}$
\EndFor
\State Train $f_C^{(t_K)}$; 
compute $\widehat{G}(t_K)$, $\widehat{\Delta V}(t_K)$;
\textbf{if} $\widehat{\Delta V}(t_K) > 0$ \textbf{then} SWITCH \textbf{else} DISCARD
\end{algorithmic}
\end{algorithm}

We provide the full procedure in Algorithm \ref{alg:seq-conf}. 
At each decision epoch $t_k \in \mathcal{T}_D$, the algorithm randomly splits the $N_{t_k}$ full-feature samples into a training set of size $(1-\rho)N_{t_k}$ and a validation set of size $\rho N_{t_k}$, and retrains the challenger $f_C^{(t_k)}$ on the training set. The empirical per-sample gap is
$
    \widehat{G}(t_k)
    := \frac{1}{\rho N_{t_k}} 
       \sum_{i \in \mathcal{V}_{t_k}}
       \big(
           \ell(f_C^{(t_k)}(x_i), y_i)
           -
           \ell(f_I(x_i), y_i)
       \big),
$
where $(x_i, y_i)$ denotes the $i$th sample in the validation set \(\mathcal V_{t_k}\).
To account for estimation error in \(\widehat G(t_k)\), we use the confidence half-width
\(
\delta_k=\frac{\gamma}{\sqrt{\rho N_{t_k}}},
\)
so that, with the chosen confidence level, the true gap satisfies
\(
G(t_k)\in\big[\widehat G(t_k)-\delta_k,\widehat G(t_k)+\delta_k\big].
\)
Confidence grows as more samples accumulate (since $\delta_k$ decreases as $N_{t_k}$ grows), whereas \(\gamma>0\) controls how conservative the bound is.
We then construct the following lower and upper confidence bounds on the value of switching:
$\widehat{V}_{\text{switch}}^{\mathrm{LB}}(t_k; \gamma)
    =
V_{\rm switch}\!\left(t_k;\widehat G(t_k)-\delta_k\right),$
$ \widehat{V}_{\text{switch}}^{\mathrm{UB}}(t_k; \gamma)
=
V_{\rm switch}\!\left(t_k;\widehat G(t_k)+\delta_k\right).$
If $\widehat{V}_{\text{switch}}^{\mathrm{LB}}(t_k; \gamma) > 0$, the decision maker switches immediately.
If $\widehat{V}_{\text{switch}}^{\mathrm{UB}}(t_k; \gamma) < 0$, the estimated value of switching is negative even under optimistic gaps, so the decision maker stops training.
Although it is now estimated that the returned value will be negative, there may still be some value to switching to the challenger.
Indeed, since all acquisition and training costs up to $t_k$ are sunk, we still need to compare the challenger to discarding.  
We compute the conservative incremental value of switching as
$\widehat{\Delta V}^{\mathrm{LB}}(t_k; \gamma)
=
\Delta V\!\left(t_k;\widehat G(t_k)-\delta_k\right)$.
If $\widehat{\Delta V}^{\mathrm{LB}}(t_k; \gamma) > 0$, the algorithm switches to the challenger; 
otherwise it discards.
If neither condition holds, i.e.,
\(
\widehat{V}_{\text{switch}}^{\mathrm{LB}}(t_k; \gamma) < 0
\ \text{and}\ 
\widehat{V}_{\text{switch}}^{\mathrm{UB}}(t_k; \gamma) > 0,
\)
the available data are insufficient to make a statistically confident decision,
and the algorithm proceeds to the next decision epoch.
If neither confidence condition is met before the final epoch, the algorithm makes the terminal empirical decision at \(t_K\): switch if $\widehat{\Delta V}(t_{K}) = \Delta V\!\left(t_{K};\widehat G(t_{K})\right)>0$, and discard otherwise.

\begin{remark}\label{rem:gse-confidence}
The half-width \(\delta_k\) is chosen to make the gap estimates uniformly valid across
decision epochs. Specifically, using a union bound over \(K=|\mathcal T_D|\) epochs and a
Hoeffding bound for bounded validation gains, if \(\ell\in[-1,1]\), then taking
\(
\gamma=\sqrt{8\log(2K/\eta)}
\)
ensures that, with probability at least \(1-\eta\),
\(
|\widehat G(t_k)-G(t_k)|
\le
\delta_k
=
\frac{\gamma}{\sqrt{\rho N_{t_k}}}
\)
for all $k=[K].$
On this event,
\(
\widehat V_{\rm switch}^{\rm LB}(t_k;\gamma)
\le
V_{\rm switch}(t_k)
\le
\widehat V_{\rm switch}^{\rm UB}(t_k;\gamma)
\)
for all $k$.
Thus a positive lower bound certifies that switching has positive value, while a negative
upper bound certifies that the current challenger has negative switching value. In the
latter case, GSE stops experimentation and makes the terminal switch/discard decision using
the conservative incremental value \(\widehat{\Delta V}^{\rm LB}(t_k;\gamma)\). This validates
the confidence logic of GSE, but does not imply that GSE satisfies the meta-regret
conditions of Theorem~\ref{thm:regretbound}.
\end{remark}

\subsection{One-Shot Evaluation (OSE) Algorithm} \label{ssec:alg-ose}
The GSE algorithm is intuitive and provides a useful sequential benchmark, but it need not satisfy the timing condition of Theorem~\ref{thm:regretbound}: depending on the confidence bounds and the realized empirical estimates, it may stop too early or too late. We therefore introduce a deliberately simple benchmark designed to satisfy the structural requirements of Theorem~\ref{thm:regretbound}. The algorithm evaluates the challenger only once, at a pre-specified epoch of order $T^{2/3}$, and then decides whether to switch or discard based on the empirical incremental value of switching.

\begin{algorithm}[h!] 
\caption{One-Shot Evaluation (OSE)} \label{alg:straw-man}
\begin{algorithmic}[1]
\State \textbf{Input:} evaluation epoch $t_{\rm OSE} \in \mathcal{T}_D$, split ratio $\rho$
\State Collect $N_{t_{\rm OSE}}$ samples; split into training and validation set of sizes $(1-\rho)N_{t_{\rm OSE}}$ and $\rho N_{t_{\rm OSE}}$
\State Train challenger $f_C^{(t_{\rm OSE})}$ on training set; compute $\widehat{G}(t_{\rm OSE}), \widehat{\Delta V}(t_{\rm OSE})$ on validation set
\State \textbf{if} $\widehat{\Delta V}(t_{\rm OSE}) > 0$ \textbf{then} SWITCH; \textbf{else} DISCARD
\end{algorithmic}
\end{algorithm}

We provide the full procedure in Algorithm \ref{alg:straw-man}. 
At a pre-determined epoch $t_{\rm OSE}\in\mathcal T_D$, the $N_{t_{\rm OSE}}$ samples collected to date are randomly split into a training set of size $(1-\rho)N_{t_{\rm OSE}}$ and a validation set $\mathcal{V}_t$ of size $\rho N_{t_{\rm OSE}}$. The challenger $f_C^{(t_{\rm OSE})}$ is trained using the training set and its per-sample performance gap $\widehat{G}(t_{\rm OSE})$  is estimated using validation set. 
The algorithm computes
$\widehat{\Delta V}(t_{\rm OSE})
=
\Delta V\!\left(t_{\rm OSE};\widehat G(t_{\rm OSE})\right)$;
it switches if \(\widehat{\Delta V}(t_{\rm OSE})>0\), and discards otherwise.
By design, with  appropriately chosen evaluation epochs, the algorithm satisfies the conditions of Theorem \ref{thm:regretbound}:
\begin{proposition}
\label{prop:strawmanverifiesregret}
    Under the main finite-horizon setting, there exist constants $w_2, w_3,$ and $w_4$ (defined in the 
    EC) such that the OSE algorithm,  
    when implemented with a decision epoch $t_{\rm OSE}\asymp T^{2/3}$, satisfies conditions (C1) and (C2) of Theorem~\ref{thm:regretbound} with probability at least $1 - 2/T$.
\end{proposition}
Condition~(C1) holds by construction because the evaluation epoch is chosen on the $T^{2/3}$ scale. Condition~(C2) follows from concentration of the validation estimate $\widehat G(t_k)$: since the validation set has size of order $T^{2/3}$, the per-period gap estimate is accurate up to order $T^{-1/3}\sqrt{\log T}$, which translates into an error of order $T^{2/3}\sqrt{\log T}$ for the value difference $\Delta V(t_{\rm OSE})$.
The prescription \(t_{\rm OSE}\asymp T^{2/3}\) is an asymptotic order condition. In practice, one must still choose the constant multiplying \(T^{2/3}\), or equivalently choose a feasible evaluation epoch from \(\mathcal T_D\). This is the main practical weakness of OSE. In the empirical analysis, we therefore report several fixed OSE benchmarks rather than presenting OSE as an adaptive rule.

\subsection{Look-ahead Sequential Evaluation (LSE) Algorithm} \label{ssec:alg-lse}
The previous algorithms evaluate the challenger based solely on its current estimated performance, possibly adjusted for statistical uncertainty, but they do not explicitly reason about how the challenger may continue to improve as additional data arrive. We now introduce a {sequential} evaluation algorithm that attempts to {anticipate} such future improvements by extrapolating trends in estimated gaps. Specifically, the algorithm fits a local linear approximation to the empirical learning curve and uses its slope to construct an optimistic projection of how much further performance could increase, thereby assessing whether waiting for more data may offer a better opportunity to switch.
\begin{algorithm}[h!]
\caption{Look-ahead Sequential Evaluation with/without Confidence Adjustment (LSE/LSEc)} \label{alg:seq-slope}
\begin{algorithmic}[1]
\State \textbf{Input:} epochs $\mathcal{T}_D$, split ratio $\rho$, confidence parameter $\gamma$, smoothing parameter $w$
\State Train $f_C^{(t_1)},\dots,f_C^{(t_{w-1})}$; compute $\widehat{G}(t_1),\dots,\widehat{G}(t_{w-1})$
\For{$k = w,\dots,|\mathcal{T}_D|-1$}
    \State Train $f_C^{(t_k)}$; compute $\widehat{G}(t_k)$; estimate $\hat{s}_{k}$ from $\widehat{G}(t_{k-w+1}),\dots,\widehat{G}(t_k)$ with confidence $\gamma$
    \State Compute projected values $\widehat{V}_{\text{switch}}^{\mathrm{UB}}(t_{k'})$ for $k'> k$ and current values $\widehat{V}_{\text{switch}}(t_k), \widehat V_{\rm discard}(t_k)$
    \If{$\max\!\Big(\widehat{V}_{\text{switch}}(t_k),\; \widehat V_{\rm discard}(t_k) \Big)
     \;\ge\; \max_{k'> k}\widehat{V}_{\text{switch}}^{\mathrm{UB}}(t_{k'})$}
     \State \textbf{if} $\widehat{\Delta V}(t_k) > 0$ \textbf{then} SWITCH; \textbf{else} DISCARD
    \EndIf
\EndFor
\State Train $f_C^{(t_K)}$; 
Compute $\widehat{G}(t_K)$, $\widehat{\Delta V}(t_K)$;
\textbf{if} $\widehat{\Delta V}(t_K) > 0$ \textbf{then} SWITCH \textbf{else} DISCARD
\end{algorithmic}
\end{algorithm}

We provide the full procedure in Algorithm~\ref{alg:seq-slope}. 
At each decision epoch $t_k \in \mathcal{T}_D$, we observe $N_{t_k}$ full-feature samples and compute the empirical per-sample gap $\widehat{G}(t_k)$, as in the previous algorithms. 
The slope-based algorithm then proceeds in three steps: (i) estimate a local slope of the empirical learning curve using recent evaluations, (ii) use this slope to construct optimistic projections of future gaps, and (iii) compare the current estimated value of switching with these projections to determine whether waiting is potentially beneficial. We now detail each component.


\paragraph{Slope estimation.}
The algorithm maintains a local estimate $\hat{s}_k$ of the slope of the empirical learning curve at epoch $k$. 
Intuitively, the slope estimate $\hat{s}_k$ can be thought of as the incremental improvement in the challenger’s gap per additional sample, as captured by a simple finite difference such as
\(
\frac{\widehat{G}(t_k)-\widehat{G}(t_{k-1})}{(1-\rho)\cdot(N_{t_k}-N_{t_{k-1}})}.
\)
In practice, direct finite differences are noisy, and we consider the following robustification procedures, each resulting in a different variant of Algorithm~\ref{alg:seq-slope}:
\begin{itemize}
    \item[(i)] {Smoothing:}  
    Instead of relying only on the most recent estimate of the gaps, we smooth the finite difference by looking at the last $w\geq2$ points. We first apply a monotonicity correction: if the gaps decrease across $t_{k-w+1},\dots,t_k$, the values are replaced by their average. We then perform a least-squares linear regression of these corrected gaps on the sample counts $N_{t_{k-w+1}},\dots,N_{t_k}$, and take the fitted slope as $\hat{s}_k$.  
    In practice, we find that a window size of $w=3$ is sufficient: learning rates typically decrease quickly, so using four or more points tends to oversmooth the curve and underestimate the local slope. We refer to this variant as LSE.

    \item[(ii)] {Confidence-adjustment:}  
    We adjust the finite difference using a confidence term. Let $\delta_k := \gamma / \sqrt{\rho N_{t_k}}$ be the confidence adjustment.  
    If the most recent improvement is negative, i.e., $\widehat{G}(t_k)-\widehat{G}(t_{k-1})<0$, we set
    \(
    \hat{s}_k = \frac{2\delta_k}{(1-\rho)\cdot(N_{t_k}-N_{t_{k-1}})}.
    \)
    Otherwise, we use the uncertainty-adjusted finite difference
    \(
    \hat{s}_k
    =
    \frac{(\widehat{G}(t_k)+\delta_k)-(\widehat{G}(t_{k-1})-\delta_k)}{(1-\rho)\cdot(N_{t_k}-N_{t_{k-1}})}.
    \)
    In practice, we find that the confidence adjustment counterbalances the noise that motivates a $w\geq3$ window, so $w=2$ is sufficient. We refer to this variant as LSEc. 
\end{itemize}

\begin{remark}
In both variants, $\hat{s}_k$ is a nonnegative, optimism-adjusted estimate of the local slope. 
We compare the effect of smoothing and confidence adjustment in two cases:
\begin{itemize}
    \item When $w=2$, the (non-adjusted) LSE slope is
\(
\hat{s}_k^{\text{LSE}}
=
\max\!\left\{
    0,\;
    \frac{\widehat{G}(t_k)-\widehat{G}(t_{k-1})}
         {(1-\rho)\,(N_{t_k}-N_{t_{k-1}})}
\right\},
\)
whereas the confidence-adjusted LSEc slope becomes
\(
\hat{s}_k^{\text{LSEc}}
=
\max\!\left\{
    \frac{2\delta_k}{(1-\rho)\,(N_{t_k}-N_{t_{k-1}})},\;
    \frac{(\widehat{G}(t_k)+\delta_k)-(\widehat{G}(t_{k-1})-\delta_k)}
         {(1-\rho)\,(N_{t_k}-N_{t_{k-1}})}
\right\}.
\)
Since $\delta_k>0$, the numerator of $\hat{s}_k^{\text{LSEc}}$ is always at least as large as that of $\hat{s}_k^{\text{LSE}}$, so $\hat{s}_k^{\text{LSEc}} \ge \hat{s}_k^{\text{LSE}}$ for all $k$. Thus, confidence adjustment strictly increases optimism and tends to delay stopping.

    \item When $w=3$, LSE fits a regression slope over $\widehat{G}(t_{k-2}),\widehat{G}(t_{k-1}),\widehat{G}(t_k)$.
In contrast, LSEc still relies on the two-point, confidence-adjusted difference. In our empirical trajectories, gaps often accelerate so that
$\widehat{G}(t_{k-1}) - \widehat{G}(t_{k-2})\gg \widehat{G}(t_{k}) - \widehat{G}(t_{k-1})$,
and the confidence term $\delta_k$ is relatively small. As a result, with $w=3$ for LSE and $w=2$ for LSEc, the (non-adjusted) LSE can produce larger slopes and stop later than LSEc.
\end{itemize}
\end{remark}

\paragraph{Projected gaps and decision logic.}
Given a slope estimate $\widehat s_k$, the projected upper bound on the future gap is
\(
\widehat G^{\rm UB}(t_{k'})
=
\widehat G(t_k)
+
(1-\rho)\big(N_{t_{k'}}-N_{t_k}\big)\widehat s_k,
\ k'>k.
\)
Then, the projected value of switching at future epoch \(t_{k'}\) is
\(
\widehat V_{\rm switch}^{\rm UB}(t_{k'})
=
V_{\rm switch}\!\left(t_{k'};\widehat G^{\rm UB}(t_{k'})\right)
\).
The current estimated switching value is
\(
\widehat V_{\rm switch}(t_k)
=
V_{\rm switch}\!\left(t_k;\widehat G(t_k)\right)
\),
whereas the current discard value is
\(
\widehat V_{\rm discard}(t_k)=V_{\rm discard}(t_k)
\).
The algorithm stops when the best current terminal action dominates every optimistic
future switching value, i.e., 
\(
\max\left\{
\widehat V_{\rm switch}(t_k),
\widehat V_{\rm discard}(t_k)
\right\}
\ge
\max_{k'>k}\widehat V_{\rm switch}^{\rm UB}(t_{k'}).
\)
If it stops, it switches if
\(
\widehat{\Delta V}(t_k)
=
\Delta V\!\left(t_k;\widehat G(t_k)\right)>0
\), and discards otherwise.

\paragraph{Theoretical guarantees of the algorithm.}

The LSEc algorithm implicitly assumes that the challenger’s learning curve is, on average, increasing and concave (up to noise): it extrapolates local slopes to construct optimistic projections of future gaps. One may therefore wonder how robust the procedure remains when this structure does not hold, and whether theoretically grounded upper bounds on future gains—such as those derived from Rademacher complexity or VC-dimension—would be better suited for constructing projections. However, it turns out that the algorithm still satisfies Theorem~\ref{thm:regretbound}'s requirement irrespectively of the learning curve's actual behavior thanks to the confidence adjustment. The reason behind this phenomenon is that the regret guarantee does not require the local linear projection to recover the full learning curve. Instead, the confidence adjustment prevents the algorithm from stopping before the \(T^{2/3}\) scale, while validation concentration ensures that the terminal switch/discard decision is correct up to the tolerance required by Theorem~\ref{thm:regretbound}.
\begin{proposition}\label{prop:algorithmslopebasedverifiesregret}
    Under the main finite-horizon setting, if $g^*+c>0$, there exist constants $w_2, $ $w_3,$ and $w_4$ (defined in the 
    EC) such that the LSEc algorithm, implemented with a geometric epoch schedule $t_k = \Lambda \lambda^{k-1}$ with ratio $\lambda>1$, satisfies conditions (C1) and (C2) of Theorem~\ref{thm:regretbound} with probability at least $1 - 2/T$.
\end{proposition}
The proof has two parts. First, the confidence adjustment prevents the algorithm from stopping too early: before the $T^{2/3}$ scale, the optimistic projection of future improvement remains large enough to justify continuation. Second, once enough data have accumulated, validation concentration ensures that the empirical switch/discard decision is correct up to the value tolerance required by (C2). Note that the excluded case, $g^*+c=0$ is the one where only the switching cost is non-zero, and thus where the decision point does not matter.

\section{Empirical Validation: Credit Scoring Case Study} \label{sec:numerical}
In this section, we evaluate the algorithms introduced earlier using a large, real-world credit-scoring dataset with alternative data. We show that optimal switching times vary systematically with learning-curve dynamics and cost parameters, consistent with the predictions in Section~\ref{sec:theory}.
We further find that using an appropriate sequential evaluation algorithm can deliver value close to the oracle benchmark and substantially outperform common one-shot baselines.

\subsection{Data}
Our empirical case study uses a comprehensive dataset from Lending Club, a large peer-to-peer lending platform, covering 705{,}302 loans issued between 2014 and 2017. 
For each loan, we observe seven traditional features describing the loan contract (e.g., amount, duration) and borrower characteristics (e.g., monthly income, job tenure). The dataset also provides alternative data on loan applicants, including 20 additional banking features (e.g., number of open credit lines, bank cards, installments, mortgages, and revolving accounts), as well as the loan grade assigned by LendingClub (2 additional features), which can be viewed as a proprietary signal combining standard credit features with nontraditional information (e.g., the applicant’s digital footprint). We define the incumbent feature set using only the 7 traditional features, and the challenger feature set by augmenting these variables with the alternative banking features and LendingClub grade.
The target variable is the repayment outcome (full repayment or default), encoded as $1$ for default and $0$ for non-default. Throughout the sample period, yearly default rates range from 19.23\% to 21.42\%, with an average of 20.87\%.
Finally, the dataset records the funding date of each loan, allowing us to align the empirical analysis with our theoretical framework by (i) specifying a date at which the alternative data become available and (ii) naturally defining the time steps of our process based on loan dates. We provide details on data pre-processing and feature selection for the incumbent and the challenger in Section~\ref{sec:data} of the EC.
 
\subsection{Experimental methodology} \label{ssec:experimental}
Here, we describe the general experimental methodology, which follows the framework introduced in Section~\ref{sec:problem-formulation}. 
The dataset is chronologically ordered by loan funding date. In the experiments, the
sequential process is indexed by decision epochs \(t_k\), and each epoch corresponds to
an incoming batch of \(n_{t_k}\) observations (i.e., we allow for varying batch sizes) and a cumulative number \(N_{t_k}\) of full-feature observations. To generate statistically diverse trajectories while preserving temporal ordering, each sample path draws without
replacement one half of each chronological block: if the batch at epoch \(k\) contains \(n_{t_k}\)
observations, we sample \(n_{t_k}\) observations from the next \(2 n_{t_k}\) observations in the
ordered stream. Blocks are disjoint across epochs.

First, {the incumbent model} is trained on the first $N_0$ data points, covering 2014--2015, resulting in an effective dataset of size $N_0=189{,}250$. It uses an initial feature set $\mathcal{I}$ of size $|\mathcal{I}|=7$: annual income, job tenure, debt-to-income ratio, FICO score, funding amount, loan duration, and loan purpose. Starting in 2016, alternative data become available (e.g., loan grade, number of open credit lines, bank cards, installments, mortgages, revolving accounts), expanding the feature set to $\mathcal{C}$ with $|\mathcal{C}|=29$, and training of the challenger begins. 


At each epoch $k$ (whose design is detailed in the next subsection), we execute three steps:
\begin{enumerate}
\item {Data collection:} collect the next $N_{t_k}-N_{t_{k-1}}$ observations from the temporally ordered dataset.
\item {Model training:} split the $N_{t_k}$ collected observations evenly into a training set and a holdout set; train the challenger on the training set and empirically estimate the economic performance gap between challenger and incumbent on the holdout set.
\item {Decision:} the algorithm chooses whether to switch, discard, or continue based on this empirical estimate and all relevant parameters (e.g., costs, sample counts).
\end{enumerate}
Once samples for all epochs are generated and challenger models are trained, we enter a decision-evaluation phase in which we empirically estimate the value of switching at each epoch along the realized sample path. At epoch $k$, we estimate the average economic performance gap between the incumbent
and challenger on samples from each future timestep $\tau>t_k$, denoted $\widehat G_{\tau,k}$,
and compute:
$
\overline{V}_{\text{switch}}(t_k)
=
-\sum_{\tau \le t_k}\beta^{\tau} C_{\mathrm{pre}}(\tau)
 - \beta^{t_k} c_s
 + \sum_{\tau > t_k} 
    \beta^{\tau}\big[n_\tau\,\hat{G}_{\tau,k} - C_{\mathrm{post}}(\tau)\big].\ 
$
The oracle reported in the experiments bases its decision on $\overline{V}_{\text{switch}}(t_k)$. This yields a stronger benchmark than the oracle defined in Section~\ref{ssec:theory-oracle}: the latter selects the optimal epoch {on average} across trajectories, whereas the experimental oracle adapts to the realized sample path. Note that $C_{\mathrm{pre}}(\tau)$ is not necessarily the same for our deployed algorithms and the oracle: typically (except for the one-shot algorithm), our algorithms incur training costs at each decision epoch before stopping, whereas the oracle incurs training cost only at its switching epoch. We hide this nuance from the notation.

\begin{figure}[htb]
\centering 
\begin{minipage}{0.8\textwidth}
  \centering
  \begin{subfigure}[b]{0.3\textwidth}
  \hspace{-14pt}
    \includegraphics[trim={0 0 0 0.6cm},clip,width=1\textwidth]{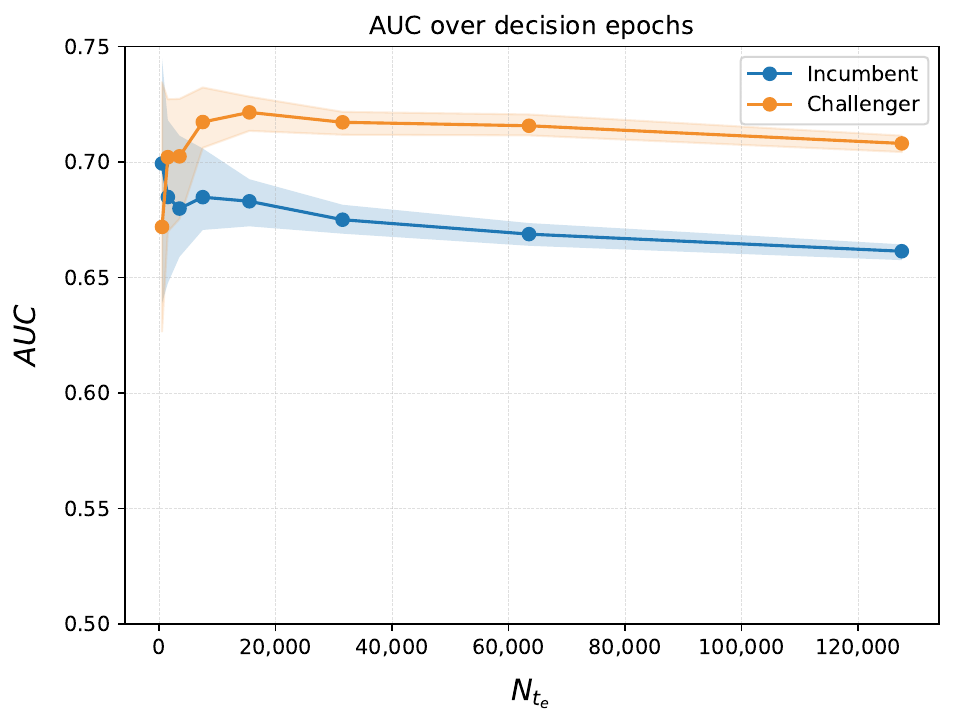}
    \caption{LR}
  \end{subfigure} 
  \hfill 
  \begin{subfigure}[b]{0.3\textwidth}
  \hspace{-14pt}
    \includegraphics[trim={0 0 0 0.6cm},clip,width=1\textwidth]{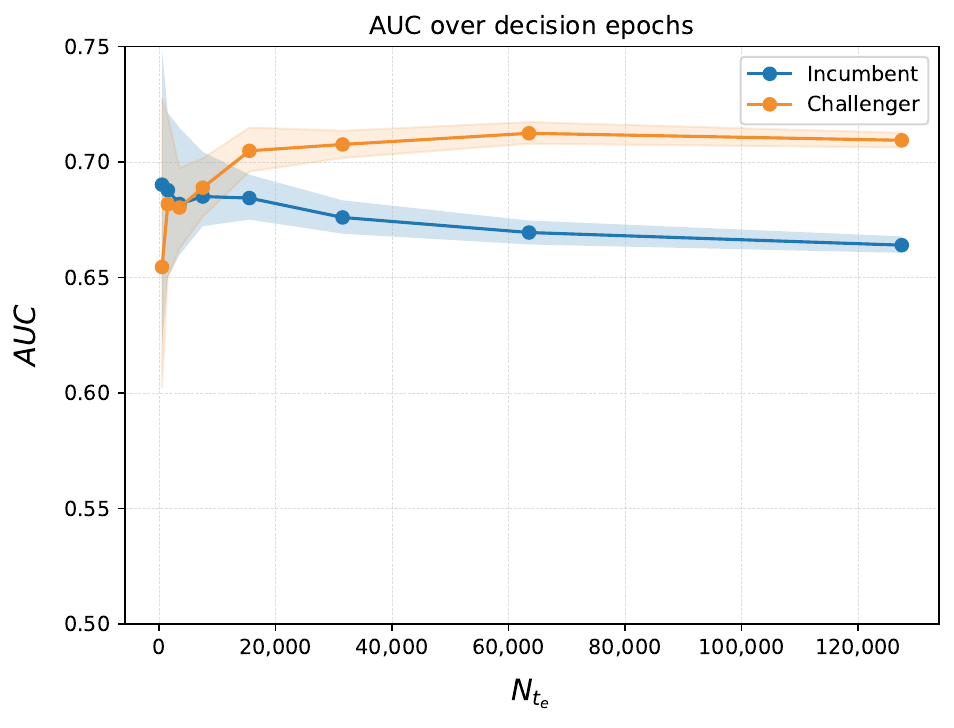}
    \caption{LightGBM} 
  \end{subfigure} 
\end{minipage}
    \caption{AUC of incumbent and challenger } \label{fig:AUC}
 \footnotesize
    \parbox{0.95\linewidth}{\textit{Note.} This figure reports the AUC of the incumbent (blue curve) and challenger (orange curve) models as a function of \(N_{t_e}\), the sample size available at each decision epoch \(t_k \in \mathcal{T}_D\), with shaded areas indicating 90\% confidence intervals.. The challenger is a LR in panel (a) and a LightGBM in panel (b). At each decision epoch, the challenger is retrained using half of the available sample \(N_{t_e}\). AUC for both the incumbent and the challenger is computed on the corresponding cumulative test set, consisting of the remaining half of the data from the current epoch through the end of the process. Each curve shows the average AUC across 30 sample paths (see Section \ref{ssec:experimental}). The decreasing AUC is due to a small distribution shift in the data.}
\end{figure}

We adopt this evaluation methodology for two reasons. First, obtaining reliable average economic-gap estimates would require holding out additional data, which would reduce the data available to generate diverse sample paths. Second, the dataset exhibits mild but non-negligible time variability (see Figure \ref{fig:AUC}); evaluating performance via $\overline{V}_{\text{switch}}(t_k)$ preserves this variability and demonstrates that our approach remains robust under small distributional shifts.


\subsection{Core scenarios: early vs. late switch } \label{ssec:scenario_selection}

We focus on two core scenarios that induce qualitatively distinct behavior concerning the optimal switching point:
an {early-switch} scenario (E.1) and a {late-switch} scenario (E.2). 
These scenarios are defined by the parameters outlined below.
We report additional parameter settings and scenarios (e.g., cases in which discarding is optimal), and robustness checks in the EC (Section~\ref{ec:empirical_results}).  

\paragraph{Core scenarios.}
The two scenarios differ in the learning environment, induced by the model class and the training cost (these are the key drivers for early vs. late switch). In E.1 ({early switch}), the model is logistic regression (LR) and the per-sample training cost is $c_{\mathrm{train}}=0.075$ (we explain the meaning of this value shortly). In E.2 ({late switch}), the model is LightGBM and the per-sample training cost is $c_{\mathrm{train}}=0.005$. 
E.1 and E.2 share the following parameters:
\begin{itemize}
\item \textbf{Sample size parameters.} The incumbent is trained on $N_0=189{,}250$ observations. Over the sequential process, the challenger has access to $N_{t_e}=163{,}400$ observations (the subscript $t_e$ denotes the end of the process). At each decision epoch, we split data evenly into training and testing sets.

\item \textbf{Timing and epoch design parameters.} We treat the data as a time-ordered stream and implement a geometric schedule in which batch sizes grow by a factor of $2$ across $8$ decision epochs. We set the first decision-epoch sample size to $500$, so the cumulative samples available by epoch are $500$, $1{,}500$, $3{,}500$, $7{,}500$, $15{,}500$, $31{,}500$, $63{,}500$, and $127{,}500$. Hence, by construction, our two core scenarios consist of eight decision epochs. The subsequent $35{,}900$ observations are reserved as future data, reflecting the fact that our last decision epoch is not the end of the operational horizon: the decision maker anticipates receiving additional data thereafter and earning potential profits.  

\item \textbf{Economic parameters.} In all experiments, we measure the {economic performance gap} as the difference in AUC between the challenger and the incumbent. 
We write the per-sample economic gap as $(\mathrm{AUC}_{\text{challenger}}-\mathrm{AUC}_{\text{incumbent}})$, so that all costs are expressed in units of ``per-sample gain from a unit increase in AUC.''
For example, $c_{\mathrm{acq}}=0.0025$ means that the acquisition cost per sample is $0.25\%$ of the per-sample gain from a one-point AUC improvement. 
Since scaling gaps and costs is equivalent up to a rescaling of the value function, this normalization is without loss of generality. To simplify parameter selection and better illustrate the findings of
Sections~\ref{sec:theory}-\ref{sec:algs}, we assume a fixed per-sample gain; a constant per-sample acquisition cost, identical before and after the decision; and no switching costs.
In both experiment families (E.1) and (E.2), we hold the discount factor and baseline costs fixed at $\beta=0.95$, $c_{\mathrm{acq}}=0.0025$, and $c_s=0$, varying only the training-related costs across scenarios. 
We revisit those choices in the EC.
\end{itemize}

\begin{remark}
Section~\ref{sec:add_exp} in the EC examines complementary scenarios that vary
(i) sample size, timing, and epoch design parameters (smaller total challenger sample size; smaller batches;  larger batches with higher acquisition costs; no temporal ordering allowing us to investigate the effect of distribution shift), as well as
(ii) economic parameters (no costs; higher acquisition costs; positive switching cost), each evaluated for both LR and LightGBM.
\end{remark}

\begin{figure}[h!]\centering
\begin{minipage}{0.8\textwidth}
  \centering
  \begin{subfigure}[b]{0.475\textwidth}
    \hspace{-10pt}
    \includegraphics[width=1\textwidth]{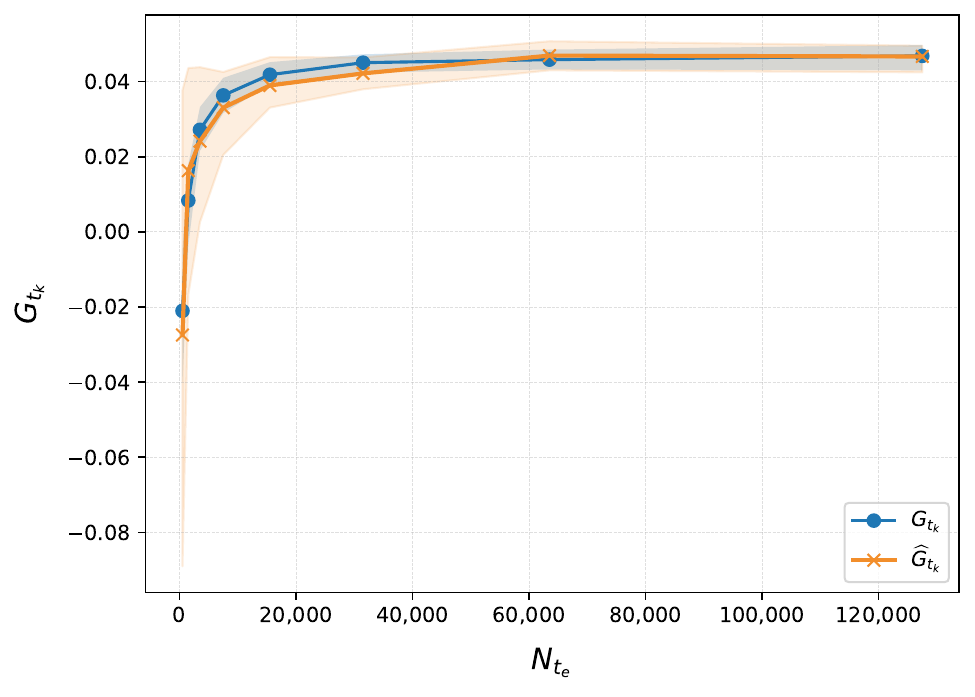}
    \caption{LR}
    \label{fig:sub1}
  \end{subfigure}
  \hfill 
  \begin{subfigure}[b]{0.475\textwidth}
  \hspace{-10pt}
    \includegraphics[width=1\textwidth]{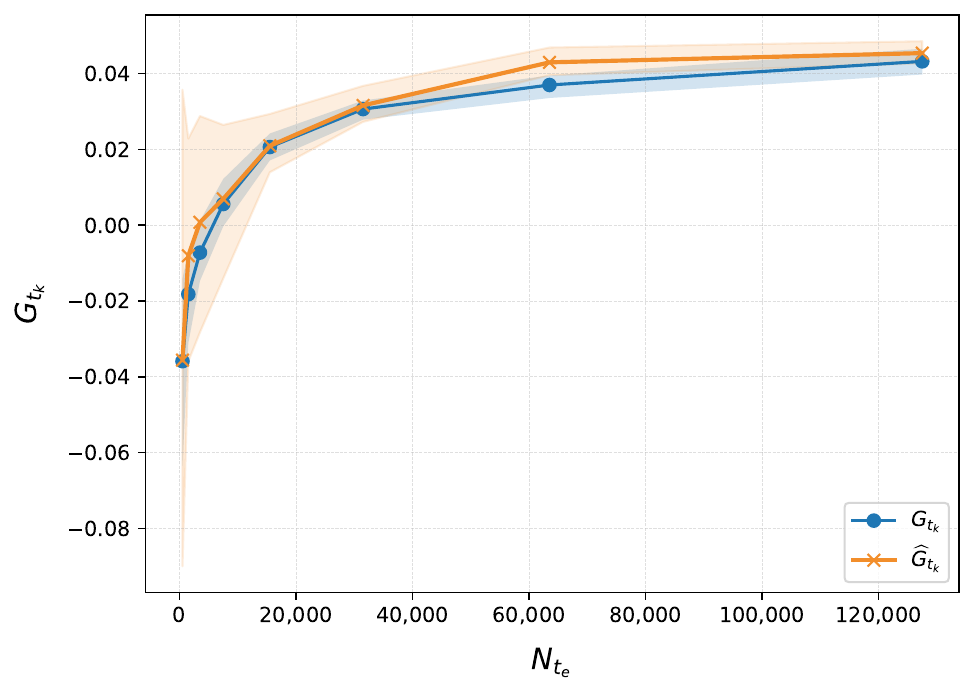}
    \caption{LightGBM}
    \label{fig:sub1}
  \end{subfigure}
\end{minipage}
  \vspace{0.2cm}
  \caption{Learning curves}
  \label{fig:learning_curve}
    \medskip
    
  \footnotesize \parbox{0.95\linewidth}{\textit{Note.} This figure displays the learning curve of LR and LightGBM. The learning curve is defined as the expected economic performance gap ~$G(t_k)$, expressed as a function of $N_{t_e}$ the sample size available at each decision epoch $t_k \in \mathcal{T}_D$. The orange line reports the empirical estimate using the holdout set at time $t_k$, while the blue line reports estimates based on the future data available after $t_k$. Each curve represents the average over the 30 generated sample paths (see Section \ref{ssec:experimental}). The orange and blue intervals represent 90\% confidence intervals.}
\end{figure}

\paragraph{Learning environment.}
Since the shape of the learning curve is central to our problem (Section~\ref{sec:problem-formulation}), we consider two models commonly used in the finance literature to generate qualitatively different learning-curve patterns. We use LR, which remains the standard in credit scoring \citep{lessmann2015benchmarking}, and LightGBM \citep{ke2017lightgbm}, a gradient-boosting framework known for its fast training speed and scalability to large datasets. The hyperparameters of the LightGBM model are selected using cross-validation. 
Figure~\ref{fig:learning_curve} displays the corresponding learning curves. 
Two results stand out: (i) the performance of LR reaches its plateau much more quickly than that of LightGBM; (ii) the performance of LR at the last epoch is close to the performance of LightGBM.

\begin{figure}[htb]
  \centering
  \begin{subfigure}[b]{0.24\textwidth}
    \includegraphics[trim={0 0 0 0.75cm},clip, width=1\linewidth]{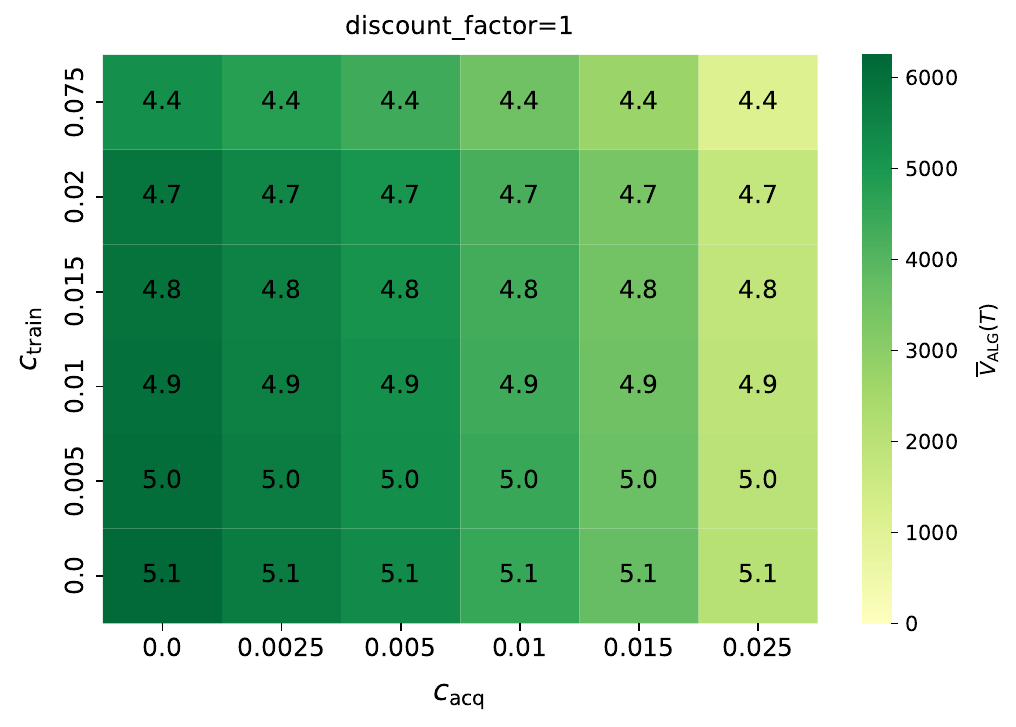}
    \caption{LR: $\beta=1$}
    \label{fig:placeholder}
  \end{subfigure}
  \hfill 
  \begin{subfigure}[b]{0.24\textwidth}
    \includegraphics[trim={0 0 0 0.75cm},clip, width=1\linewidth]{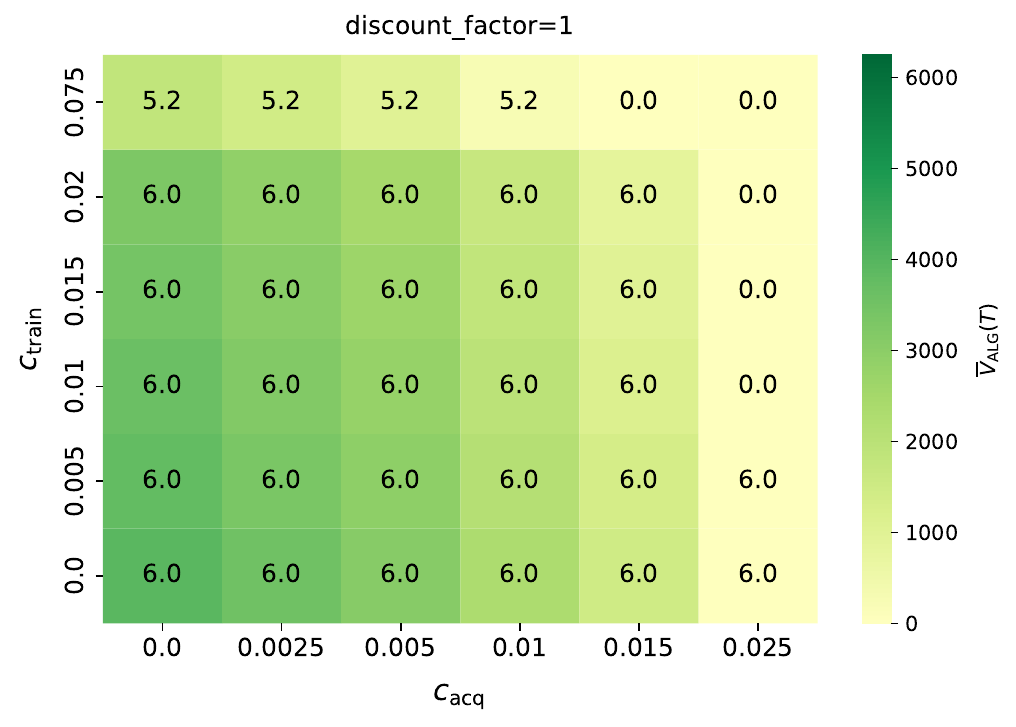}
    \caption{LightGBM: $\beta=1$}
    \label{fig:placeholder}
  \end{subfigure}
  \hfill 
    \begin{subfigure}[b]{0.24\textwidth}
    \includegraphics[trim={0 0 0 0.75cm},clip, width=1\linewidth]{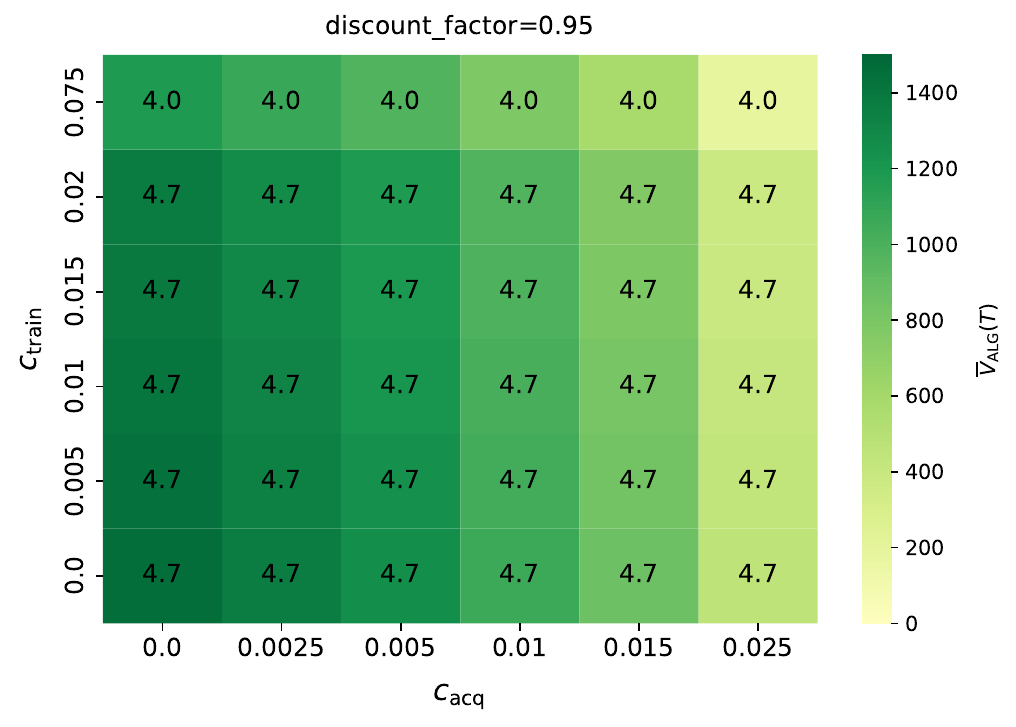}
    \caption{LR: $\beta=0.95$}
    \label{fig:placeholder}
  \end{subfigure}
  \hfill 
  \begin{subfigure}[b]{0.24\textwidth}
    \includegraphics[trim={0 0 0 0.75cm},clip, width=1\linewidth]{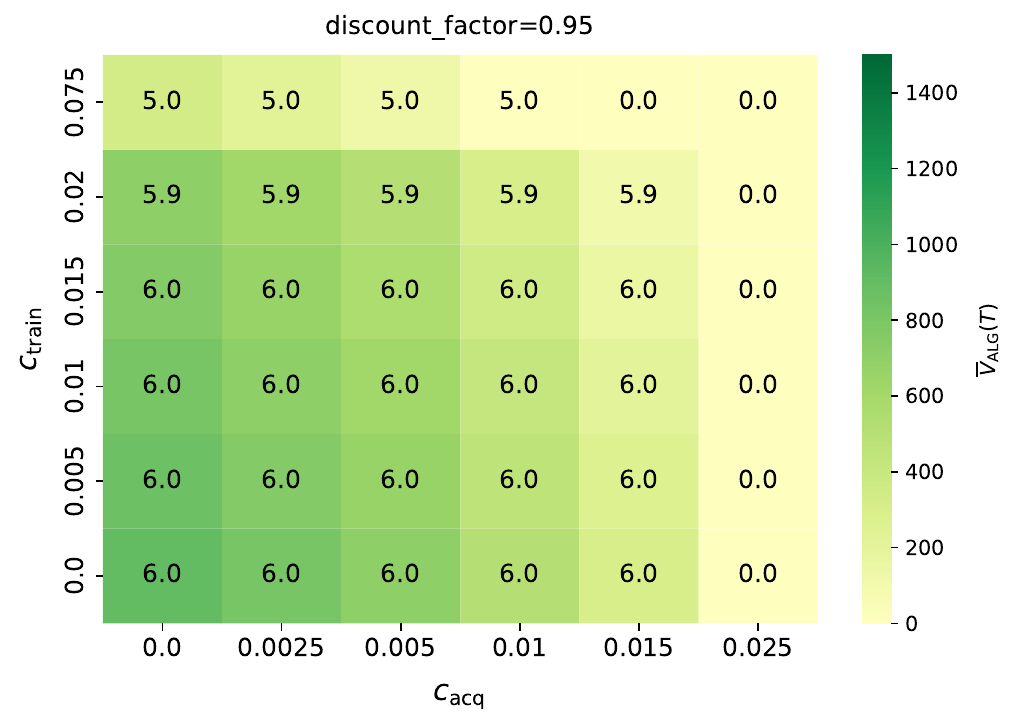}
    \caption{LightGBM: $\beta=0.95$}
    \label{fig:placeholder}
  \end{subfigure}
  \caption{Oracle's performance and optimal stopping time} \label{fig:heat_map_oracle}
    
  \footnotesize \parbox{0.95\linewidth}{\textit{Note.} Illustration of the oracle’s average value and stopping point across 30 sample paths over a grid of per-sample acquisition costs \(c_{\text{acq}}\), training costs \(c_{\text{train}}\), and discount factors \(\beta\), with \(c_s=0\), using LR (first column) and LightGBM (second column). Cell shading indicates average performance, while the value inside each cell reports the stopping point.}
\end{figure}
\begin{figure}[h!]
  \centering
  \begin{subfigure}[b]{0.3\textwidth}
  \hspace{-15pt}
    \includegraphics[width=1\linewidth]{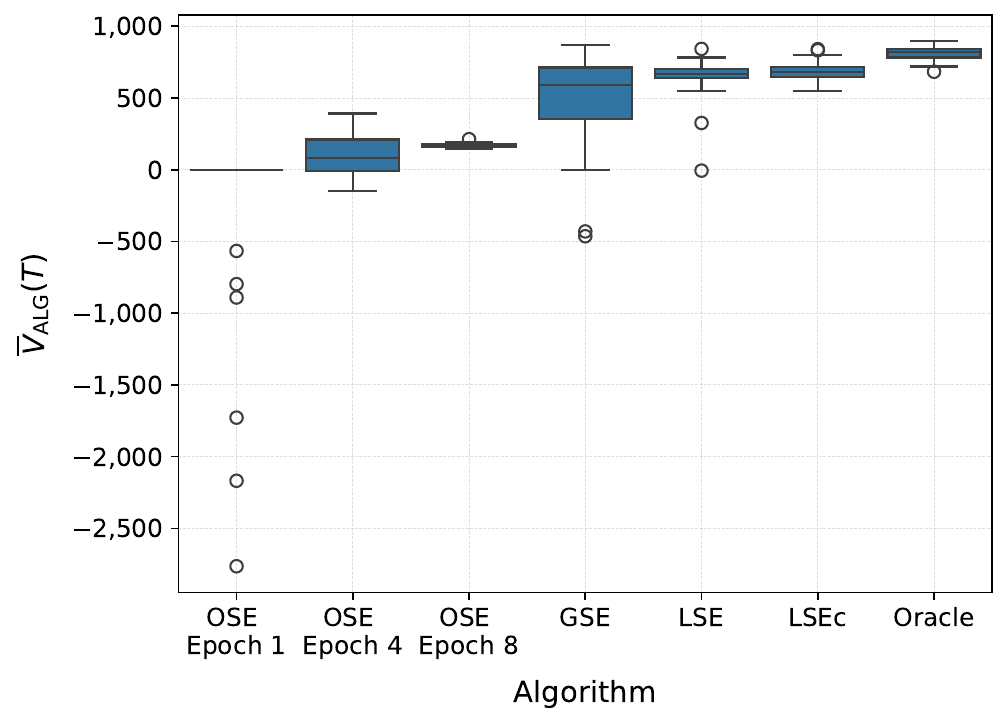}
    \caption{$c_{\text{train}}=0$}
    \label{fig:sub1} 
  \end{subfigure} 
  \hfill 
  \begin{subfigure}[b]{0.3\textwidth}
  \hspace{-15pt}
    \includegraphics[width=1\linewidth]{Figures/LightGBM_Box_plot_High_gap_E2}
    \caption{$c_{\text{train}}=0.005$}
    \label{fig:sub1}
  \end{subfigure} 
  \hfill
  \begin{subfigure}[b]{0.3\textwidth}
  \hspace{-15pt}
    \includegraphics[width=1\linewidth]{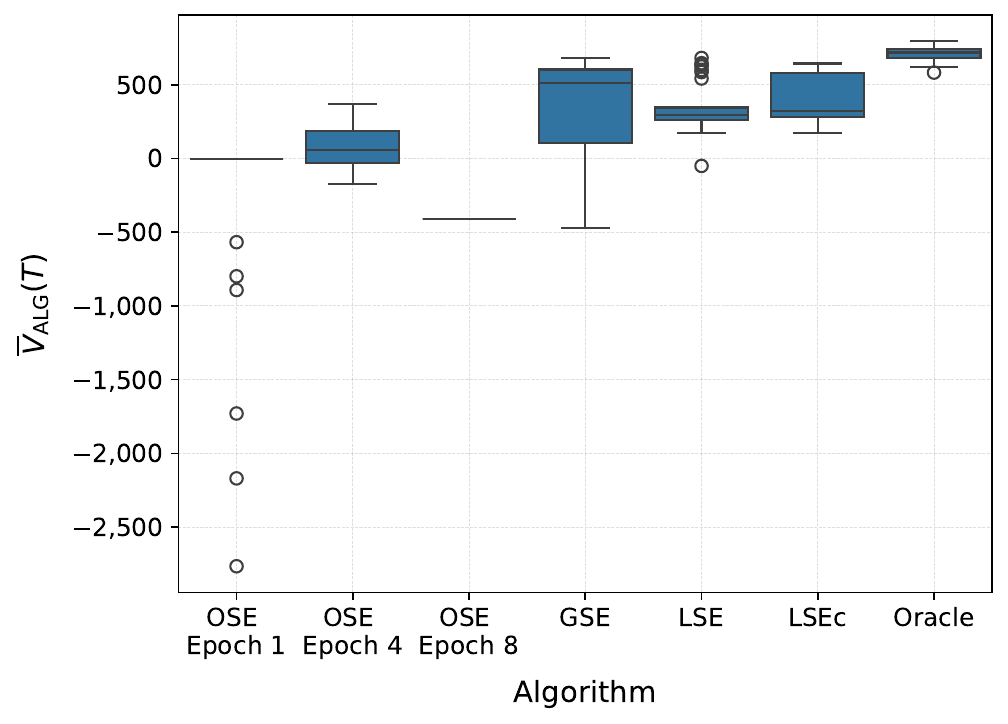}
    \caption{$c_{\text{train}}=0.01$}
    \label{fig:sub1}
  \end{subfigure}
  
  \begin{subfigure}[b]{0.3\textwidth}
  \hspace{-15pt}
    \includegraphics[width=1\linewidth]{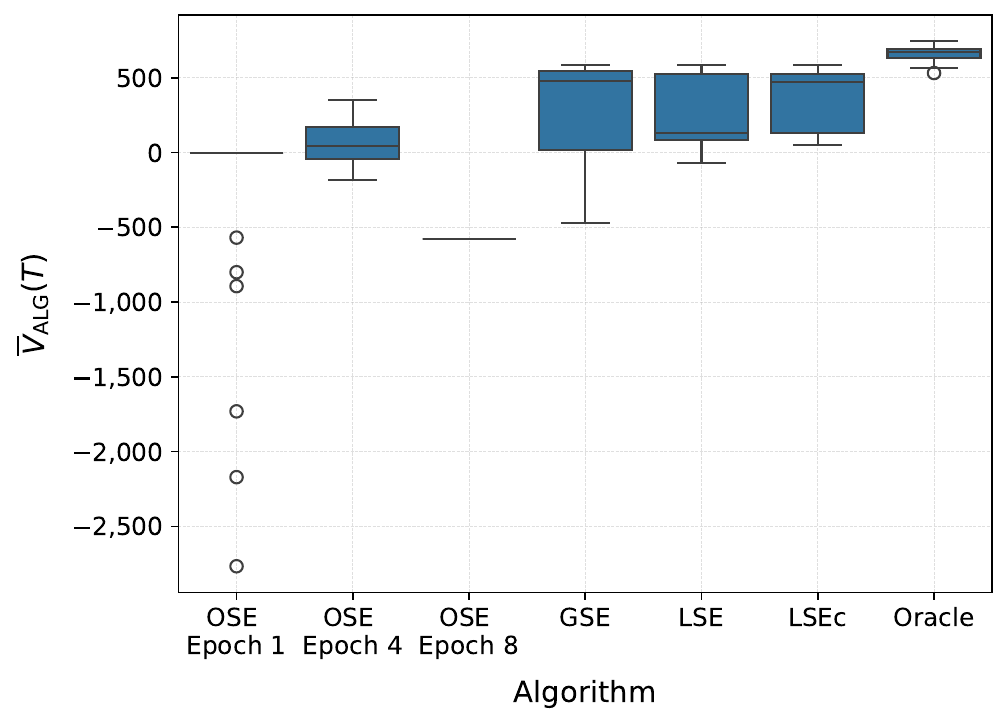}
    \caption{$c_{\text{train}}=0.015$}
    \label{fig:sub1}
  \end{subfigure}
  \vspace{0.25cm}
  \hfill   
  \begin{subfigure}[b]{0.3\textwidth}
  \hspace{-15pt}
    \includegraphics[width=1\linewidth]{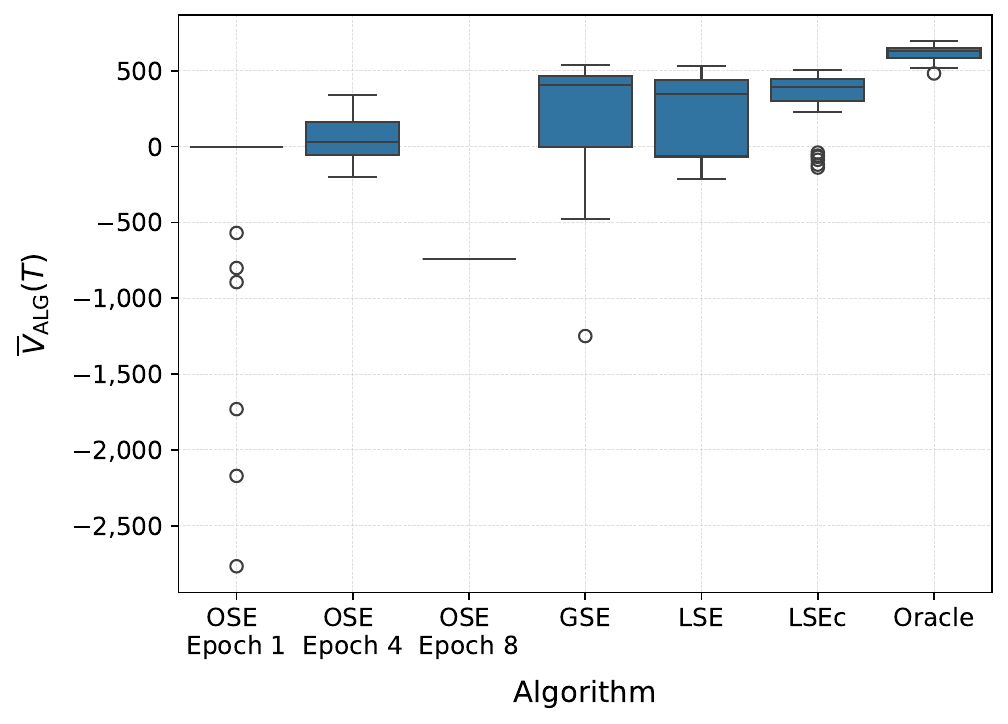}
    \caption{$c_{\text{train}}=0.02$}
    \label{fig:sub1}
  \end{subfigure}
  \hfill 
  \begin{subfigure}[b]{0.3\textwidth}
  \hspace{-15pt}
    \includegraphics[width=1\linewidth]{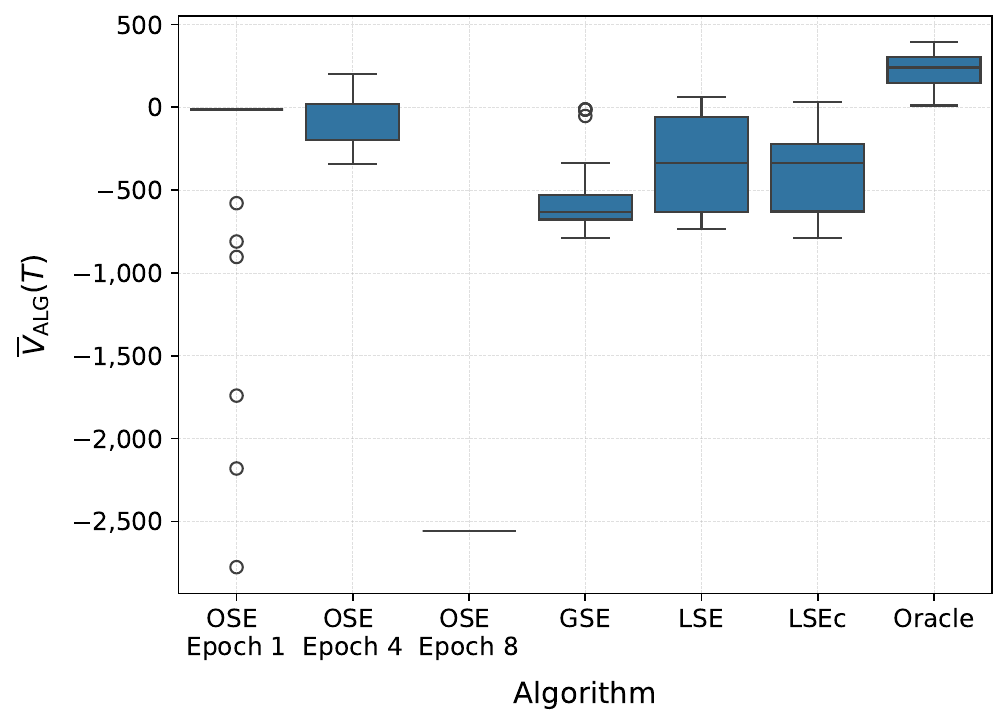}
    \caption{$c_{\text{train}}=0.075$}
    \label{fig:sub1}
  \end{subfigure}
  \vspace{0.2cm}
  \caption{E.2 for different per-sample training costs $c_{\text{train}}$} \label{fig:E_2_robust}
  \medskip
    
  \footnotesize \parbox{0.95\linewidth}{\textit{Note.} This figure displays the performance of our algorithms across the 30 sample paths for experiment E.2, when varying the per-sample training cost $c_{\text{train}}$.}
\end{figure}

\paragraph{Algorithm value and oracle stopping time.}
We evaluate algorithm performance using the notion of {empirical value} of an algorithm, denoted
\(\overline{V}_{\textsc{alg}}(T)\).
Let denote $\tilde{t} \in \mathcal{T}_D$ the epoch at which the algorithm decides to switch or discard. If the algorithm switches at epoch \(k\), then $\tilde{t}=t_k$ and 
\(\overline{V}_{\textsc{alg}}(T)=\overline{V}_{\text{switch}}(\tilde{t})\).
Otherwise, if it discards the challenger at epoch \(k\), then $\tilde{t}=t_k$ and 
\(\overline{V}_{\textsc{alg}}(T)= -\sum_{\tau \le \tilde{t}}\beta^{\tau} C_{\mathrm{pre}}(\tau)\). 
This definition also covers the oracle: if it discards, it does so before starting the sequential process, yielding zero performance.
In the EC, we study the oracle's optimal switching time and performance across six values of the per-sample acquisition cost, six values of the per-sample training cost, and two discount-factor values, for both the LR and the LightGBM models. 
The full oracle grid is reported in Figure~\ref{fig:heat_map_oracle} of the EC. Consistent with the
structure of Section~\ref{sec:theory}, acquisition costs do not affect the optimal stopping point when
pre- and post-decision acquisition costs are identical; higher training costs and lower
discount factors move the optimal switching point earlier; and sufficiently high costs make
discarding optimal. We obtain E.1 and E.2 by selecting an early and a late switching point
from the grid of Figure~\ref{fig:heat_map_oracle}. Specifically, over our parameter grid, the earliest average switching point occurs around the fourth epoch using LR with \(c_{\text{train}}=0.075\) and \(\beta=0.95\), independently of \(c_{\text{acq}}\). Using LightGBM, the latest average switching point occurs around the sixth epoch for multiple parameter combinations. We fix \(\beta=0.95\) and \(c_{\text{acq}}=0.0025\), set \(c_{\text{train}}=0.075\) (early) and \(0.005\) (late), and show in Figure~\ref{fig:E_2_robust} that results are qualitatively unchanged under alternative specifications.

\paragraph{Algorithm hyperparameter selection.}
Here, we briefly explain how we select each algorithm's hyperparameters (we defer the detailed discussion to Section~\ref{subsec:sensitivity} of the EC). OSE is not an adaptive rule, so we report three benchmark
evaluation epochs: the first epoch corresponding to purchasing only the first batch; the final epoch corresponding to purchasing the entire dataset; and the fourth epoch because it yields the best OSE performance in the first core scenario (E.1).
For GSE and LSEc, the key tuning parameter is the confidence parameter $\gamma$. 
In GSE, larger $\gamma$ widens
the confidence interval around the current gap estimate, making the rule more conservative
and delaying terminal decisions. 
In LSEc, larger $\gamma$ enters the confidence adjustment in
the local slope estimate, making future gap projections more optimistic and also delaying
stopping. We choose one value of $\gamma$ for each procedure using the sensitivity analysis and hyperparameter tuning procedure outlined in
Section~\ref{subsec:sensitivity}, and then hold it fixed throughout the main analysis. We observe that GSE is sensitive to this choice and no value dominates across the grid; in contrast, LSEc is much less
sensitive, which allows us to use a value from the stable region of its sensitivity curve without incurring much loss. 

\subsection{Results}
\label{sec:result_core_scenario}
\begin{figure}[ht]
  \centering
  \begin{subfigure}[b]{0.475\textwidth}
    \includegraphics[width=1\textwidth]{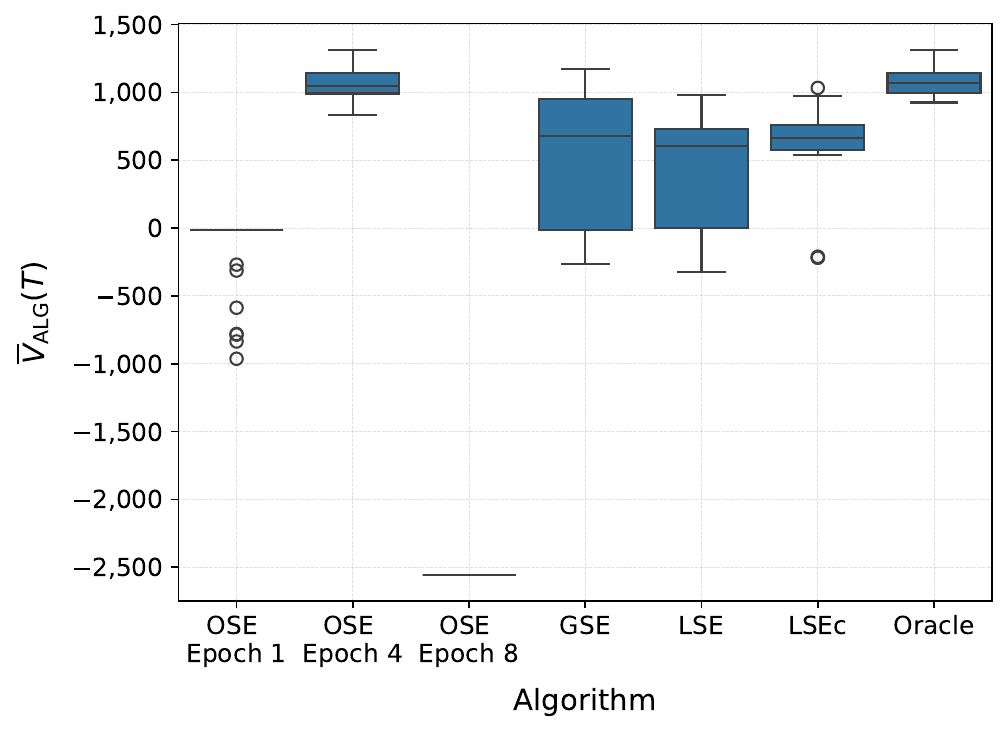}
    \caption{Scenario E.1 (early)}
  \end{subfigure}
    \hfill
    \begin{subfigure}[b]{0.475\textwidth}
    \includegraphics[width=1\textwidth]{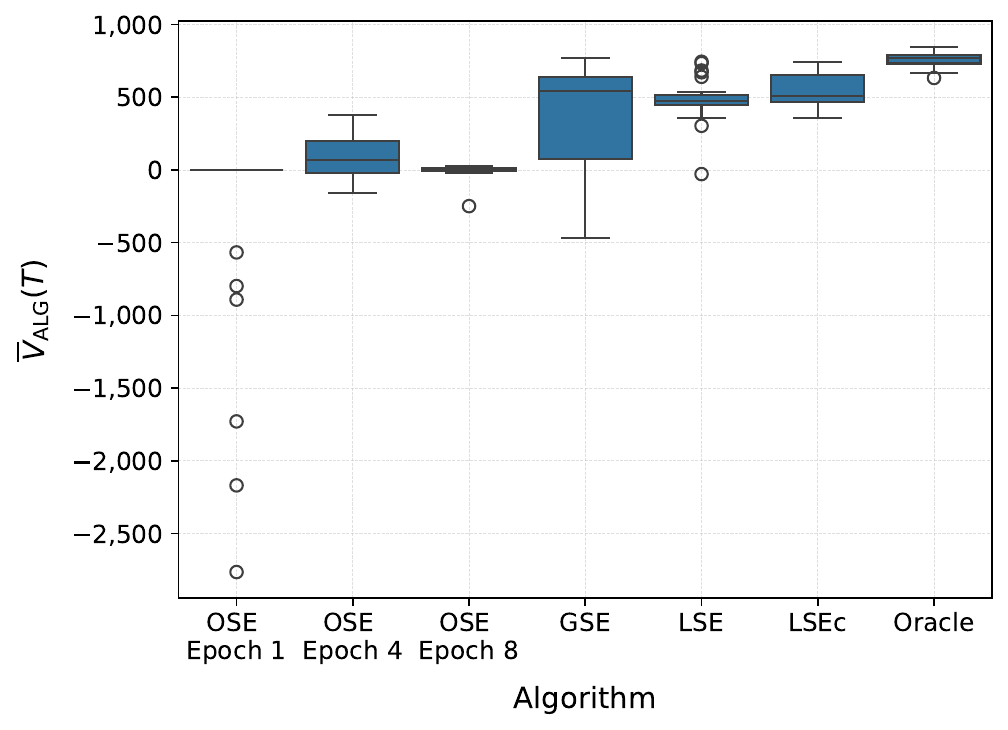}
    \caption{Scenario E.2 (late)}
  \end{subfigure}  
  \caption{Algorithm terminal values}
  \label{fig:E1_E2}
  \medskip
    
  \footnotesize \parbox{0.95\linewidth}{\textit{Note.} This figure displays the performance of our algorithms across the 30 sample paths for the early- and late-switch scenarios.}
\end{figure}

\begin{figure}[ht] \centering
\begin{minipage}{\textwidth}
  \centering
\begin{subfigure}[b]{0.32\textwidth}
    \includegraphics[width=1\textwidth]{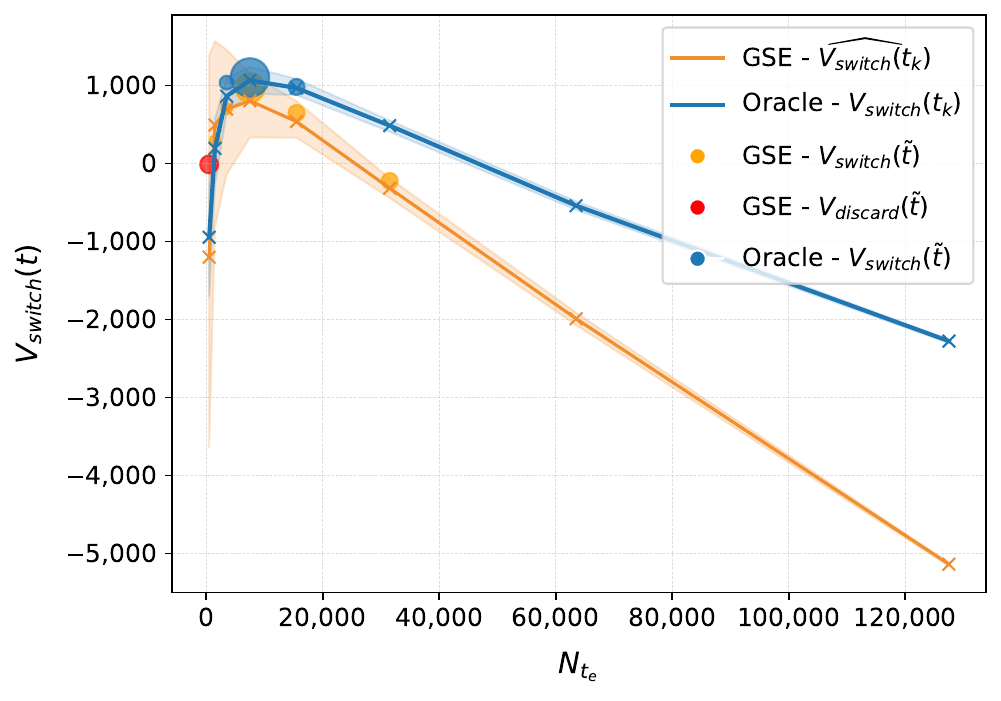}
    \caption{Scenario E.1 (early): GSE}
    \label{fig:sub1}
  \end{subfigure}
  \hfill
  \begin{subfigure}[b]{0.32\textwidth}
    \includegraphics[width=1\textwidth]{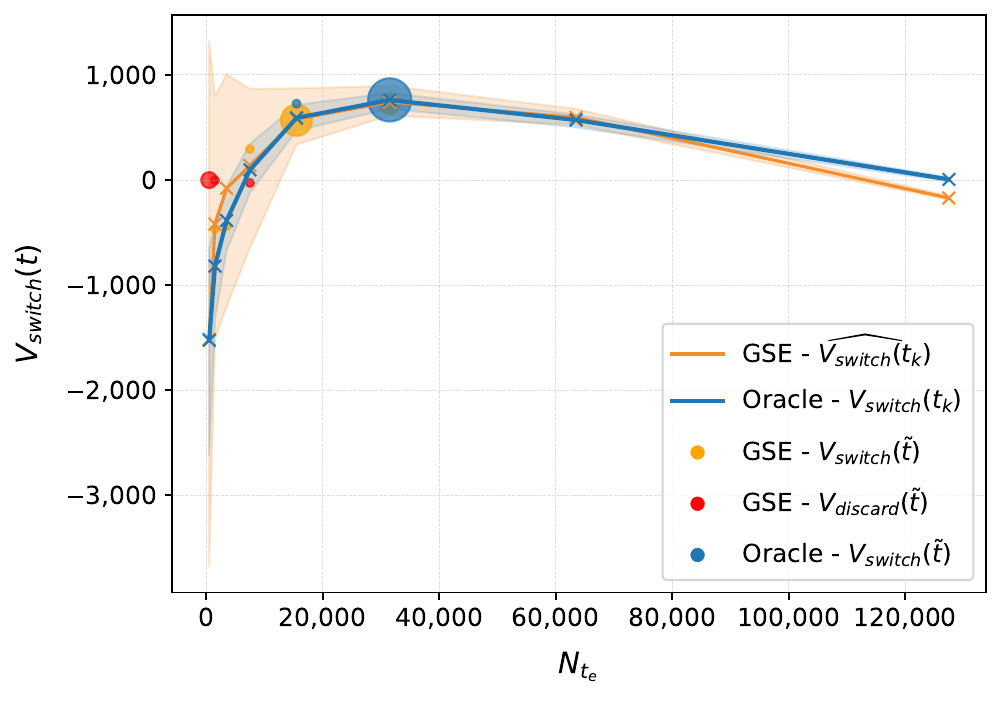}
    \caption{Scenario E.2 (late): GSE}
    \label{fig:sub1}
  \end{subfigure}
  \hfill
    \begin{subfigure}[b]{0.32\textwidth}
    \includegraphics[width=1\textwidth]{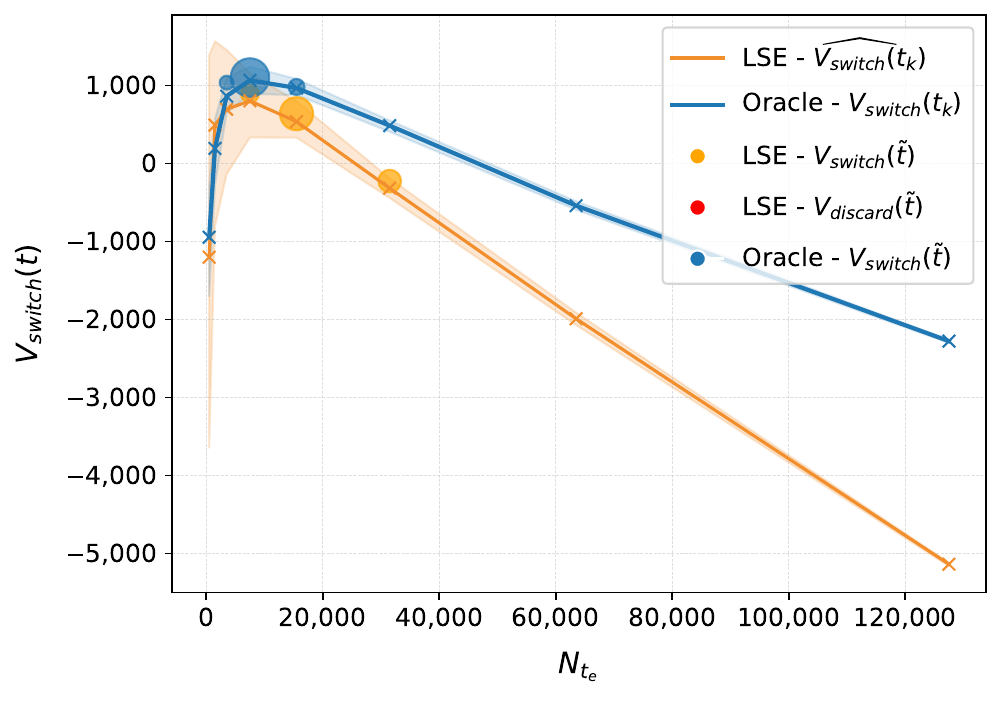}
    \caption{Scenario E.1 (early): LSE}
    \label{fig:sub1}
  \end{subfigure} 
  \hfill
  \begin{subfigure}[b]{0.32\textwidth}
    \includegraphics[width=1\textwidth]{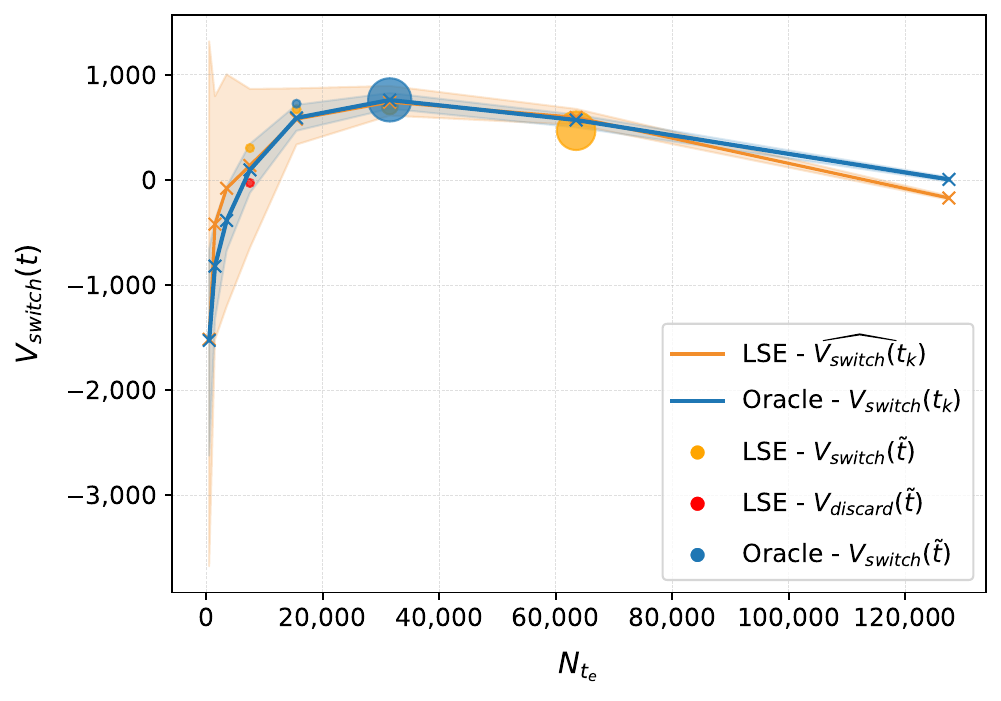}
    \caption{Scenario E.2 (late): LSE}
    \label{fig:sub1}
  \end{subfigure} 
  \hfill
  \begin{subfigure}[b]{0.32\textwidth}
    \includegraphics[width=1\textwidth]{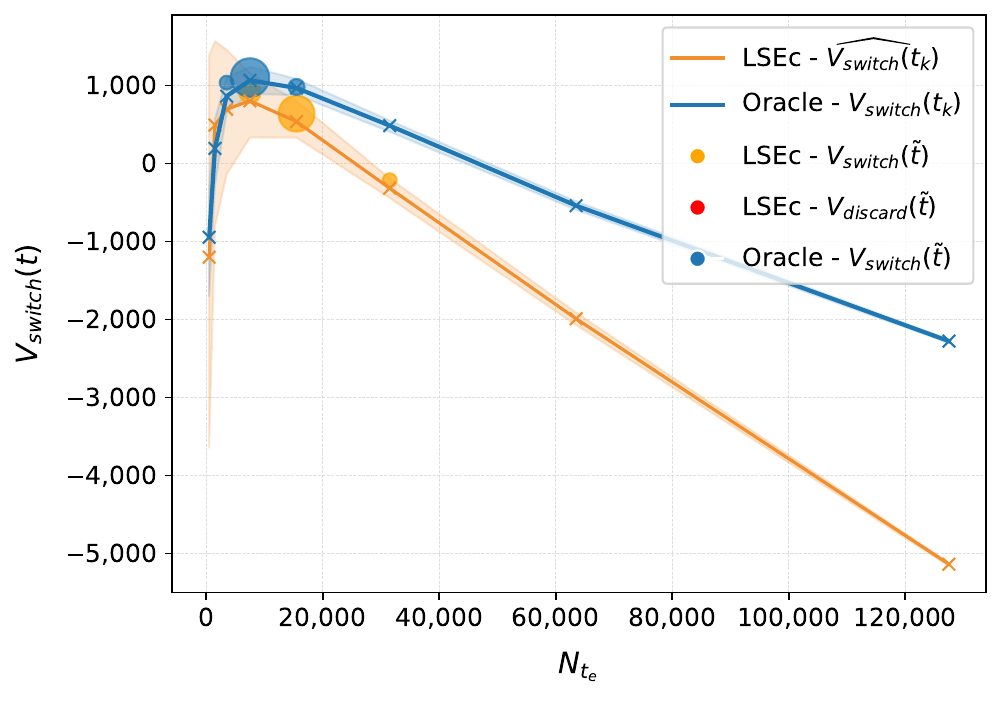}
    \caption{Scenario E.1 (early): LSEc}
    \label{fig:sub1}
  \end{subfigure}
  \hfill
  \begin{subfigure}[b]{0.32\textwidth}
    \includegraphics[width=1\textwidth]{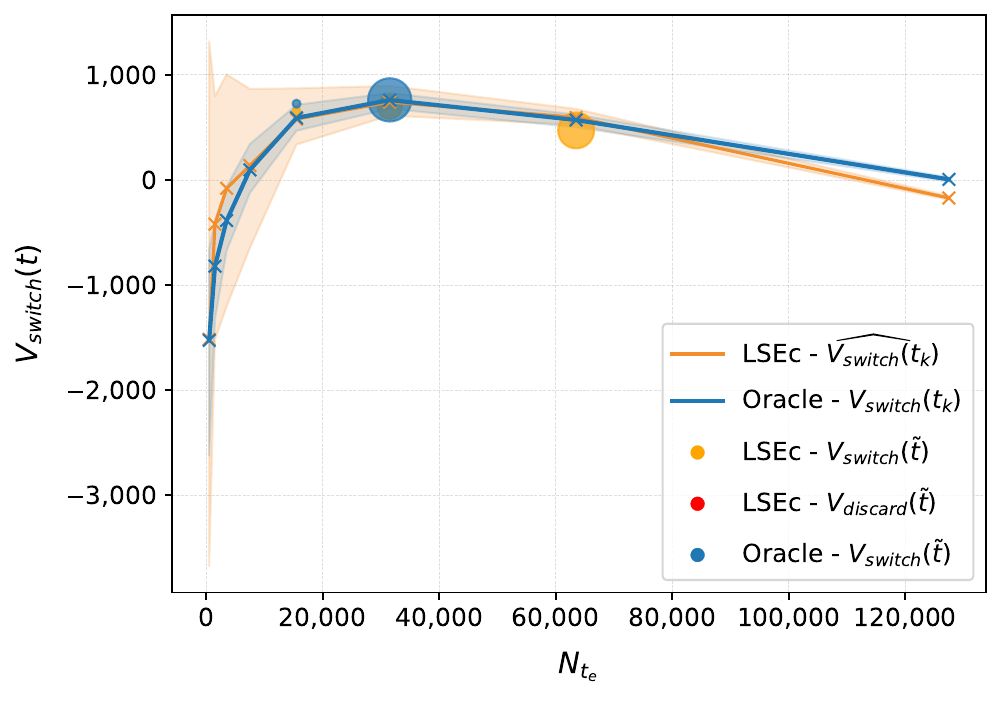}
    \caption{Scenario E.2 (late): LSEc}
    \label{fig:sub1}
  \end{subfigure} 
\end{minipage}
  \caption{Algorithm decisions and terminal values compared with oracle across 30 sample paths}
  \label{fig:E1_E2_utility}
  \medskip
    
  \footnotesize \parbox{0.95\linewidth}{\textit{Note.} Solid lines report average switching values at each \(t_k\) (shaded areas indicate 90\% confidence intervals). Blue and orange dots denote oracle and algorithm performance at the switching epoch \(\tilde{t}\), respectively, while red dots indicate discard decisions. Dot sizes are proportional to the frequency of the corresponding decision across sample paths. Dot heights represent the average of the switching or discarding value at the corresponding epoch. Each panel corresponds to a different algorithm.}
  
\end{figure}

\begin{figure}[htb]
\centering 
\begin{minipage}{0.8\textwidth}
  \centering
  \begin{subfigure}[b]{0.45\textwidth}
    \includegraphics[width=1\textwidth]{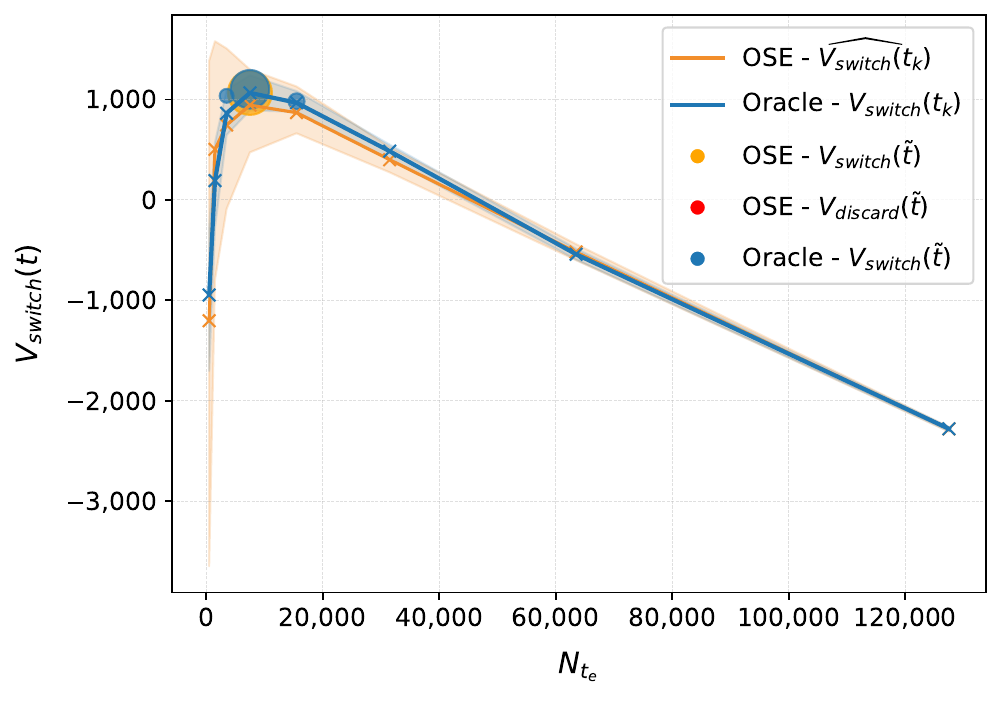}
    \label{fig:sub1}
  \end{subfigure}
  \hfill
    \begin{subfigure}[b]{0.45\textwidth}
    \includegraphics[width=1\textwidth]{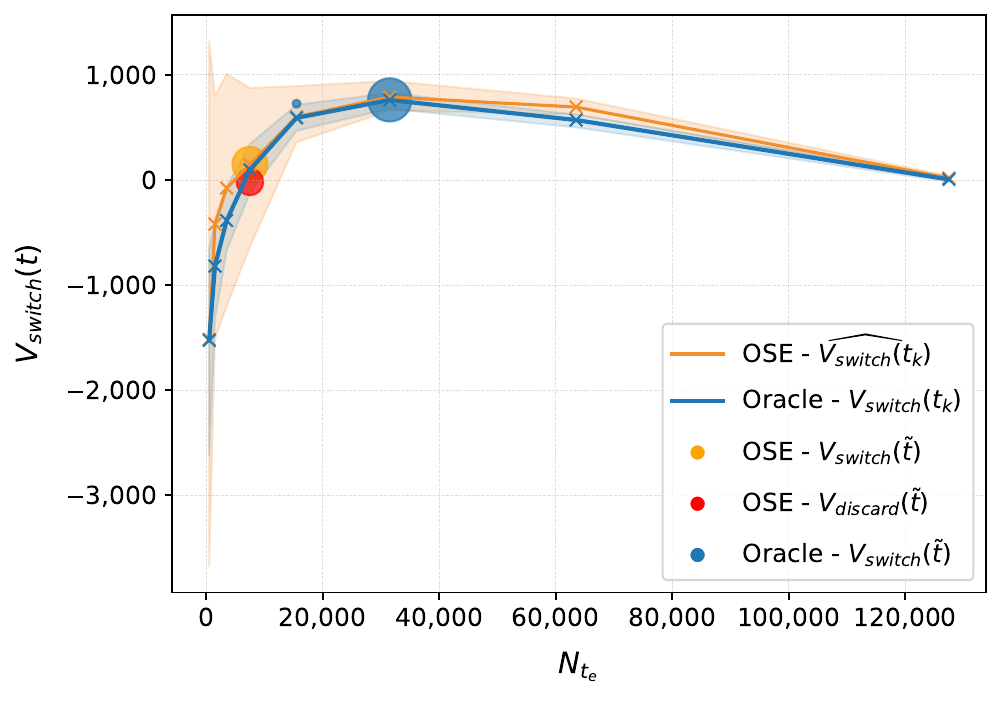}
    \label{fig:sub1}
  \end{subfigure}
\end{minipage}
  \vspace{-0.5cm}
  \caption{Early switch vs late switch - OSE algorithm} \label{fig:E_1_OSE}
    
  \footnotesize \parbox{0.95\linewidth}{\textit{Note.} The figure shows the utility (switching value) and decisions of OSE algorithm and the oracle across 30 sample paths for experiments E.1 and E.2. Solid lines report average switching values at each \(t_k\), with shaded areas indicating 90\% confidence intervals. Blue and orange dots denote oracle and algorithm performance at the switching epoch \(\tilde{t}\), respectively, while red dots indicate discard decisions. Dot sizes are proportional to the frequency of the corresponding decision across sample paths. Dot heights represent the average of the switching or discarding value at the corresponding epoch.}

\end{figure}

We now analyze results for the two core scenarios. Figure~\ref{fig:E1_E2} summarizes algorithm performance across the 30 sample paths, with the corresponding utility curves in Figures \ref{fig:E1_E2_utility} and \ref{fig:E_1_OSE}. 
These utility curves report the switching value at each epoch $t_k$.
We distinguish between the oracle’s switching value (blue), $\overline{V}_{\text{switch}}(t_k)$, and the algorithm’s empirically estimated switching value (orange), $\widehat{V}_{\text{switch}}(t_k)$.
The blue curve plots the oracle’s switching value, $\overline{V}_{\text{switch}}(t_k)$, computed using future data and assuming training occurs only at the candidate epoch (so no value is lost to excess training).
The orange curve plots the empirically estimated switching value available to the algorithms, $\widehat{V}_{\text{switch}}(t_k)$, constructed along the realized path using the holdout set.
Consequently, the two curves differ for two reasons: algorithms (i) estimate performance gaps from the holdout set rather than future data and (ii) typically incur cumulative training costs prior to stopping. Next, we summarize the main takeaways from Figures~\ref{fig:E1_E2} and \ref{fig:E1_E2_utility}.
\vspace{-0.25cm}

\paragraph{Performance of GSE.}
GSE is highly volatile across the 30 sample paths in both E.1 and E.2. A first source of volatility is premature discarding: GSE occasionally fails to switch and instead discards the challenger at early stages, as indicated by the red dots in the utility-curve plots. In the first three epochs, the estimated switching values $\widehat{V}_{\text{switch}}(t)$ (orange curves), on which the algorithm bases its decisions, are highly noisy and frequently negative. 
We observe two behaviors. When the required confidence level is low, GSE can be too aggressive: because it is not forward-looking and relies on the current estimate, early noise may lead it to conclude that the challenger underperforms the incumbent and discard it prematurely. When the required confidence level is high, GSE can instead be too conservative: even when it does not discard, it often remains uncertain about whether switching dominates continuing, so early-epoch noise causes switching at different epochs across sample paths, further increasing dispersion in realized values.

\paragraph{Performance of OSE.}
Setting the evaluation epoch to the fourth epoch yields strong performance for OSE in E.1, but performance deteriorates markedly in E.2. This is expected: the oracle stops at the fourth epoch in E.1 but at the sixth epoch in E.2, so fixing OSE to evaluate at epoch four forces it to act too early in the latter scenario. More broadly, purchasing only the initial batch or the entire dataset upfront is unprofitable in both scenarios. Together, these findings show that OSE is highly sensitive to the evaluation epoch and that no single fixed epoch performs well when the optimal stopping point varies across settings.

\paragraph{Performance of LSE.}
LSE behaves differently across the two scenarios. In E.1, its performance is also volatile, but for a different reason than GSE’s. Because LSE is forward-looking, it forms expectations about future gaps based on the evolution of past value estimates. This optimism can cause LSE to switch later than is optimal; when the switching value declines sharply after its peak, switching one or two epochs too late produces a substantial opportunity loss, explaining the volatility in E.1. In E.2, by contrast, LSE typically stops only one epoch after the peak; since values decline only mildly thereafter, realized performance remains tightly concentrated and close to the oracle’s. Thus, the difference in LSE performance is primarily driven by utility-curve shape: the much higher training cost in E.1 causes switching values to fall steeply after their peak, amplifying the penalty for even slight delays.

\paragraph{Performance of LSEc.}
LSEc stands out as the most stable, delivering performance that remains consistently close to the oracle across sample paths in both E.1 and E.2. Although closely related to LSE, LSEc estimates its slope using only the most recent finite difference and incorporates a confidence adjustment, whereas in our empirical analysis LSE uses a window of $w=3$ and thus relies on the two preceding epochs. As discussed in Section~4.3, these design choices affect how optimistic each algorithm is when extrapolating future gaps: using three points can amplify optimism when gaps increase sharply, while confidence adjustment reinforces optimism when recent estimates are noisy. This pattern aligns with the observed behavior. In E.1, the learning curve exhibits steep early improvements (Figure~\ref{fig:learning_curve}), so LSE’s three-point slope becomes inflated and it tends to switch later than LSEc. In E.2, the learning curve is less steep and early estimates are noisier, making confidence-adjusted LSEc slightly more optimistic; e.g., LSEc defers switching until epochs five, six, or seven, while LSE occasionally switches or even discards at epoch four.

\paragraph{Model choice.}
Higher model complexity does not automatically translate into higher profitability. In particular, LR in E.1 achieves higher value than LightGBM in E.2 for most algorithms, including the oracle, and this pattern holds throughout our empirical analysis. The reason is straightforward: LR’s learning curve is consistently steeper and lies above LightGBM’s at all decision epochs (Figure~\ref{fig:learning_curve}). As a result, for a given economic structure, LR dominates in economic value.

\section{Conclusion} \label{sec:conclusion}
We study the model-switching problem faced by organizations that already operate an incumbent predictive model when a new data source becomes available. Because the historical training data lack the new features, the challenger must be trained on a small but growing full-feature dataset. We develop a framework linking learning-curve dynamics, model-switching economics, and statistical uncertainty; characterize oracle timing; provide regret guarantees for implementable procedures; and validate the framework in a credit-scoring application.

\textit{Managerial implications.} Our results show that evaluation timing should be treated as a model-governance decision. New data sources should not be adopted simply because they improve eventual predictive performance, and they should not be rejected because early challenger estimates are noisy. In our framework, the learning curve determines when experimentation should stop, whereas costs determine whether switching is worthwhile. Organizations should therefore track both the current performance gap between the challenger and the incumbent, and the rate at which this gap is improving.
A second implication concerns the design of evaluation procedures. One-shot evaluations are simple, but brittle: they work well only when the chosen evaluation date happens to match the unknown optimal scale. Greedy sequential evaluation reacts to current evidence, but can discard the challenger too early when early estimates are noisy. Look-ahead evaluation offers a middle ground: it uses local learning-curve trends to preserve the option value of continued learning, while stopping once projected improvement no longer justifies additional delay and cost. Operationally, this means that model-lifecycle policies should specify the evaluation cadence, the level of statistical confidence required before switching or discarding, and the degree of optimism allowed about future learning.
Finally, the empirical results caution against equating model complexity with economic value. In the credit-scoring study, the simpler model can generate higher value when its learning curve improves faster and reaches useful performance earlier. The relevant comparison is net value over the deployment horizon after data-collection, validation, retraining, and switching costs are included.

\paragraph{Future research.} Future research could extend the framework to multiple competing challengers, retrained incumbents, or richer forms of distributional shift and model misspecification. Another direction is to integrate risk and regulatory constraints directly into the switching rule, especially in high-stakes settings such as finance where new data sources can improve decisions but also introduce governance and compliance costs. 
Finally, although our empirical analysis centers on finance, the proposed framework generalizes to other domains, including healthcare. Consider a hospital deploying a ML model for disease diagnosis: the incumbent relies on demographic and laboratory data, whereas the challenger augments these with genetic markers and real-time monitoring information. While the expanded feature set may enhance diagnostic accuracy, it also entails additional expenditures related to genetic testing, data integration, and infrastructure investment.

\bibliographystyle{informs2014} 
\bibliography{bib-ms} 

@article{Babina2025,
  title={Customer data access and fintech entry: Early evidence from open banking},
  author={Babina, Tania and Bahaj, Saleem and Buchak, Greg and De Marco, Filippo and Foulis, Angus and Gornall, Will and Mazzola, Francesco and Yu, Tong},
  journal={Journal of Financial Economics},
  volume={169},
  pages={103950},
  year={2025},
  publisher={Elsevier}
}

@article{Berg2020,
	author = {Berg, Tobias and Burg, Valentin and Gombović, Ana and Puri, Manju},
	title = {On the rise of FinTechs: Credit scoring using digital footprints},
	journal = {Review of Financial Studies},
	year = {2020},
	volume={33},
	number={7},
	pages={2845-2897}
}

@article{Bonelli2025,
    author = {Bonelli, Maxime},
    title = {Data-Driven Investors},
    journal = {The Review of Financial Studies},
    volume = {39},
    number = {7},
    pages = {1909-1969},
    year = {2026},
    month = {07},
}

@article{Bryan2024,
        Author = {Bryan, Gharad and Karlan, Dean and Osman, Adam},
        Title = {Big Loans to Small Businesses: Predicting Winners and Losers in an Entrepreneurial Lending Experiment},
        Journal = {American Economic Review},
        Volume = {114},
        Number = {9},
        Year = {2024},
        Month = {September},
        Pages = {2825–60},
}

@article{Cao2024,
        title = {From Man vs. Machine to Man + Machine: The art and {AI} of stock analyses},
        journal = {Journal of Financial Economics},
        volume = {160},
        pages = {103910},
        year = {2024},
        author = {Sean Cao and Wei Jiang and Junbo Wang and Baozhong Yang},
}

@article{ChenMS2024,
        author = {Chen, Luyang and Pelger, Markus and Zhu, Jason},
        title = {Deep Learning in Asset Pricing},
        journal = {Management Science},
        volume = {70},
        number = {2},
        pages = {714-750},
        year = {2024}
}

@article{Chi2025,
        author = {Chi, Feng and Hwang, Byoung-Hyoun and Zheng, Yaping},
        title = {The Use and Usefulness of Big Data in Finance: Evidence from Financial Analysts},
        journal = {Management Science},
        volume = {71},
        number = {6},
        pages = {4599-4621},
        year = {2025}
}

@article{Dessaint2024,
        author = {Dessaint, Olivier and Foucault, Thierry and Fresard, Laurent},
        title = {Does Alternative Data Improve Financial Forecasting? {T}he Horizon Effect},
        journal = {Journal of Finance},
        volume = {79},
        number = {3},
        pages = {2237-2287},
        year = {2024}
}

@article{Djeundje2021,
        title = {Enhancing credit scoring with alternative data},
        journal = {Expert Systems with Applications},
        volume = {163},
        pages = {113766},
        year = {2021},
        author = {Viani B. Djeundje and Jonathan Crook and Raffaella Calabrese and Mona Hamid}
}

@techreport{Foucault2025,
        title={Artificial Intelligence in Finance},
        author={Foucault, Thierry and Gambacorta, Leonardo and Jiang, Wei and Vives, Xavier},
        institution = "Centre for Economic Policy Research (CEPR)",
        type = "Report",
        year={2025}
}

@article{Gambacorta2023,
        author = {Gambacorta, Leonardo and Huang, Yiping and Li, Zhenhua and Qiu, Han and Chen, Shu},
        title = {Data versus Collateral},
        journal = {Review of Finance},
        volume = {27},
        number = {2},
        pages = {369-398},
        year = {2022}
}

@article{Gambacorta2024,
        title = {How do machine learning and non-traditional data affect credit scoring? {New} evidence from a Chinese fintech firm},
        journal = {Journal of Financial Stability},
        volume = {73},
        pages = {101284},
        year = {2024},
        author = {Leonardo Gambacorta and Yiping Huang and Han Qiu and Jingyi Wang},
}

@article{Green2019,
        title = {Crowdsourced employer reviews and stock returns},
        journal = {Journal of Financial Economics},
        volume = {134},
        number = {1},
        pages = {236-251},
        year = {2019},
        author = {T. Clifton Green and Ruoyan Huang and Quan Wen and Dexin Zhou},
}

@article{GuKellyXiu2020,
    author = {Gu, Shihao and Kelly, Bryan and Xiu, Dacheng},
    title = {Empirical Asset Pricing via Machine Learning},
    journal = {Review of Financial Studies},
    volume = {33},
    number = {5},
    pages = {2223-2273},
    year = {2020}
}

@article{He2023,
        title = {Open banking: Credit market competition when borrowers own the data},
        author = {Zhiguo He and Jing Huang and Jidong Zhou},
        journal = {Journal of Financial Economics},
        volume = {147},
        number = {2},
        pages = {449-474},
        year = {2023}
}

@article{ke2017lightgbm,
  title={{LightGBM: A highly efficient gradient boosting decision tree}},
  author={Ke, Guolin and Meng, Qi and Finley, Thomas and Wang, Taifeng and Chen, Wei and Ma, Weidong and Ye, Qiwei and Liu, Tie-Yan},
  journal={{Advances in Neural Information Processing Systems}},
  volume={30},
  year={2017}
}

@article{Iyer2016,
        author = {Iyer, Rajkamal and Khwaja, Asim Ijaz and Luttmer, Erzo F. P. and Shue, Kelly},
        title = {Screening Peers Softly: Inferring the Quality of Small Borrowers},
        journal = {Management Science},
        volume = {62},
        number = {6},
        pages = {1554-1577},
        year = {2016}
}

@article{kelly2022financial,
  title        = {Financial Machine Learning},
  author       = {Kelly, Bryan and Xiu, Dacheng},
  journal      = {Foundations and Trends in Finance},
  volume       = {13},
  number       = {3-4},
  pages        = {205--363},
  year         = {2022},
  publisher    = {Now Publishers}
}

@article{Kelly2024,
        author = {Kelly, Bryan and Malamud, Semyon and Zhou, Kankying},
        title = {The Virtue of Complexity in Return Prediction},
        journal = {Journal of Finance},
        volume = {79},
        number = {1},
        pages = {459-503},
        year = {2024}
}

@article{lessmann2015benchmarking,
        title={{Benchmarking State-of-the-Art Classification Algorithms for Credit Scoring: An Update of Research}},
        author={Lessmann, Stefan and Baesens, Bart and Seow, Hsin-Vonn and Thomas, Lyn C},
        journal={European Journal of Operational Research},
        volume={247},
        number={1},
        pages={124--136},
        year={2015}
        }

@article{MAHADEVAN2024,
        title = {Cost-aware retraining for machine learning},
        journal = {Knowledge-Based Systems},
        volume = {293},
        pages = {111610},
        year = {2024},
        author = {Ananth Mahadevan and Michael Mathioudakis}
}

@article{Oskarsdottir_Baesens2019,
        title = {The value of big data for credit scoring: Enhancing financial inclusion using mobile phone data and social network analytics},
        journal = {Applied Soft Computing},
        volume = {74},
        pages = {26-39},
        year = {2019},
        author = {María Oskarsdottir and Cristian Bravo and Carlos Sarraute and Jan Vanthienen and Bart Baesens},
}

@article{ZLIOBAITE2015,
        title = {Towards cost-sensitive adaptation: When is it worth updating your predictive model?},
        journal = {Neurocomputing},
        volume = {150},
        pages = {240-249},
        year = {2015},
        author = {Indrė Žliobaitė and Marcin Budka and Frederic Stahl},
}

@book{mohri2018foundations,
  title={Foundations of machine learning},
  author={Mohri, Mehryar and Rostamizadeh, Afshin and Talwalkar, Ameet},
  year={2018},
  publisher={MIT press}
}

@article{viering2022shapelearningcurvesreview,
        title={The Shape of Learning Curves: {A} Review}, 
        author={Tom Viering and Marco Loog},
        year = {2023},
        volume = {45},
        number = {6},
        journal = {IEEE Trans. Pattern Anal. Mach. Intell.},
        pages = {7799–7819}
}

@article{Mohr_2024,
        title={Learning curves for decision making in supervised machine learning: a survey},
        volume={113},
        number={11–12},
        journal={Machine Learning},
        publisher={Springer Science and Business Media LLC},
        author={Mohr, Felix and van Rijn, Jan N.},
        year={2024},
        month=dec, pages={8371–8425}
}

@article{Ravina2025,
        title={Love \& Loans: {T}he Effect of Beauty and Personal Characteristics in Credit Markets},
        author={Ravina, Enrichetta},
        journal={Management Science,},
        year={2025},
        volume={forthcoming}
}

@inproceedings{10.1145/312129.312188,
        author = {Provost, Foster and Jensen, David and Oates, Tim},
        title = {Efficient progressive sampling},
        year = {1999},
        publisher = {Association for Computing Machinery}, address = {New York, NY, USA},
        booktitle = {Proceedings of the Fifth ACM SIGKDD International Conference on Knowledge Discovery and Data Mining},
        pages = {23–32},    
        location = {San Diego, California, USA}, series = {KDD '99}
}

@InProceedings{pmlr-vR2-frey99a,
        title = {Modeling decision tree performance with the power law},
        author = {Frey, Lewis J. and Fisher, Douglas H.},
        booktitle = {Proceedings of the Seventh International Workshop on Artificial Intelligence and Statistics},
        year = {1999},
        editor = {Heckerman, David and Whittaker, Joe},
        volume = {R2},
        series = {Proceedings of Machine Learning Research},
        publisher = {PMLR}
}

@INPROCEEDINGS{4476671,
        author={Last, Mark},
        booktitle={Seventh IEEE International Conference on Data Mining Workshops (ICDMW 2007)}, 
        title={Predicting and Optimizing Classifier Utility with the Power Law}, 
        year={2007},
        volume={},
        number={},
        pages={219-224}
}

@article{10.1007/s10618-007-0082-x,
        author = {Weiss, Gary M. and Tian, Ye},
        title = {Maximizing classifier utility when there are data acquisition and modeling costs},
        year = {2008},
        publisher = {Kluwer Academic Publishers},
        address = {USA},
        volume = {17},
        number = {2},
        journal = {Data Min. Knowl. Discov.},
        month = oct,
        pages = {253–282}
}

@inproceedings{10.1145/1557019.1557076,
        author = {Last, Mark},
        title = {Improving data mining utility with projective sampling},
        year = {2009},
        booktitle = {Proceedings of the 15th ACM SIGKDD International Conference on Knowledge Discovery and Data Mining},
        pages = {487–496},
        series = {KDD '09}
}

@inproceedings{sabharwal2016selectingnearoptimallearnersincremental,
        title={Selecting Near-Optimal Learners via Incremental Data Allocation}, 
        author={Ashish Sabharwal and Horst Samulowitz and Gerald Tesauro},
        year={2016}, 
        booktitle = {Proceedings of the Thirtieth AAAI Conference on Artificial Intelligence},
        pages = {2007–2015},
        series = {AAAI'16}
}

@inproceedings{john1996static,
  title={Static Versus Dynamic Sampling for Data Mining.},
  author={John, George H and Langley, Pat},
  booktitle={{KDD}-96 Proceedings},
  volume={96},
  pages={367--370},
  year={1996}
}

@article{bertsimas2024towards,
  title={Towards Stable Machine Learning Model Retraining via Slowly Varying Sequences},
  author={Bertsimas, Dimitris and Digalakis Jr, Vassilis and Ma, Yu and Paschalidis, Phevos},
  year={2024},
  journal={arXiv preprint},
  volume={2403.19871}
}

@article{kabra2024limitations,
  title={The limitations of model retraining in the face of performativity},
  author={Kabra, Anmol and Patel, Kumar Kshitij},
  journal={arXiv preprint},
  year={2024},
  volume={2408.08499}
}

@inproceedings{bifet2007learning,
  title={Learning from time-changing data with adaptive windowing},
  author={Bifet, Albert and Gavalda, Ricard},
  booktitle={Proceedings of the 2007 SIAM international conference on data mining},
  pages={443--448},
  year={2007},
  organization={SIAM}
}

@inproceedings{pesaranghader2016fast,
  title={Fast hoeffding drift detection method for evolving data streams},
  author={Pesaranghader, Ali and Viktor, Herna L},
  booktitle={Joint European conference on machine learning and knowledge discovery in databases},
  pages={96--111},
  year={2016},
  organization={Springer}
}

@inproceedings{schwinn2022improving,
  title={Improving robustness against real-world and worst-case distribution shifts through decision region quantification},
  author={Schwinn, Leo and Bungert, Leon and Nguyen, An and Raab, Ren{\'e} and Pulsmeyer, Falk and Precup, Doina and Eskofier, Bj{\"o}rn and Zanca, Dario},
  booktitle={International Conference on Machine Learning},
  pages={19434--19449},
  year={2022},
  organization={PMLR}
}

@article{misic2020data,
  author  = {Mi{\v{s}}i{\'c}, Velibor V. and Perakis, Georgia},
  title   = {Data Analytics in Operations Management: A Review},
  journal = {Manufacturing \& Service Operations Management},
  year    = {2020},
  volume  = {22},
  number  = {1},
  pages   = {158--169},
  doi     = {10.1287/msom.2019.0805}
}

@article{davis2024best,
  author  = {Davis, Andrew M. and Mankad, Shawn and Corbett, Charles J. and Katok, Elena},
  title   = {{OM Forum---The Best of Both Worlds: Machine Learning and Behavioral Science in Operations Management}},
  journal = {Manufacturing \& Service Operations Management},
  year    = {2024},
  volume  = {26},
  number  = {5},
  pages   = {1605--1621},
  doi     = {10.1287/msom.2022.0553}
}

@article{bertsimas2020predictive,
  author  = {Bertsimas, Dimitris and Kallus, Nathan},
  title   = {From Predictive to Prescriptive Analytics},
  journal = {Management Science},
  year    = {2020},
  volume  = {66},
  number  = {3},
  pages   = {1025--1044},
  doi     = {10.1287/mnsc.2018.3253}
}

@article{elmachtoub2022smart,
  author  = {Elmachtoub, Adam N. and Grigas, Paul},
  title   = {Smart ``Predict, then Optimize''},
  journal = {Management Science},
  year    = {2022},
  volume  = {68},
  number  = {1},
  pages   = {9--26},
  doi     = {10.1287/mnsc.2020.3922}
}

@article{ban2019big,
  author  = {Ban, Gah-Yi and Rudin, Cynthia},
  title   = {The Big Data Newsvendor: Practical Insights from Machine Learning},
  journal = {Operations Research},
  year    = {2019},
  volume  = {67},
  number  = {1},
  pages   = {90--108},
  doi     = {10.1287/opre.2018.1757}
}

@article{ferreira2016analytics,
  author  = {Ferreira, Kris Johnson and Lee, Bin Hong Alex and Simchi-Levi, David},
  title   = {Analytics for an Online Retailer: Demand Forecasting and Price Optimization},
  journal = {Manufacturing \& Service Operations Management},
  year    = {2016},
  volume  = {18},
  number  = {1},
  pages   = {69--88},
  doi     = {10.1287/msom.2015.0561}
}

@article{caro2023believing,
  author  = {Caro, Felipe and S{\'a}ez de Tejada Cuenca, Anna},
  title   = {Believing in Analytics: Managers' Adherence to Price Recommendations from a {DSS}},
  journal = {Manufacturing \& Service Operations Management},
  year    = {2023},
  volume  = {25},
  number  = {2},
  pages   = {524--542},
  doi     = {10.1287/msom.2022.1166}
}

@article{kawaguchi2021workers,
  author  = {Kawaguchi, Kohei},
  title   = {When Will Workers Follow an Algorithm? A Field Experiment with a Retail Business},
  journal = {Management Science},
  year    = {2021},
  volume  = {67},
  number  = {3},
  pages   = {1670--1695},
  doi     = {10.1287/mnsc.2020.3599}
}

@article{feng2025tests,
  author  = {Feng, Kai and Hong, Han and Tang, Ke and Wang, Jingyuan},
  title   = {Statistical Tests for Replacing Human Decision Makers with Algorithms},
  journal = {Management Science},
  year    = {2025},
  volume  = {71},
  number  = {11},
  pages   = {9145--9170},
  doi     = {10.1287/mnsc.2023.01845}
}

@article{digalakis2025ml,
  title={{ML} Compass: Navigating Capability, Cost, and Compliance Trade-offs in {AI} Model Deployment},
  author={Digalakis Jr, Vassilis and Krishnan, Ramayya and Fernandez, Gonzalo Martin and Orfanoudaki, Agni},
  journal={arXiv preprint arXiv:2512.23487},
  year={2025}
}

@article{chan2026deployment,
  title={Deployment of AI-Assisted Interventions: Capacity Constraints and Noisy Compliance},
  author={Chan, Carri W and Han, Yi and Li, Hannah and Ranard, Benjamin L},
  journal={arXiv preprint arXiv:2604.14370},
  year={2026}
}

@article{cai2026diagnosing,
  title={Diagnosing model performance under distribution shift},
  author={Cai, Tiffany and Namkoong, Hongseok and Yadlowsky, Steve},
  journal={Operations Research},
  volume={74},
  number={2},
  pages={898--916},
  year={2026},
  publisher={INFORMS}
}

@article{kaps2025choose,
  title={How to Choose Among Technologies With Learning Curves: Making Better Investment Decisions},
  author={Kaps, Christian and Anderer, Arielle},
  journal={Available at SSRN 5168106},
  year={2025}
}

\ECSwitch

\section{Supplement for Section \ref{sec:theory}} \label{ec:sec3}

This section provides technical proofs and additional results for Section~\ref{sec:theory} of the main paper.

\subsection{Proofs for Section \ref{ssec:theory-optimal}} \label{ec:sec3.2}

\begin{proof}{Proof of Theorem~\ref{thm:finite-stopping-main}.}
In the finite-horizon setting,
\(
V_{\rm switch}(t)
=
-cnT-c_s+n(T-t)G(t)
=
-cnT-c_s+n\Phi(t).
\)
Thus maximizing \(V_{\rm switch}(t)\) is equivalent to maximizing \(\Phi(t)\), and
switching is profitable if and only if
$
n\max_{t\in\mathcal T}\Phi(t)\ge cnT+c_s.
$

For real \(t\in(0,T)\), write
$
G(t)=g^*-g_0n^{-\alpha}t^{-\alpha}
$. 
Then
$
\Phi'(t)
=
-g^*
+
g_0n^{-\alpha}t^{-\alpha-1}\left[\alpha T+(1-\alpha)t\right],
$
and
$
\Phi''(t)
=
-\alpha g_0n^{-\alpha}t^{-\alpha-2}
\left[(\alpha+1)T+(1-\alpha)t\right]<0.
$
Hence \(\Phi\) is strictly concave. If \(G(T)>0\), then
\(\Phi'(0^+)=+\infty\) and \(\Phi'(T)=-G(T)<0\), so the continuous maximizer is unique.
Strict concavity implies that the integer maximizer is one of the two nearest integers
to \(t^\dagger\).

Finally, because the discrete increments \(\Phi(t+1)-\Phi(t)\) are decreasing in \(t\),
the earliest integer maximizer is the first epoch at which the one-step gain from
waiting becomes nonpositive:
$
\Phi(t+1)-\Phi(t)\le 0.
$
This is exactly
$
(T-t)G(t)\ge (T-t-1)G(t+1).
$
\qedsymbol
\end{proof}

\begin{proof}{Proof of Proposition~\ref{prop:finite}.}
The first-order condition from the proof of Theorem~\ref{thm:finite-stopping-main} is
$
g^* (t^\dagger)^{\alpha+1}
=
g_0n^{-\alpha}\left[\alpha T+(1-\alpha)t^\dagger\right].
$
Since the right-hand side is \(O(T+t^\dagger)\), the solution satisfies
\(t^\dagger=O(T^{1/(1+\alpha)})\), hence \(t^\dagger=o(T)\). Therefore,
\(
\alpha T+(1-\alpha)t^\dagger\sim \alpha T.
\)
Substituting into the first-order condition gives
$
(t^\dagger)^{\alpha+1}
\sim
\frac{\alpha g_0T}{g^* n^\alpha}
\quad \Rightarrow \quad
t^\dagger
\sim
\left(\frac{\alpha g_0T}{g^* n^\alpha}\right)^{1/(1+\alpha)}.
$
\qedsymbol
\end{proof}

\subsection{Additional results for Section \ref{ssec:theory-optimal}: Extended setting} \label{ec:sec3-extended-setting}
This section generalizes the finite-horizon stopping rule of Section~\ref{ssec:theory-optimal} to a more general setting. Thm.~\ref{thm:threshold} and Prop.~\ref{prop:infinite} are the analogues of Thm.~\ref{thm:finite-stopping-main} and Prop.~\ref{prop:finite}, respectively. We omit their proofs, since the arguments are identical to those of their finite-horizon counterparts.

\paragraph{Extended setting description.}
As previewed in Remark~\ref{rem:disc}, we move from a finite horizon to an infinite
deployment horizon with discounting, where all cash flows are evaluated at time zero (before data collection begins)
using a per-period discount factor \(\beta\in(0,1)\). As previewed in
Remark~\ref{rem:costs}, we also allow unequal pre- and post-deployment per-sample costs:
$
C_{\rm pre}(t)=nc_{\rm pre},
\ 
C_{\rm post}(t)=nc_{\rm post}.
$
(The retraining-cost extension in Remark~\ref{rem:costs} affects the design of decision
epochs and is analyzed separately in Section~\ref{ec:sec4-epoch-design}.)
We keep the fixed sample flow \(N_t=nt\), full monitoring
\(\mathcal T_D=\mathcal T=\mathbb N\), and the power-law gap in
Assumption~\ref{assume:learning-curve}:
\(
G(t)=g^*-g_0(nt)^{-\alpha},
\  \text{with } g^*,g_0,\alpha>0.
\)

\paragraph{Value functions and switching rule.}
In this extended setting, switching at epoch \(t\) has present value
$
V_{\rm switch}(t)
=
-\sum_{\tau=1}^{t}\beta^\tau nc_{\rm pre}
-\beta^t c_s
+
\sum_{\tau=t+1}^{\infty}\beta^\tau n\{G(t)-c_{\rm post}\}.
$
Discarding at epoch \(t\) gives
$
V_{\rm discard}(t)
=
-\sum_{\tau=1}^{t}\beta^\tau nc_{\rm pre}.
$
Hence
\(
\Delta V(t)
=
V_{\rm switch}(t)-V_{\rm discard}(t)
=
-\beta^t c_s
+
\sum_{\tau=t+1}^{\infty}\beta^\tau n\{G(t)-c_{\rm post}\}.
\)
The switching value can be rewritten as
$
V_{\rm switch}(t)
=
-\frac{\beta nc_{\rm pre}}{1-\beta}
+
\frac{n\beta^{t+1}}{1-\beta}
\{G(t)-c_{\rm diff}\},
$
where
$
c_{\rm diff}
:=
c_{\rm post}-c_{\rm pre}
+
\frac{(1-\beta)c_s}{\beta n}.
$
Thus, conditional on switching, the oracle maximizes
$
\Phi_\beta(t)
:=
\beta^{t+1}\{G(t)-c_{\rm diff}\}.
$
The term \(c_{\rm diff}\) is the effective per-sample hurdle for switching: it combines
the post-minus-pre cost difference with the discounted per-sample equivalent of the
one-time switching cost.

\paragraph{Optimal stopping rule.}
Theorem~\ref{thm:threshold} is the discounted analog of
Theorem~\ref{thm:finite-stopping-main}. The main difference is that costs now affect both
the profitability and the timing of switching through \(c_{\rm diff}\). A higher
post-deployment cost \(c_{\rm post}\) or switching cost \(c_s\) raises the hurdle and
delays or prevents switching. A higher pre-deployment cost \(c_{\rm pre}\) lowers the
relative hurdle, making switching more attractive. 
\begin{theorem}[Exact discounted stopping rule]\label{thm:threshold}
Under the extended setting described above, define
\(
K_\beta
:=
\frac{n^\alpha(g^*-c_{\rm diff})}{g_0}.
\)
Then:
\begin{enumerate}
    \item \textbf{Feasibility.}
    Switching is globally optimal if and only if
    \(
    \max_{t\in\mathbb N}\Phi_\beta(t)\ge \beta c_{\rm pre}.
    \)
    A necessary condition is \(K_\beta>0\), equivalently \(g^*>c_{\rm diff}\). If
    \(K_\beta\le0\), immediate discard is optimal.

    \item \textbf{Existence and uniqueness.}
    If \(K_\beta>0\), the continuous problem
    \(
    \max_{t>0}\Phi_\beta(t)
    \)
    has a unique maximizer \(t_\beta^\dagger\) that satisfies
    \(
    t_\beta^\dagger
    >
    t_0
    :=
    \big((\alpha+1)K_\beta\big)^{-1/\alpha}.
    \)

    \item \textbf{Stopping rule.}
    The integer candidate is
    \(
    t^*_{\beta,{\rm int}}
    =
    \min\left\{
    t\in\mathbb N:
    G(t)-\beta G(t+1)\ge (1-\beta)c_{\rm diff}
    \right\}
    \). The optimal policy is to switch at \(t^*_{\beta,{\rm int}}\) if
    \(
    V_{\rm switch}(t^*_{\beta,{\rm int}})\ge V_{\rm discard}(0)=0,
    \)
    o/w discard immediately.
\end{enumerate}
\end{theorem}

Similar with the main setting, we continue with an asymptotic analysis of the optimal stopping time. Proposition~\ref{prop:infinite} shows that discounting plays the same role as a finite
horizon. As \(\beta\uparrow1\), the effective horizon is of order
\((1-\beta)^{-1}\). If \(c_{\rm pre}=c_{\rm post}\), then
\(
c_{\rm diff}
=
\frac{(1-\beta)c_s}{\beta n}
\to 0,
\)
and the scale in Proposition~\ref{prop:infinite} matches the finite-horizon scale in
Proposition~\ref{prop:finite} after replacing \(T\) by \((1-\beta)^{-1}\). 
This is coherent with a known fact from the literature on MDPs: discounted MDPs with a discount factor of $\beta$ and finite horizon MDPs with $T=\frac{1}{1-\beta}$ are good approximations to each other.
\begin{proposition}[Asymptotic discounted stopping rule]\label{prop:infinite}
Suppose \(g^*>c_{\rm post}-c_{\rm pre}\), so that \(g^*>c_{\rm diff}\) for all
\(\beta\uparrow1\). Then the continuous optimizer satisfies
\(
t_\beta^\dagger
\sim_{\beta\uparrow1}
\left(
\frac{\alpha g_0}
{n^\alpha (g^*-c_{\rm diff})(1-\beta)}
\right)^{1/(1+\alpha)}.
\)
\end{proposition}

\subsection{Proofs for Section \ref{ssec:theory-regretbound}} \label{ec:sec3.3}

\begin{proof}{Proof of Theorem ~\ref{thm:regretbound}.}
Define 
\(
g^*:= \max_{f_C \in \mathcal{F}_C}\mathbb{E}[\ell(f_C^{(t_k)}(X_\mathcal{C}),Y)]
-
\mathbb{E}[\ell(f_I(X_\mathcal{I}),Y)],
\) the maximal potential gain of the challenger. Note that since $f_I\in \mathcal{F}_C$, we have $g^*\geq 0$.
The following bound on the value of the oracle holds:$
    V_{\textsc{oracle}}(T) \leq \max\left(\left[g^*-c\right]n T-c_s,0\right). 
$
We show that under the conditions of the theorem, sublinear regret with respect to the upper bound of that inequality is achievable.
Let us start  by introducing the event that the generalization error of the challenger is well behaved. Set 
$
\delta'_{N_t}
:= 
2\sqrt{\frac{2d \,\log\!\bigl(e(1-\rho)N_t/d\bigr)}{(1-\rho)N_t}}
\;+\;
\sqrt{\frac{2\log(2T^2)}{(1-\rho)N_t}}.
$

Let \(\mathcal A_{\rm alg}\) be the event on which (C1) and (C2) hold.
By assumption, \(\mathbb P(\mathcal A_{\rm alg})\ge 1-w_1/T\). Let \(\mathcal A_{\rm gen}\) be the event on which
$
g^*-G(t)\le \delta'_{N_t},\qquad t=1,\ldots,T.
$ Set \(\mathcal A=\mathcal A_{\rm alg}\cap\mathcal A_{\rm gen}\).
By Corollary 3.19 of \cite{mohri2018foundations} and a union bound over
\(t\le T\), 
\vspace{-20pt}
\begin{align}\label{eq:lowerboundprobalearningtheory}
\mathbb P(\mathcal A_{\rm gen})\ge 1-1/T \text{ and }\mathbb P(\mathcal A^c)\le \frac{1+w_1}{T}.
\end{align}
Decomposing the regret along event $\mathcal{A}$, we obtain:
\[
    R_{\text{alg}}(T)=\underbrace{\mathbb{E}[(V_{\textsc{oracle}}(T)-V_{\textsc{alg}}(T))\Ind{\mathcal{A}}]}_{(i)}+\underbrace{\mathbb{E}[(V_{\textsc{oracle}}(T)-V_{\textsc{alg}}(T))\Ind{\mathcal{A}^c}].}_{(ii)}
\]
We start by upper bounding (ii). Since the loss takes value in $\{-1;1\}$, $g^*\leq 2$, hence, by the upper bound on the value of the oracle, $ V_{\textsc{oracle}}(T) \leq 2 nT.$ On the other hand, the algorithm can at worse pay all the costs while switching to the worst possible challenger, which implies $V_{\textsc{alg}}(T)\geq -c_s-(c+2)nT$. Combining those two facts:
$
    (ii)\leq ((4+c)n+\frac{c_s}{T})T\mathbb{P}(\mathcal{A}^c)\leq  (1+w_1)((c+4)n+\frac{c_s}{T}),
$
where the second inequality holds by Equation ~\ref{eq:lowerboundprobalearningtheory}. 

We now work on upper bounding $(i)$. Consider any $t \geq w_2 T^{2/3}$. Under event $\mathcal{A}$, we have $
g^* \leq G(t) +\delta'_{w_2 n T^{2/3}}.
$
This implies:
\vspace{-10pt}
\begin{align*}
    V_{\mathrm{switch}}(t)\geq& -c  n T -c_s+ n(T-t) \left(g^*-2\sqrt{\frac{2d \,\log\!\bigl(e(1-\rho)w_2nT^{2/3}/d\bigr)}{(1-\rho)w_2nT^{2/3}}}
\;-\;
\sqrt{\frac{2\log(2T^2)}{(1-\rho)w_2nT^{2/3}}}\right),\\
    \geq& [g^*-c]nT-c_S -ntg^*-nT^{2/3}\left(2\sqrt{\frac{2d \,\log\!\bigl(e(1-\rho)w_2nT^{2/3}/d\bigr)}{(1-\rho)w_2n}}
\;+\;
\sqrt{\frac{2\log(2T^2)}{(1-\rho)w_2n}}\right).
\end{align*}
Therefore, for any $t \in [w_2 T^{2/3};w_3 T^{2/3}\sqrt{\log(T)}]$, there exists a constant $w_5$, such that for any large enough $T$, we have:
$
    V_{\mathrm{switch}}(t)\geq
    [g^*-c]nT-c_S -w_5T^{2/3}\sqrt{\log(T)}.
$
This, combined with condition (C1) implies:
$
    V_{\mathrm{switch}}(t_{\textsc{alg}})\geq
     [g^*-c]nT-c_S -w_5T^{2/3}\sqrt{\log(T)}.
$
Condition (C1) also implies:
$
    V_{\mathrm{discard}}(t_{\textsc{alg}})\geq
     -w_3n cT^{2/3}\sqrt{\log(T)}.
$
By condition (C2),
$
    V_{\textsc{alg}}(T)\geq \max (V_{\mathrm{switch}}(t_{\textsc{alg}});V_{\mathrm{discard}}(t_{\textsc{alg}}))-w_4 T^{2/3}\sqrt{\log(T)}.
$
Putting those last inequalities together yields:
\vspace{-10pt}
\begin{align*}
    V_{\textsc{alg}}(T)\geq& \max ([g^*-c]nT-c_S -w_5T^{2/3}\sqrt{\log(T)}; -w_3n cT^{2/3}\sqrt{\log(T)})-w_4 T^{2/3}\sqrt{\log(T)} \\
    \geq & \max([g^*-c]nT-c_S ; 0) -(w_3n c+w_4+w_5)T^{2/3}\sqrt{\log(T)}.
\end{align*}
The first term on the right hand side matches the upper bound on the value of the oracle, which gives a bound on (i). Combining with the bound on (ii) yields the theorem. 
\qedsymbol
\end{proof}

\section{Supplement for Section \ref{sec:algs}} \label{ec:sec4}

This section provides additional results and technical proofs for theoretical guarantees presented in Section~\ref{sec:algs} of the main paper.
\subsection{Designing Decision Epochs} \label{ec:sec4-epoch-design}

This subsection formalizes the cost-responsiveness trade-off in the choice of decision epochs.
We work in the main finite-horizon setting with \(N_t=nt\) and
\(G(t)=g^*-g_0(nt)^{-\alpha}\). We compare the uniform schedule
\(t_k=\Lambda k\) and the geometric schedule \(t_k=\Lambda\lambda^{k-1}\), \(\lambda>1\).
Integer rounding of the geometric schedule does not affect the asymptotic orders below.
As previewed in Remark~\ref{rem:costs}, suppose that retraining at time \(t\) incurs
\(
C_{\rm train}(t)=c_{\rm train}N_t^q,\ q\ge0.
\)
Let \(t^\dagger\) denote the continuous oracle stopping time from
Theorem~\ref{thm:finite-stopping-main}. We evaluate schedules along two dimensions:
the cumulative retraining cost up to \(t^\dagger\), and the normalized value loss from
restricting switching to the available epochs,
\(
W(\{t_k\})
:=
\frac{
V_{\rm switch}(t^\dagger)-\max_{t_k\le T}V_{\rm switch}(t_k)
}{nT}.
\)
We have the following proposition:
\begin{proposition}[Cost and responsiveness]
\label{prop:epoch-cost}
\label{prop:epoch-responsiveness}
Fix \(t^\dagger\in(0,T)\).
\begin{enumerate}
    \item[{(a)}] \textbf{Retraining cost.}
    If \(q>0\), then
    \(
    C_{\rm unif}(t^\dagger)
    =
    \Theta\!\left(\frac{(t^\dagger)^{q+1}}{\Lambda}\right),
    \ 
    C_{\rm geom}(t^\dagger)
    =
    \Theta\!\left((t^\dagger)^q\right)
    =
    \Theta\!\left(C_{\rm train}(t^\dagger)\right).
    \)
    
    If \(q=0\), then the number of retraining rounds satisfies
    \( 
    K_{\rm unif}(t^\dagger)
    =
    \Theta\!\left(\frac{t^\dagger}{\Lambda}\right),
    \ 
    K_{\rm geom}(t^\dagger)
    =
    \Theta(\log t^\dagger).
    \)

    \item[{(b)}] \textbf{Responsiveness loss.}
    There exist constants \(C_1,C_2>0\) such that, for \(t^\dagger\) sufficiently large,
    \(
    W_{\rm unif}
    \le
    C_1\frac{\Lambda^2}{(t^\dagger)^{\alpha+2}},
    \)
    where \(\tilde t\) is the uniform epoch closest to \(t^\dagger\). For the geometric schedule,
    if \([t^\dagger/\lambda,\lambda t^\dagger]\subset(0,T)\), then
    \(
    W_{\rm geom}
    \le
    C_2\frac{(\lambda-1)^2}{(t^\dagger)^\alpha}.
    \)
\end{enumerate}
\end{proposition}

\begin{proof}{Proof of Proposition~\ref{prop:epoch-responsiveness}}
For part (a), since \(N_t=nt\), the cost of retraining at epoch \(t\) is
\(c_{\rm train}(nt)^q\). Under the uniform schedule, with
\(K=\lfloor t^\dagger/\Lambda\rfloor\),
\(
C_{\rm unif}(t^\dagger)
=
c_{\rm train}n^q\Lambda^q\sum_{k=1}^K k^q
=
\Theta\!\left(c_{\rm train}n^q\Lambda^qK^{q+1}\right)
=
\Theta\!\left(\frac{(t^\dagger)^{q+1}}{\Lambda}\right).
\)
Under the geometric schedule, let \(b\) be the largest integer such that
\(t_b\le t^\dagger<t_{b+1}\). Then
\(
C_{\rm geom}(t^\dagger)
=
c_{\rm train}n^q\Lambda^q\sum_{k=0}^{b-1}\lambda^{qk}
=
\Theta\!\left(c_{\rm train}n^q t_b^q\right).
\)
Because \(t^\dagger/\lambda\le t_b\le t^\dagger\), this is
\(\Theta((t^\dagger)^q)\). When \(q=0\), total cost is proportional to the number
of epochs, giving \(K_{\rm unif}=\Theta(t^\dagger/\Lambda)\) and
\(K_{\rm geom}=\Theta(\log_\lambda(t^\dagger/\Lambda))=\Theta(\log t^\dagger)\).

For part (b), constants cancel in value differences, so
\(
W(\{t_k\})
=
\frac{\Phi(t^\dagger)-\max_{t_k\le T}\Phi(t_k)}{T},
\ 
\Phi(t)=(T-t)G(t).
\)
Let \(\bar\Phi(t)=\Phi(t)/T\). Under the power-law gap,
\(
G'(t)=\alpha g_0n^{-\alpha}t^{-(\alpha+1)},
\ 
G''(t)=-\alpha(\alpha+1)g_0n^{-\alpha}t^{-(\alpha+2)}.
\)
Hence
\(
\bar\Phi''(t)
=
\frac{(T-t)G''(t)-2G'(t)}{T},
\)
and, since \(t\le T\), there is a constant \(C>0\) such that
\(
|\bar\Phi''(t)|\le C t^{-(\alpha+2)}.
\)
Because \(\bar\Phi'(t^\dagger)=0\), Taylor's theorem implies that for any nearby epoch
\(\tilde t\),
\(
\bar\Phi(t^\dagger)-\bar\Phi(\tilde t)
\le
C(t^\dagger)^{-(\alpha+2)}(\tilde t-t^\dagger)^2,
\)
after adjusting \(C\). 

For the uniform schedule, the closest epoch satisfies
\(|\tilde t-t^\dagger|\le \Lambda\), yielding
\(
W_{\rm unif}
\le
C_1\frac{\Lambda^2}{(t^\dagger)^{\alpha+2}}.
\)

For the geometric schedule, if \(t_k\le t^\dagger<t_{k+1}\), then choosing the closer
of \(t_k,t_{k+1}\) gives
\(
|\tilde t-t^\dagger|
\le
\frac{t_{k+1}-t_k}{2}
=
\frac{(\lambda-1)t_k}{2}
\le
\frac{\lambda-1}{2}t^\dagger.
\)
Substitution gives
\(
W_{\rm geom}
\le
C_2\frac{(\lambda-1)^2}{(t^\dagger)^\alpha}.
\)
\end{proof}
\subsection{Proofs for Section \ref{ssec:alg-ose}} \label{ec:sec4.2}

\begin{proof}{Proof of Proposition ~\ref{prop:strawmanverifiesregret}.}
    By design, the OSE algorithm with $t_k=T^{2/3}$ verifies (C1) with $w_2=w_3=1$. Let us show it also verifies (C2) with high enough probability.

    The OSE algorithm switches if and only if 
   $ \widehat{\Delta V(T^{2/3})}\geq 0.
    $
Therefore (C2) is valid as soon as:
$
 |\widehat{\Delta V(T^{2/3})} -\Delta V(T^{2/3})| \leq w_4 T^{2/3}\sqrt{\log(T)}.
$
    The following holds:
    $
    |\widehat{\Delta V(T^{2/3})} -\Delta V(T^{2/3})| \leq Tn|G(T^{2/3})-\widehat{G(T^{2/3})}|.
    $
    We have:
\vspace{-20pt}    
\begin{align}
  &\mathbb{P}\left(Tn|G(T^{2/3})-\widehat{G(T^{2/3})}| \geq T^{2/3}\sqrt{\frac{16n\log(T)}{\rho }} \right)=          \mathbb{P}\left(|G(T^{2/3})-\widehat{G(T^{2/3})}| \geq\sqrt{\frac{16\log(T)}{\rho n T^{2/3}}} \right) \notag\\
  &\hspace{8cm}\leq  2 \exp \left(-\frac{\rho n T^{2/3}}{8}\times\frac{16\log(T)}{\rho n T^{2/3}}\right)
            \leq \frac{2}{T},
    \end{align}
   where the second line holds by Hoeffding's inequality. 
    This implies that the OSE algorithm verifies (C2) with $w_1=2$ and $w_4 =4 \sqrt{n/\rho}$. \qedsymbol
\end{proof}
\subsection{Proofs for Section \ref{ssec:alg-lse}} \label{ec:sec4.3}

\begin{proof}{Proof of Proposition~\ref{prop:algorithmslopebasedverifiesregret}.}
Consider LSEc with confidence parameter $\gamma>0$ and geometric epoch rate $\lambda>1$. At any epoch $k\ge 3$, LSEc continues to the next epoch if and only if
\vspace{-20pt}
\begin{align}
\widehat{V}_{\mathrm{switch}}(t_k)
    &<\max_{k'>k} \widehat{V}_{\mathrm{switch}}^{\mathrm{UB}}(t_{k'}), 
    \label{eq:stopcondition1}\\
\text{and }  -nct_k
  &<\max_{k'>k} \widehat{V}_{\mathrm{switch}}^{\mathrm{UB}}(t_{k'}).
    \label{eq:stopcondition2}
\end{align}
We first prove the lower bound in condition (C1). 
By the confidence adjustment in LSEc and the geometric schedule, there exists a constant
$a>0$, independent of $T$, such that for all $k\ge 3$,
$
\widehat s_k \ge a t_k^{-3/2}.
$
Indeed, this follows from
\[
\widehat s_k
\ge
\frac{2\gamma}
{(1-\rho)(N_{t_k}-N_{t_{k-1}})\sqrt{\rho N_{t_k}}},
\qquad
t_k=\lambda t_{k-1},\quad N_{t_k}=nt_k .
\]

To verify \eqref{eq:stopcondition1}, compare epoch $k$ with $k+1$. We have
\[
\widehat{V}_{\mathrm{switch}}^{\mathrm{UB}}(t_{k+1})
-
\widehat{V}_{\mathrm{switch}}(t_k)
=
n(t_{k+1}-t_k)
\left\{(1-\rho)n(T-t_{k+1})\widehat s_k-\widehat G(t_k)\right\}.
\]
Since $|\widehat G(t_k)|\le 2$, a sufficient condition for \eqref{eq:stopcondition1} is
$
(1-\rho)n(T-t_{k+1})\widehat s_k>2.
$
Using $\widehat s_k \ge a t_k^{-3/2}$, this condition holds whenever
$t_k\le c_1T^{2/3}$, for some constant $c_1>0$ and all sufficiently large $T$.

To verify \eqref{eq:stopcondition2}, choose ${\bar k}= \lfloor \frac{\log(T/2\Lambda)}{\log(\lambda)}+1\rfloor$, so that
   $ \frac{T}{2\lambda}\leq t_{\bar k} \leq  \frac{T}{2}. $ We have,
$
\widehat{V}_{\mathrm{switch}}^{\mathrm{UB}}(t_{\bar k})+cnt_k\ge
\frac{Tn}{2}
\left[
-2-2c-\frac{2c_s}{n}
+(1-\rho)n\left(\frac{T}{2\lambda}-t_k\right)\widehat s_k
\right].
$
Thus \eqref{eq:stopcondition2} holds whenever
$
(1-\rho)n\left(\frac{T}{2\lambda}-t_k\right)\widehat s_k
>
2\left(c+1+\frac{c_s}{n}\right).
$
Together with $\widehat s_k \ge a t_k^{-3/2}$, this holds whenever $t_k\le c_2T^{2/3}$, for some
constant $c_2>0$ and all sufficiently large $T$. So, for some $w_2>0$,
$
t_k<w_2T^{2/3}
\quad\Longrightarrow\quad
\eqref{eq:stopcondition1}\text{ and }\eqref{eq:stopcondition2}\text{ both hold}.
$
Since LSEc continues whenever both conditions hold,
$
t_{\mathrm{LSEc}}\ge w_2T^{2/3}.
$

We next prove the upper bound in condition (C1). We first combine Equation \ref{eq:lowerboundprobalearningtheory} and Hoeffding's inequality together with a union bound and obtain:
\begin{lemma}\label{lem:boundrademacherestimate2}
  With probability at least $1-2/T$, uniformly over
all epochs $t_k\le T$,
\[
|G(t_k)-\widehat G(t_k)|
\le C\sqrt{\frac{\log T}{N_{t_k}}},
\qquad
|g^*-\widehat G(t_k)|
\le C\sqrt{\frac{\log T}{N_{t_k}}}, \tag{event $\mathcal{C}$}
\]
where $C$ is independent of $T$. 
\end{lemma}

Under event $\mathcal{C}$, the confidence-adjusted slope satisfies, for
another constant $C'>0$,
$
\widehat s_k\le C'\frac{\sqrt{\log T}}{t_k^{3/2}}.
$
This follows from the triangle inequality applied to
$|\widehat G(t_k)-\widehat G(t_{k-1})|$, the displayed uniform bounds, and the geometric schedule. Suppose first that $g^*>0$. For all sufficiently large $T$ and all $t_k\ge T^{2/3}$, event
$\mathcal C$ implies $\widehat G(t_k)\ge g^*/2$. For any $k'>k$,
$
\widehat{V}_{\mathrm{switch}}^{\mathrm{UB}}(t_{k'})
-
\widehat{V}_{\mathrm{switch}}(t_k)
\le
n(t_{k'}-t_k)
\left\{
(1-\rho)nT\widehat s_k-\widehat G(t_k)
\right\}.
$
Thus \eqref{eq:stopcondition1} fails whenever
$
(1-\rho)nT\widehat s_k\le \frac{g^*}{2}.
$
Using $
\widehat s_k\le C'\frac{\sqrt{\log T}}{t_k^{3/2}}.
$, this occurs for all
$
t_k\ge C_1T^{2/3}(\log T)^{1/3},
$
for some constant $C_1>0$.

Now suppose $g^*=0$. Since Proposition~\ref{prop:algorithmslopebasedverifiesregret} assumes
$g^*+c>0$, we have $c>0$. For all sufficiently large $T$ and all $t_k\ge T^{2/3}$, event
$\mathcal C$ implies $\widehat G(t_k)\le c/4$. For $t_k\le T/2$ and all $k'\ge k$,
$
\widehat{V}_{\mathrm{switch}}^{\mathrm{UB}}(t_{k'})+cnt_k
\le
-\frac{cnT}{2}-c_s+nT\left(\frac c4+nT\widehat s_k\right).
$
Hence \eqref{eq:stopcondition2} fails whenever $nT\widehat s_k\le c/4$. By
$
\widehat s_k\le C'\frac{\sqrt{\log T}}{t_k^{3/2}}.
$, this occurs for all
$
t_k\ge C_2T^{2/3}(\log T)^{1/3},
$
for some constant $C_2>0$.

So, on $\mathcal C$, one of \eqref{eq:stopcondition1} and \eqref{eq:stopcondition2} fails
once $t_k$ exceeds a constant multiple of $T^{2/3}(\log T)^{1/3}$. Since the epochs are geometric,
the terminal time is at most a factor $\lambda$ larger than this threshold. Thus, for some constant
$w_3>0$,
$
t_{\mathrm{LSEc}}
\le
w_3T^{2/3}(\log T)^{1/3}
\le
w_3T^{2/3}\sqrt{\log T}
$
for all sufficiently large $T$. Together with the lower bound on $t_{\mathrm{LSEc}}$ previously obtained, this proves condition (C1)
on $\mathcal C$.

It remains to verify condition (C2). At the terminal epoch, LSEc switches if and only if
$\widehat{\Delta V}(t_{\mathrm{LSEc}})\ge0$. Moreover,
$
\left|
\widehat{\Delta V}(t_{\mathrm{LSEc}})
-
\Delta V(t_{\mathrm{LSEc}})
\right|
\le
nT
\left|
\widehat G(t_{\mathrm{LSEc}})
-
G(t_{\mathrm{LSEc}})
\right|.
$
On $\mathcal C$,
$
nT
\left|
\widehat G(t_{\mathrm{LSEc}})
-
G(t_{\mathrm{LSEc}})
\right|
\le
nT C\sqrt{\frac{\log T}{N_{t_{\mathrm{LSEc}}}}}.
$
Using $N_{t_{\mathrm{LSEc}}}=nt_{\mathrm{LSEc}}$ and the lower bound
$t_{\mathrm{LSEc}}\ge w_2T^{2/3}$, we obtain
$
\left|
\widehat{\Delta V}(t_{\mathrm{LSEc}})
-
\Delta V(t_{\mathrm{LSEc}})
\right|
\le
w_4T^{2/3}\sqrt{\log T}
$
for some constant $w_4>0$. Hence, if
$\Delta V(t_{\mathrm{LSEc}})-w_4T^{2/3}\sqrt{\log T}>0$, then
$\widehat{\Delta V}(t_{\mathrm{LSEc}})>0$ and LSEc switches; if
$\Delta V(t_{\mathrm{LSEc}})+w_4T^{2/3}\sqrt{\log T}<0$, then
$\widehat{\Delta V}(t_{\mathrm{LSEc}})<0$ and LSEc discards. Thus condition (C2) holds on
$\mathcal C$.

Since $\mathbb P(\mathcal C)\ge 1-2/T$, LSEc satisfies conditions (C1) and (C2) with probability
at least $1-2/T$. 
\end{proof}

\section{Credit Scoring Case Study: Additional Empirical Results} \label{ec:empirical_results}

This section describes data management procedures, reports complementary results to the main analysis, and presents sensitivity analyses and robustness checks.

\subsection{Data management}
\label{sec:data}

Our empirical analysis uses a comprehensive dataset from Lending Club, a large peer-to-peer lending platform. The original dataset, obtained from Kaggle, covers the period 2007--2020 and contains 2,925,493 loans with 141 variables.\footnote{The dataset is available at \url{https://www.kaggle.com/datasets/ethon0426/lending-club-20072020q1}.}

\paragraph{Sample selection.} We restrict the sample to loans with a final status of {fully paid} or {charged off}, excluding 1,065,162 loans with other outcomes. We further remove 588,361 loans with unverified income and 44,371 loans involving co-borrowers. An additional 138,071 observations are dropped due to missing values, and 2,237 are excluded as outliers.
We exclude data from 2018--2020 to avoid the Lending Club crisis period, during which loan applications declined sharply and default rates exhibited substantial instability. Specifically, applications fell by more than 45\% per year (45.45\% in 2018 and 65.89\% in 2019), while default rates fluctuated markedly relative to 2017 (+2.96 percentage points in 2018 and --5.06 percentage points in 2019). We also omit data from 2012--2013, which display unusually low default rates compared with subsequent years (17.7\% in 2012 and 16.87\% in 2013 versus 19.4\% in 2014, the lowest among retained years). Finally, we drop the first four months of 2014 and the last eight months of 2017 due to abnormally low and high default rates, respectively. The final dataset contains 705,302 loans between 5/2014 and 4/2017.

\paragraph{Feature selection.} We begin by excluding variables that (i) contain post-issuance information, (ii) relate to co-borrowers, (iii) have more than 5\% missing values, (iv) are highly correlated with variables retained in the dataset, (v) lack a clear definition, or (vi) are unlikely to be informative for predicting loan default. In addition, to facilitate a clear illustration of the algorithms’ behavior across different regimes, including scenarios in which switching is optimal, we exclude three variables: interest rate, home ownership status, and employment title. This restriction increases the performance gap between the challenger and the incumbent models and enables a clearer analysis of the resulting switching decisions. The {incumbent model} has therefore access to a feature set $\mathcal{I}$ of size $|\mathcal{I}|=7$, which consists of annual income, job tenure, debt-to-income ratio, FICO score, funding amount, loan duration, and loan purpose. In contrast, the {challenger model} benefits from access to all the alternative data. (e.g., loan grade, number of open credit lines, bank cards, installments, mortgages, and revolving accounts), thereby expanding the feature set to $\mathcal{C}$, with $|\mathcal{C}|=29$.  Among all these features, 27 are continuous and 2 are categorical.
Continuous variables are standardized, categorical variables are one-hot encoded, and loan grade and subgrade, which are originally categorical, are transformed into numerical (ordinal) scales. Table \ref{tab:feature_sets} reports the complete list of features and their descriptions for both the incumbent and challenger models.
\begin{table}[ht] 
\centering 
\label{tab:feature_sets} 
\renewcommand{\baselinestretch}{0.8}\selectfont 
\resizebox{\textwidth}{!}{ 
\begin{tabular}{p{0.6\linewidth} p{1\linewidth}} 
\toprule 
\textbf{Incumbent features} & \textbf{Description} \\ 
\midrule 
\texttt{annual\_inc}, \texttt{dti} & Borrower's self-reported annual income and debt-to-income ratio. \\ 
\texttt{fico\_range\_high} & Upper boundary of the borrower's FICO range at loan origination. \\ 
\texttt{funded\_amnt}, \texttt{loan duration} & Total amount committed to the loan and loan duration, equal to 1 for 60 months and 0 for 36 months. \\ 
\texttt{emp\_length}, \texttt{purpose} & Employment length in years and borrower-provided loan purpose. \\ 
\midrule 
\textbf{Additional challenger features} & \textbf{Description} \\ 
\midrule 
\texttt{avg\_cur\_bal}, \texttt{bc\_open\_to\_buy}, \texttt{bc\_util}, \texttt{revol\_bal}, \texttt{revol\_util} & Account balances, available revolving bankcard credit, bankcard utilization, total revolving balance, and revolving utilization. \\ 
\texttt{delinq\_2yrs}, \texttt{pub\_rec}, \texttt{pub\_rec\_bankruptcies}, \texttt{tax\_liens} & Recent delinquencies, derogatory public records, public record bankruptcies, and tax liens. \\ 
\texttt{inq\_last\_6mths}, \texttt{mort\_acc} & Credit inquiries in past 6 months, excl. auto and mortgage inquiries, and \# of mortgage accounts. \\ 
\texttt{open\_acc}, \texttt{num\_rev\_accts}, \texttt{num\_bc\_tl}, \texttt{num\_il\_tl}, \texttt{num\_actv\_bc\_tl} & Number of open credit lines, revolving accounts, bankcard accounts, installment accounts, and currently active bankcard accounts. \\ 
\texttt{mths\_since\_recent\_bc}, \texttt{mo\_sin\_old\_rev\_tl\_op}, \texttt{mo\_sin\_rcnt\_rev\_tl\_op}, \texttt{mo\_sin\_rcnt\_tl} & Months since the most recent bankcard account, oldest revolving account, most recent revolving account, and most recent account were opened. \\ 
\texttt{grade}, \texttt{sub\_grade} & LC assigned loan grade and subgrade. \\ 
\bottomrule 
\end{tabular}} 
\caption{Incumbent and challenger features} 
\label{tab:feature_sets} 
\end{table}

\subsection{Core scenarios: Extended Numerical Results} \label{ssec:scenario_selection}

This section presents additional numerical results for the core scenarios E.1 and E.2. 

\subsection{Sensitivity Analysis: Effect of Algorithm Hyperparameters}
\label{subsec:sensitivity}

We analyze how the performance of GSE and LSEc algorithms varies with the confidence parameter $\gamma$. For GSE, a larger $\gamma$ corresponds to a more conservative decision rule and therefore leads to later stopping. For LSEc, a larger $\gamma$ results in a more optimistic assessment of future gaps, which likewise delays stopping.

\paragraph{Design.}
To assess sensitivity to \(\gamma\), we evaluate algorithm performance across a large set of experiments constructed from the same parameter grid used to define the core scenarios (see Section~\ref{ssec:scenario_selection}), for both LR and LightGBM, over a prespecified range \(\Gamma\) of \(\gamma\) values. For each algorithm, we proceed in two steps. First, we consider a wide range of \(\gamma\) values, evaluate performance across experiments, and select a reference value. Second, we examine local sensitivity by varying \(\gamma\) within a tighter neighborhood around the selected value.

\begin{figure}[h!]
  \centering
  \begin{subfigure}[b]{0.24\textwidth}
  \hspace{7pt}
    \includegraphics[trim={0 0 0 0.75cm},clip, width=0.9\linewidth]{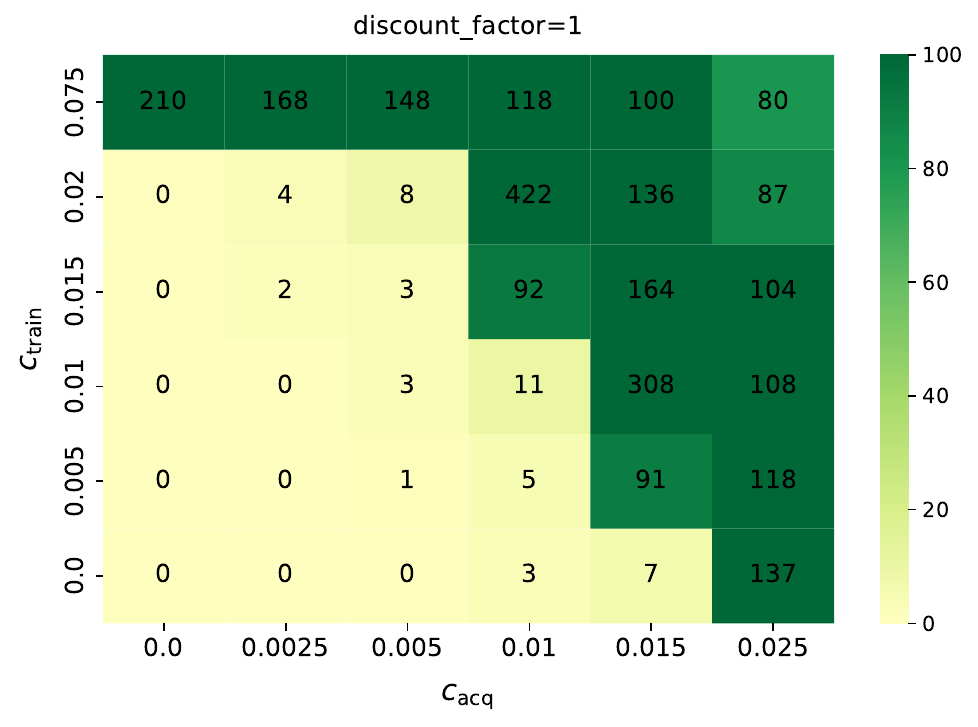}
    \caption{LR: $\tilde{\gamma} \approx 44.68$}
  \end{subfigure}
  \begin{subfigure}[b]{0.24\textwidth}
  \hspace{7pt}
    \includegraphics[trim={0 0 0 0.75cm},clip, width=0.9\linewidth]{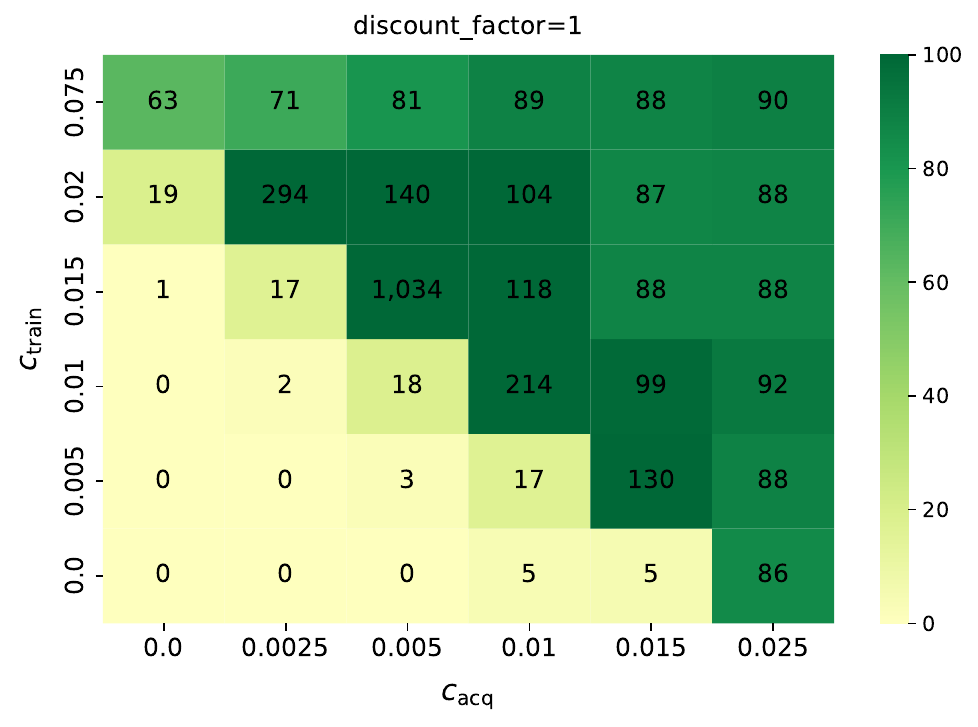}
    \caption{LightGBM: $\tilde{\gamma} \approx 44.68$}
  \end{subfigure}
    \begin{subfigure}[b]{0.24\textwidth}
    \hspace{7pt}
    \includegraphics[trim={0 0 0 0.75cm},clip, width=0.9\linewidth]{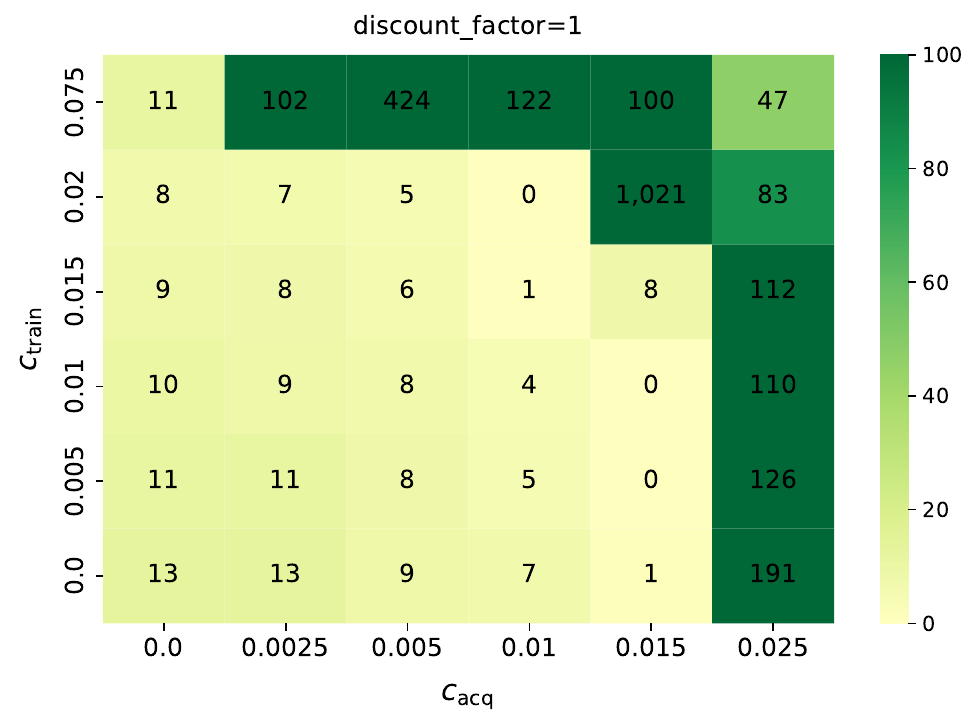}
    \caption{LR: $\tilde{\gamma} \approx 25.76$}
  \end{subfigure}
  \begin{subfigure}[b]{0.24\textwidth}
  \hspace{7pt}
    \includegraphics[trim={0 0 0 0.75cm},clip, width=0.9\linewidth]{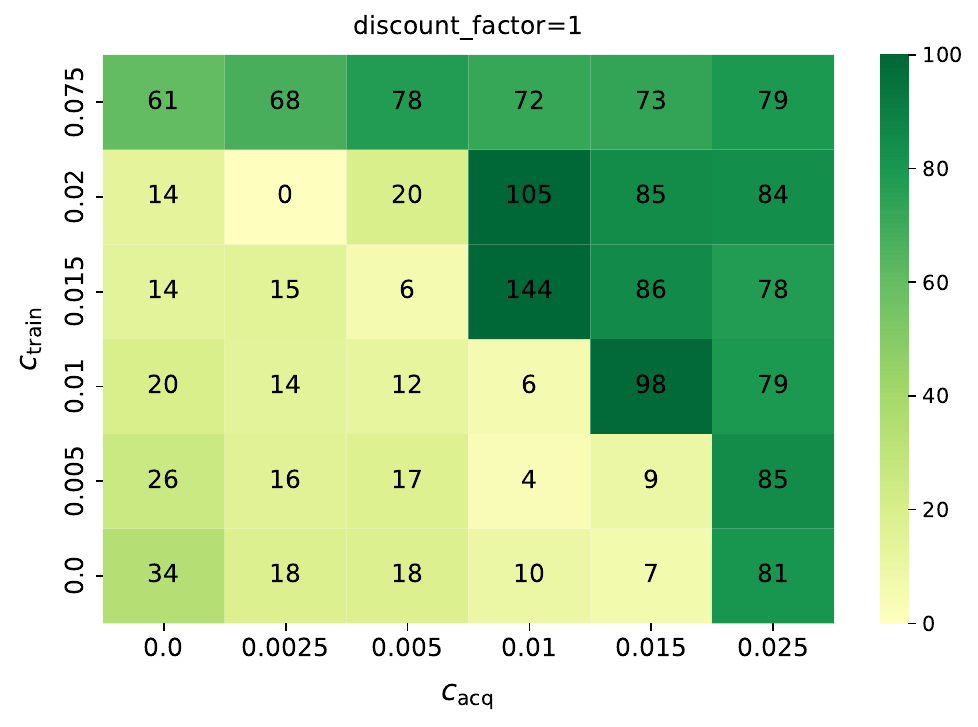}
    \caption{LightGBM: $\tilde{\gamma} \approx 25.76$}
  \end{subfigure}
    \begin{subfigure}[b]{0.24\textwidth}
    \hspace{7pt}
    \includegraphics[trim={0 0 0 0.75cm},clip, width=0.9\linewidth]{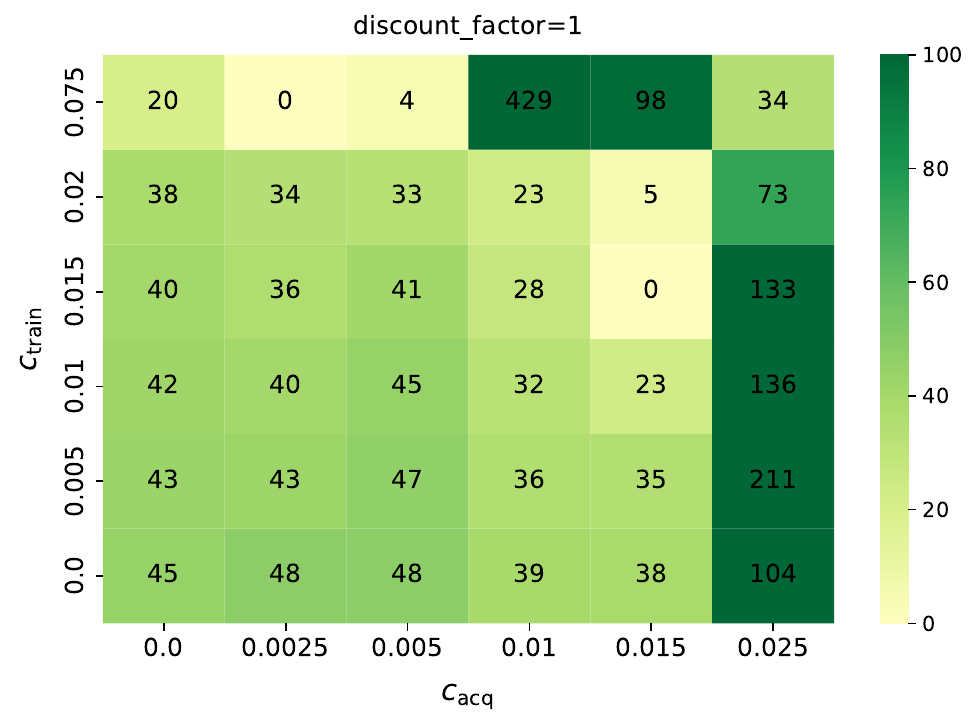}
    \caption{LR: $\tilde{\gamma} \approx 1.92$}
    \label{fig:E.1_gamma}
  \end{subfigure}
  \begin{subfigure}[b]{0.24\textwidth}
  \hspace{7pt}
    \includegraphics[trim={0 0 0 0.75cm},clip, width=0.9\linewidth]{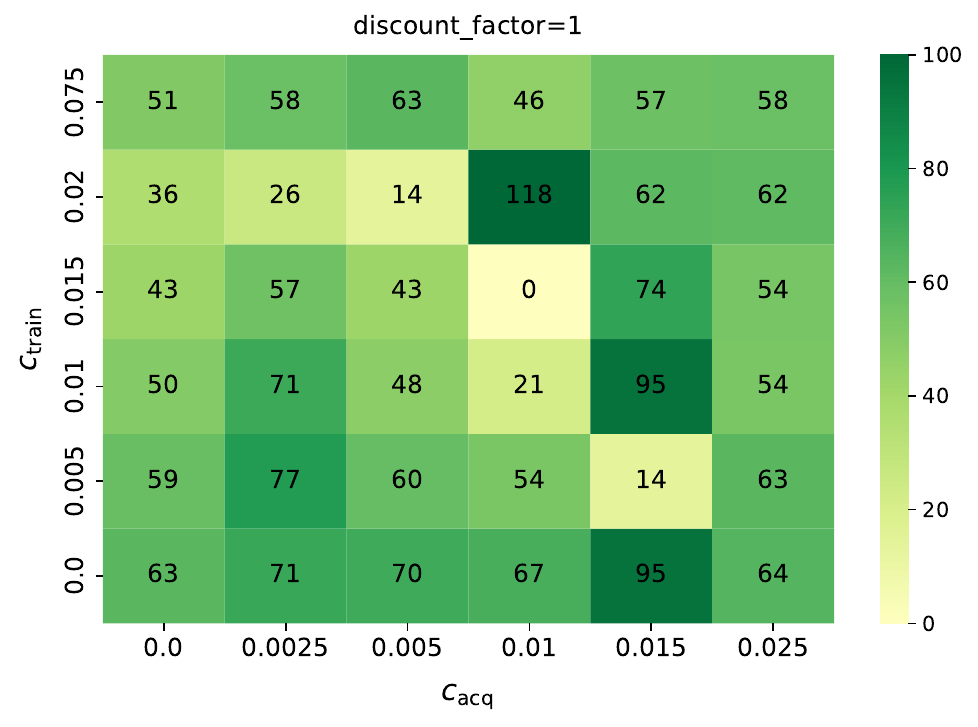}
    \caption{LightGBM: $\tilde{\gamma} \approx 1.92$}
  \end{subfigure} 
  \begin{subfigure}[b]{0.24\textwidth}
  \hspace{7pt}
    \includegraphics[trim={0 0 0 0.75cm},clip, width=0.9\linewidth]{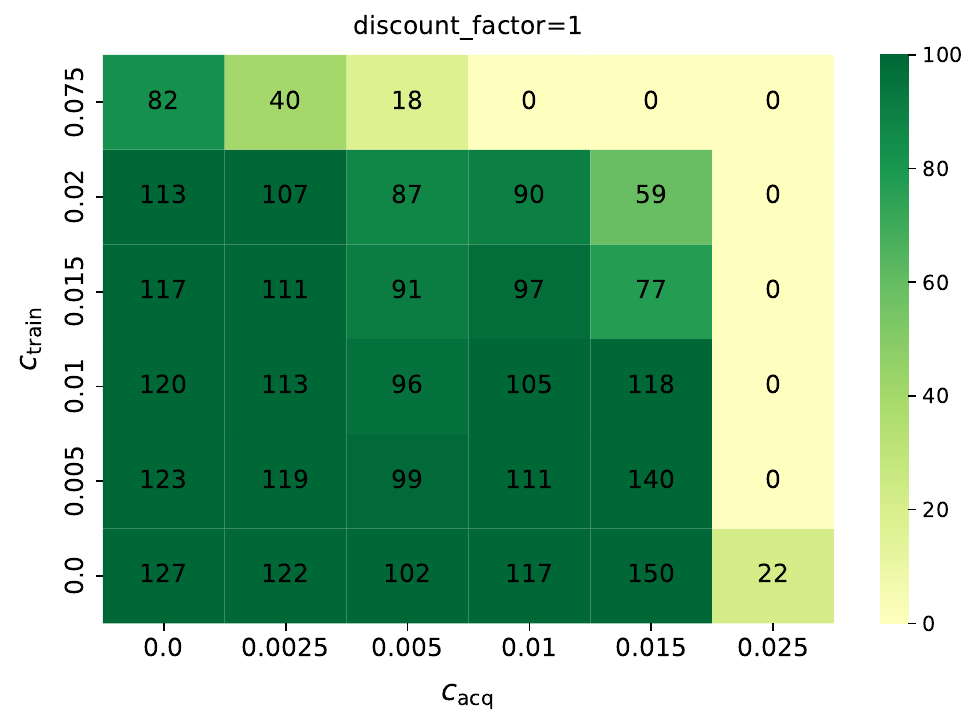}
    \caption{LR: $\tilde{\gamma} \approx 0.48$}
  \end{subfigure}
  \begin{subfigure}[b]{0.24\textwidth}
  \hspace{7pt}
    \includegraphics[trim={0 0 0 0.75cm},clip, width=0.9\linewidth]{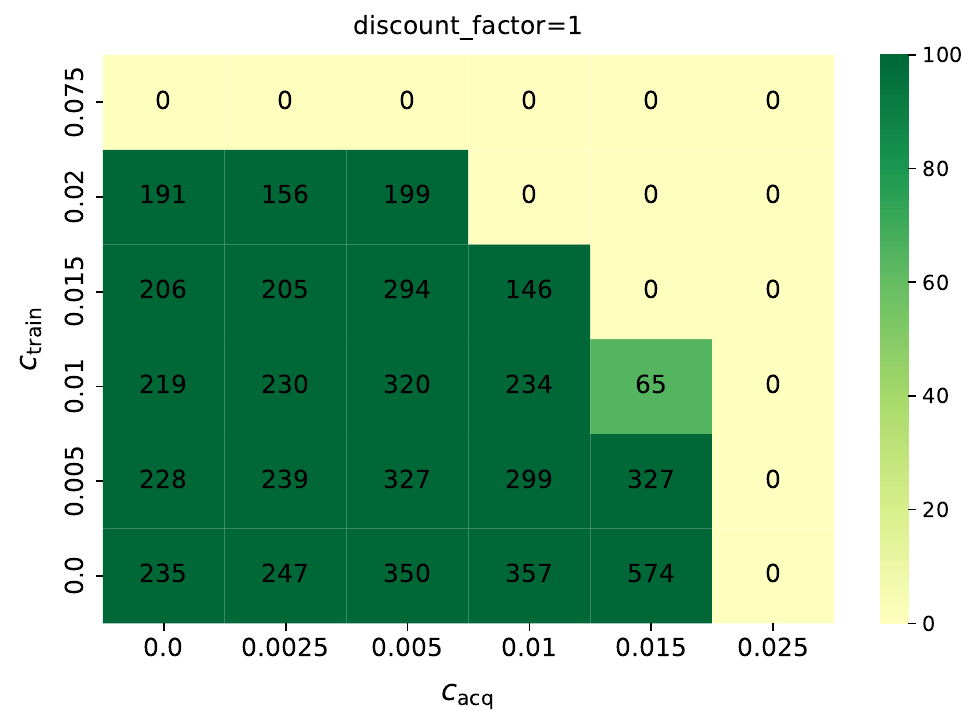}
    \caption{LightGBM: $\tilde{\gamma} \approx 0.48$}
  \end{subfigure}
  \caption{GSE's sensitivity to $\gamma$ - Wide range $\Gamma$}

  \footnotesize \parbox{0.95\linewidth}{\textit{Note.} Illustration of GSE’s sensitivity to the confidence parameter $\gamma$ across 30 sample paths over a grid of per-sample acquisition costs \(c_{\text{acq}}\), training costs \(c_{\text{train}}\), with \(c_s=0\) and \(\beta=1\), using LR (first column) and LightGBM (second column). Cell shading indicates maximum percentage performance gain $P$, as defined in Equation \ref{eq:P},\linebreak with \(\Gamma=\{44.68,25.76,19.2,12.01,7.69,5.6,4.32,1.92,0.96,0.48\}\).}
  \label{fig:GSE_sensitivity_wide}

\end{figure}

In both steps, we summarize the results using heatmaps (see Figure~\ref{fig:GSE_sensitivity_wide}). Each cell of a heatmap corresponds to a single experiment defined by values of \(c_{\text{acq}}\) and \(c_{\text{train}}\), with \(c_s=0\) and \(\beta\) fixed at a default value. The specific value of $\beta$ is reported in the subtitle of the figure or its description. 
Within each cell, we compute the sample-path–averaged performance for each \(\gamma \in \Gamma\), denoted \(\overline{V}^{\textsc{~sp}}_{\textsc{alg}}(T;\gamma)\), and report the maximum percentage performance gain $P$, defined as:
\begin{equation}
P = \max_{\gamma \in \Gamma}
\frac{\overline{V}^{\textsc{~sp}}_{\textsc{alg}}(T;\gamma)
- \overline{V}^{\textsc{~sp}}_{\textsc{alg}}(T;\tilde{\gamma})}
{\left|\overline{V}^{\textsc{~sp}}_{\textsc{alg}}(T;\tilde{\gamma})\right|}
\cdot 100, \label{eq:P}
\end{equation}
where \(\tilde{\gamma} \in \Gamma\) denotes a reference value. Thus, \(P\) measures the maximal percentage improvement in sample-path--averaged performance relative to the reference value \(\tilde{\gamma}\). Since \(\tilde{\gamma} \in \Gamma\), $P \ge 0$.

\begin{figure}[h!]
  \centering
  
    \begin{subfigure}[b]{0.24\textwidth}
    \hspace{7pt}
    \includegraphics[trim={0 0 0 0.75cm},clip, width=0.9\linewidth]{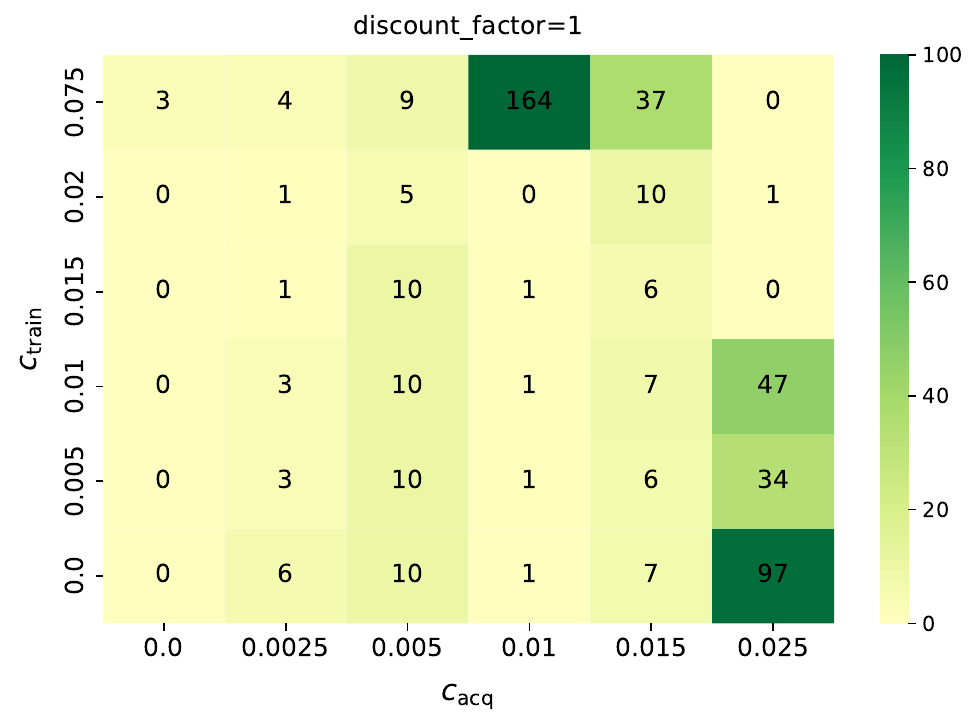}
    \caption{LR: $\beta=1$}
    \label{fig:placeholder}
  \end{subfigure}
  \hfill 
  \begin{subfigure}[b]{0.24\textwidth}
  \hspace{7pt}
    \includegraphics[trim={0 0 0 0.75cm},clip, width=0.9\linewidth]{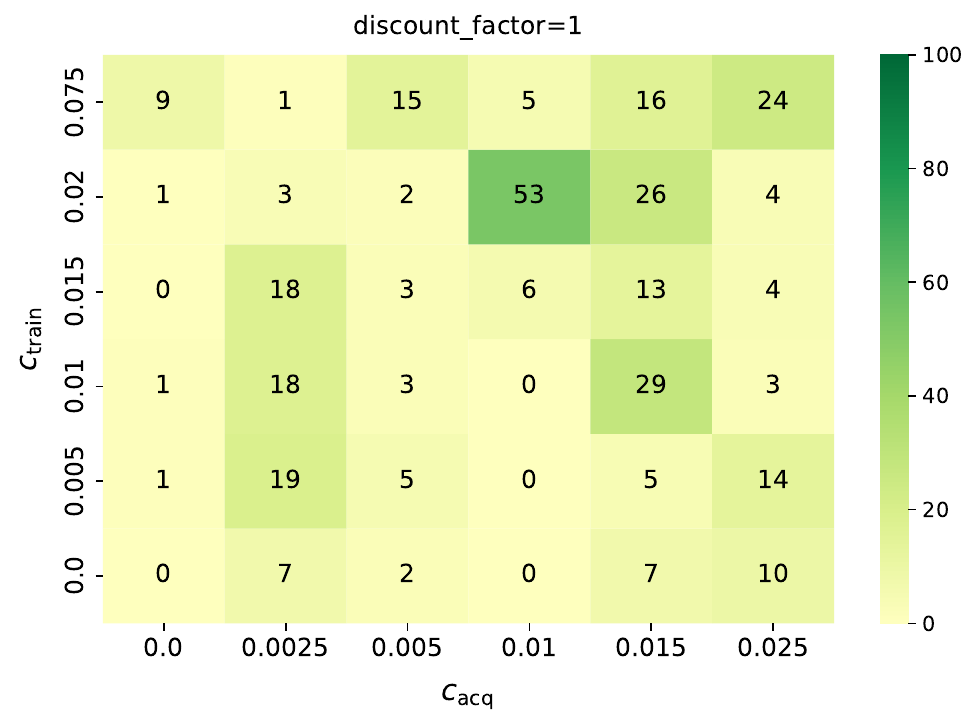}
    \caption{LightGBM: $\beta=1$}
    \label{fig:placeholder}
  \end{subfigure}
  \hfill 
    \begin{subfigure}[b]{0.24\textwidth}
    \hspace{7pt}
    \includegraphics[trim={0 0 0 0.75cm},clip, width=0.95\linewidth]{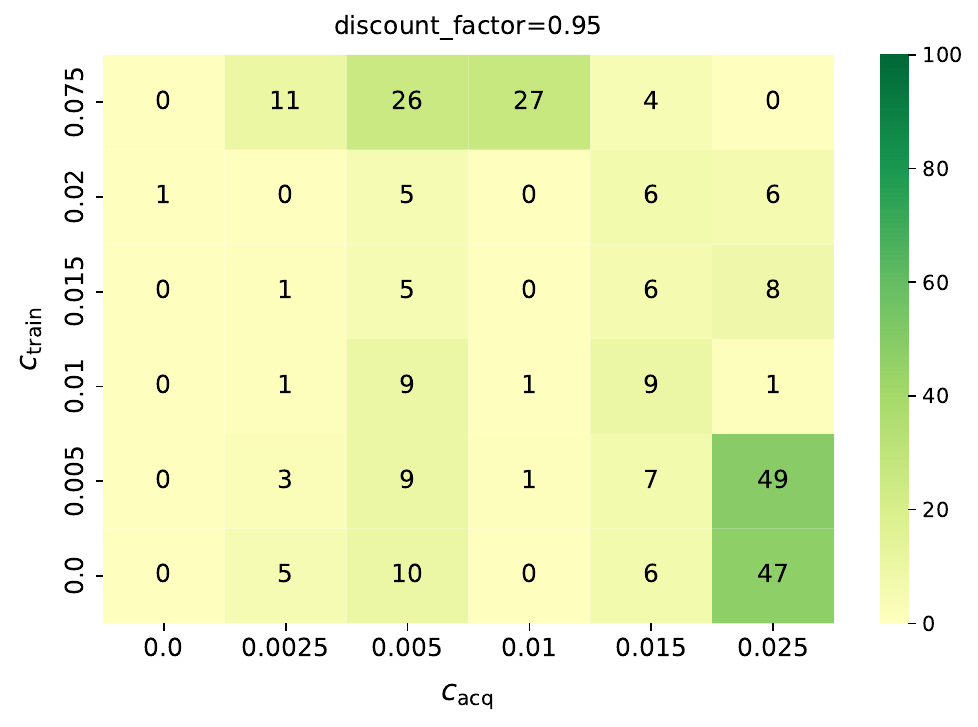}
    \caption{LR: $\beta=0.95$}
    \label{fig:placeholder}
  \end{subfigure}
  \hfill 
  \begin{subfigure}[b]{0.24\textwidth}
  \hspace{7pt}
    \includegraphics[trim={0 0 0 0.75cm},clip, width=1\linewidth]{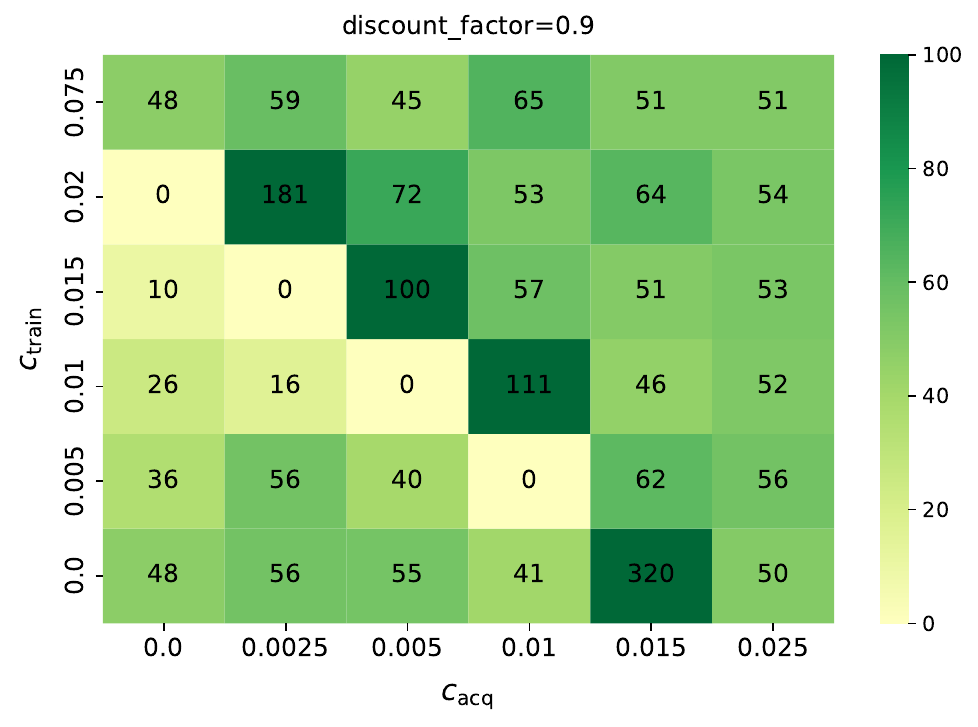}
    \caption{LightGBM: $\beta=0.9$}
    \label{fig:placeholder}
  \end{subfigure}
  \vspace{0.2cm}
  \caption{GSE's sensitivity to $\gamma$ - Close range $\Gamma$}
  \label{fig:GSE_sensitivity_close}

  \footnotesize \parbox{0.95\linewidth}{\textit{Note.} Illustration of GSE’s sensitivity to the confidence parameter $\gamma$ across 30 sample paths over a grid of per-sample acquisition costs \(c_{\text{acq}}\), training costs \(c_{\text{train}}\), and discount factor \(\beta\), with \(c_s=0\) and $\tilde{\gamma}=1.92$, using LR (first column) and LightGBM (second column). Cell shading indicates maximum percentage performance gain $P$, as defined in Equation \ref{eq:P}, with \(\Gamma=\{2.31,2.14,1.99,1.92,1.85,1.73,1.61\}\).}
\end{figure}

\paragraph{GSE results.} Overall, we find that GSE performance
is highly sensitive to $\gamma$, and no single value performs best across experiments. As shown in Figure~\ref{fig:GSE_sensitivity_wide}, there is no single value of \(\gamma\) that consistently delivers the close-to-best sample-path-averaged performance across experiments. As a single value of \(\gamma\) must be selected for the core scenarios, we choose \(\gamma=1.92\), which delivers the best performance in experiment E.1 among the selected values (see Figure \ref{fig:E.1_gamma}, column 2, row 1). We then examine local sensitivity around this reference value by fixing \(\tilde{\gamma}=1.92\). Heatmaps for \(\beta \in \{1,0.9\}\), with \(c_s=0\), are reported for both LR and LightGBM in Figure~\ref{fig:GSE_sensitivity_close}. Even in the vicinity of \(\gamma=1.92\), performance varies substantially, and \(\gamma=1.92\) is often far from delivering close-to-best sample-path--averaged performance, particularly when \(\beta=0.9\).

\begin{figure}[h!]
  \centering
  \begin{subfigure}[b]{0.24\textwidth}
  \hspace{7pt}
    \includegraphics[trim={0 0 0 0.75cm},clip, width=0.9\linewidth]{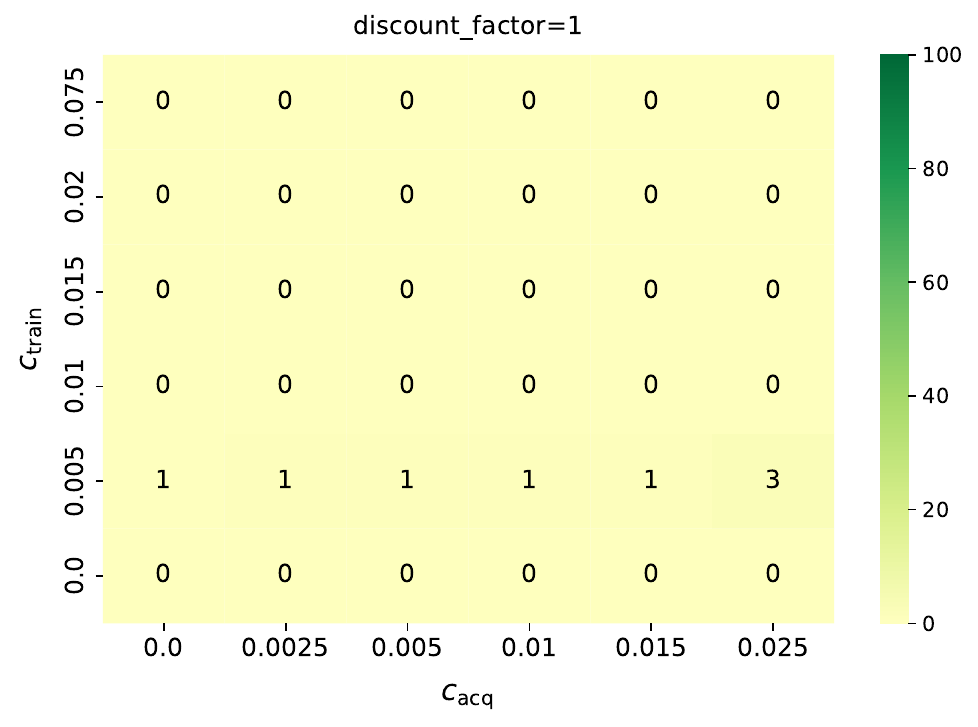}
    \caption{LR: $\tilde{\gamma}=0.0001$}
    \label{fig:placeholder}
  \end{subfigure}
  \begin{subfigure}[b]{0.24\textwidth}
  \hspace{7pt}
    \includegraphics[trim={0 0 0 0.75cm},clip, width=0.9\linewidth]{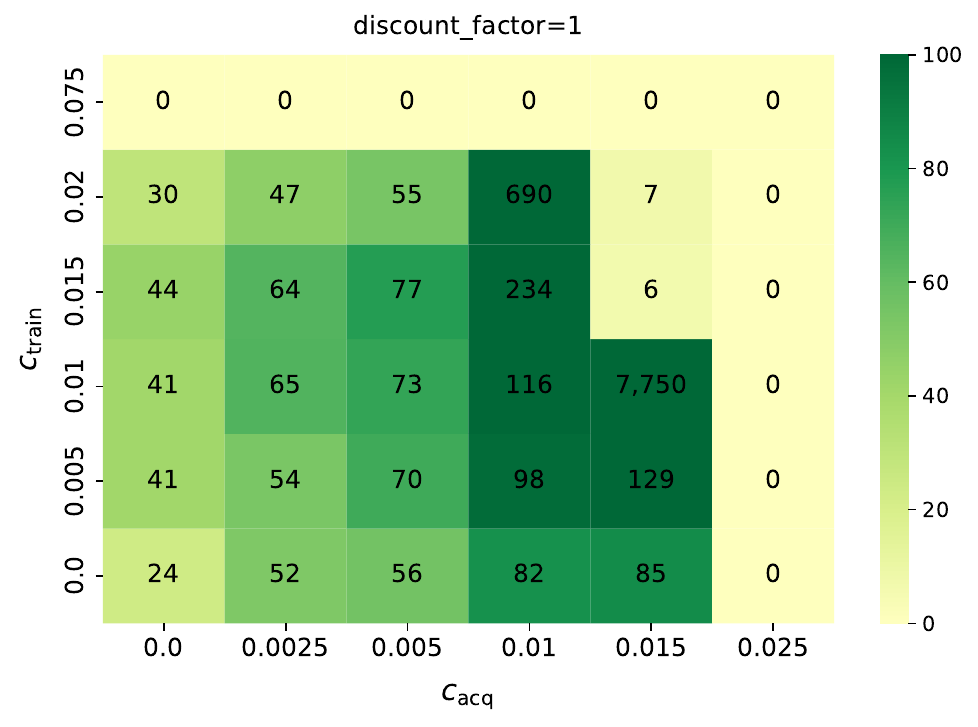}
    \caption{LightGBM: $\tilde{\gamma}=0.0001$}
    \label{fig:placeholder}
  \end{subfigure}
    \begin{subfigure}[b]{0.24\textwidth}
    \hspace{7pt}
    \includegraphics[trim={0 0 0 0.75cm},clip, width=0.9\linewidth]{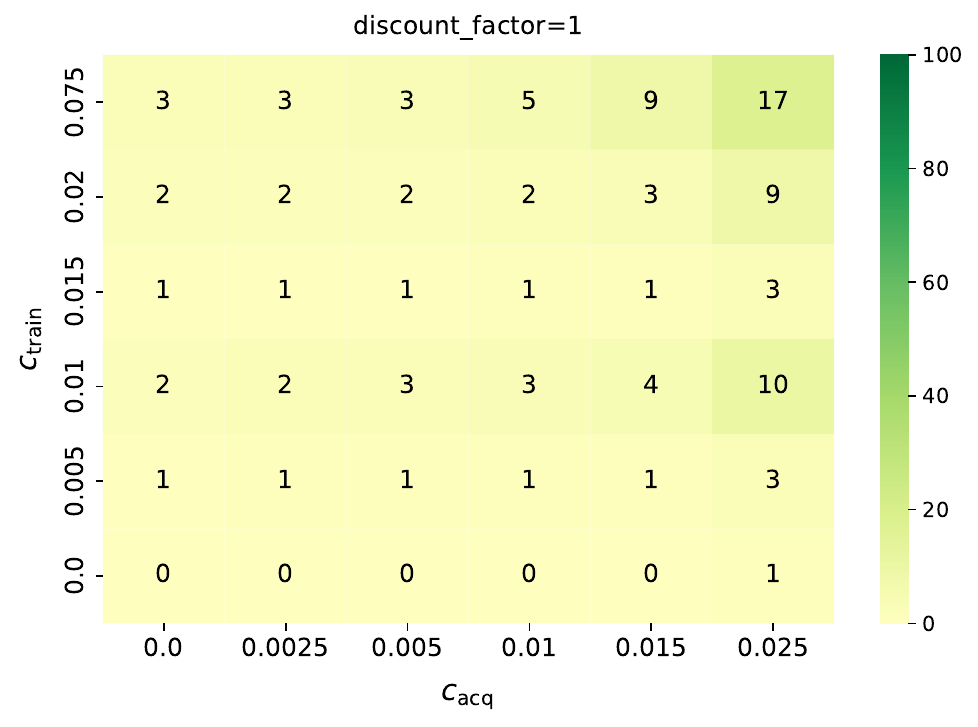}
    \caption{LR: $\tilde{\gamma}=0.05$}
    \label{fig:placeholder}
  \end{subfigure}
  \begin{subfigure}[b]{0.24\textwidth}
  \hspace{7pt}
    \includegraphics[trim={0 0 0 0.75cm},clip, width=0.9\linewidth]{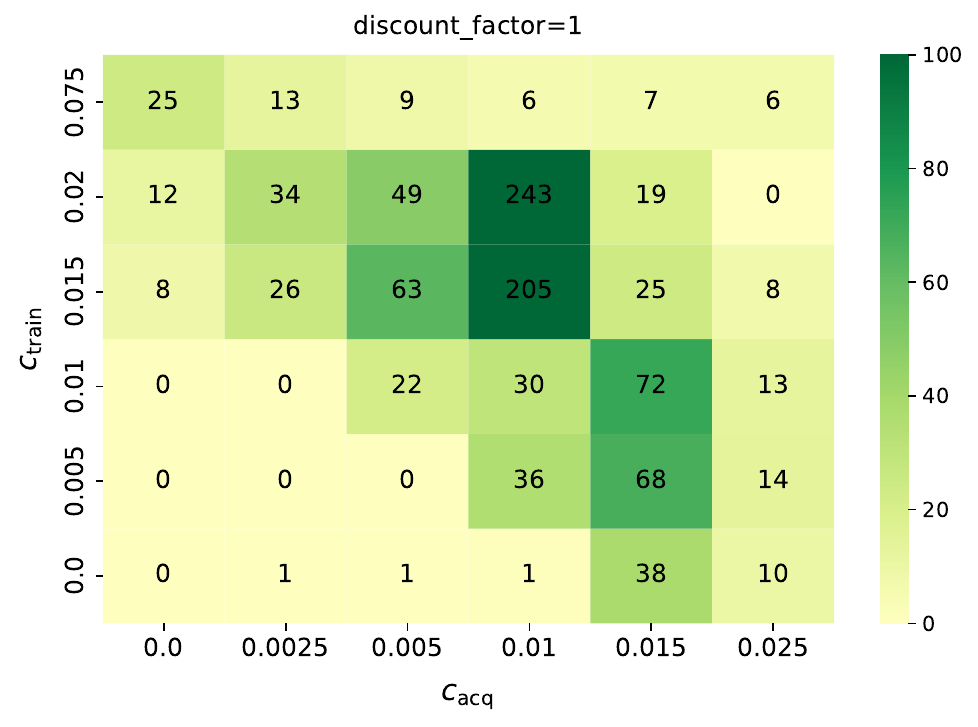}
    \caption{LightGBM: $\tilde{\gamma}=0.05$}
    \label{fig:placeholder}
  \end{subfigure}
    \begin{subfigure}[b]{0.24\textwidth}
    \hspace{7pt}
    \includegraphics[trim={0 0 0 0.75cm},clip, width=0.9\linewidth]{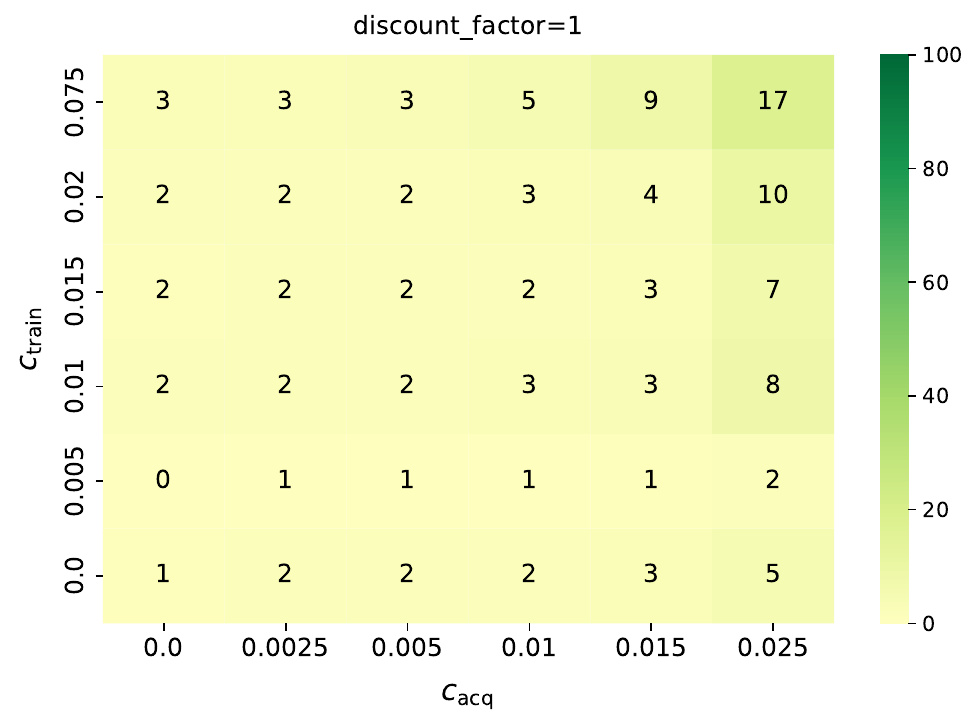}
    \caption{LR: $\tilde{\gamma}=0.1$}
    \label{fig:placeholder}
  \end{subfigure}
  \begin{subfigure}[b]{0.24\textwidth}
  \hspace{7pt}
    \includegraphics[trim={0 0 0 0.75cm},clip, width=0.9\linewidth]{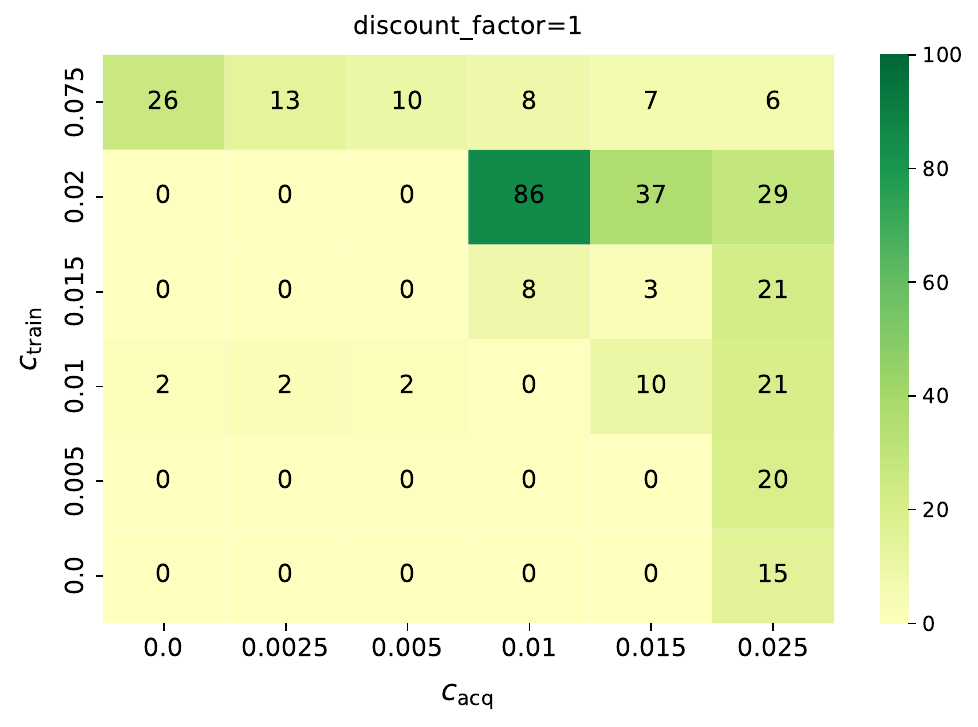}
    \caption{LightGBM: $\tilde{\gamma}=0.1$}
    \label{fig:placeholder}
  \end{subfigure} 
  \begin{subfigure}[b]{0.24\textwidth}
  \hspace{7pt}
    \includegraphics[trim={0 0 0 0.75cm},clip, width=0.9\linewidth]{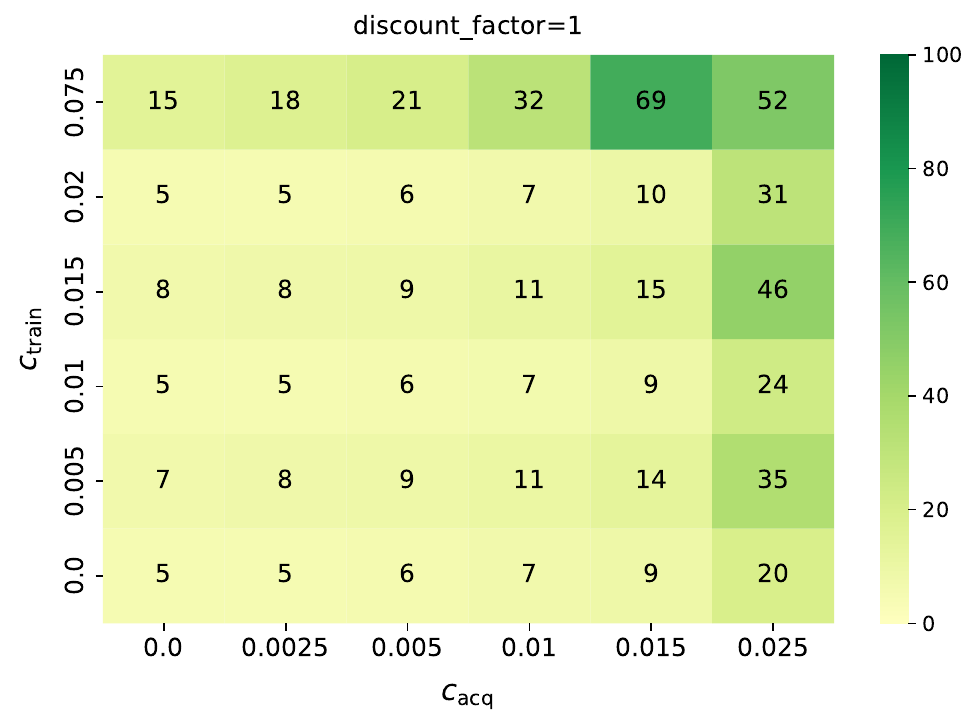}
    \caption{LR: $\tilde{\gamma}=0.2$}
    \label{fig:placeholder}
  \end{subfigure} 
  \begin{subfigure}[b]{0.24\textwidth}
  \hspace{7pt}
    \includegraphics[trim={0 0 0 0.75cm},clip, width=0.9\linewidth]{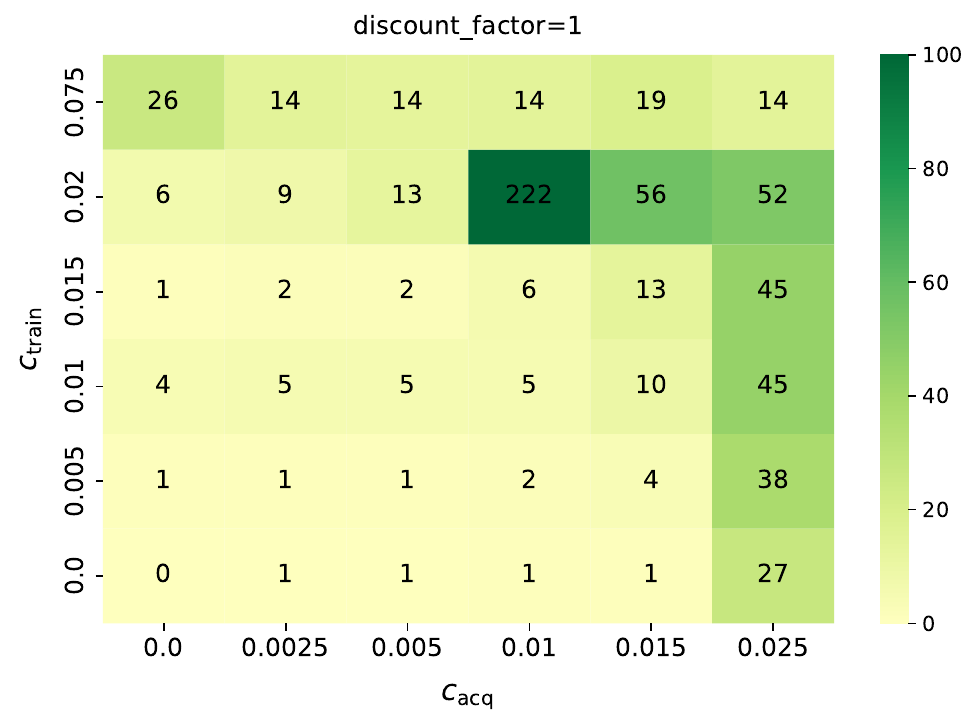}
    \caption{LightGBM: $\tilde{\gamma}=0.2$}
    \label{fig:placeholder}
  \end{subfigure}
  \caption{LSEc's sensitivity to $\gamma$ - Wide range $\Gamma$}
  \label{fig:LSEc_sensitivity_wide}

  \footnotesize \parbox{0.95\linewidth}{\textit{Note.} Illustration of LSEc’s sensitivity to the confidence parameter $\gamma$ across 30 sample paths over a grid of per-sample acquisition costs \(c_{\text{acq}}\), training costs \(c_{\text{train}}\), with \(c_s=0\) and \(\beta=1\), using LR (first column) and LightGBM (second column). Cell shading indicates maximum percentage performance gain $P$, as defined in Equation \ref{eq:P}, \linebreak with \(\Gamma=\{0.001,0.01,0.025,0.05,0.075,0.09,0.1,0.125,0.15,0.2\}\).}
\end{figure}

\begin{figure}
  \centering
    \begin{subfigure}[b]{0.24\textwidth}
    \hspace{7pt}
    \includegraphics[trim={0 0 0 0.75cm},clip, width=0.9\linewidth]{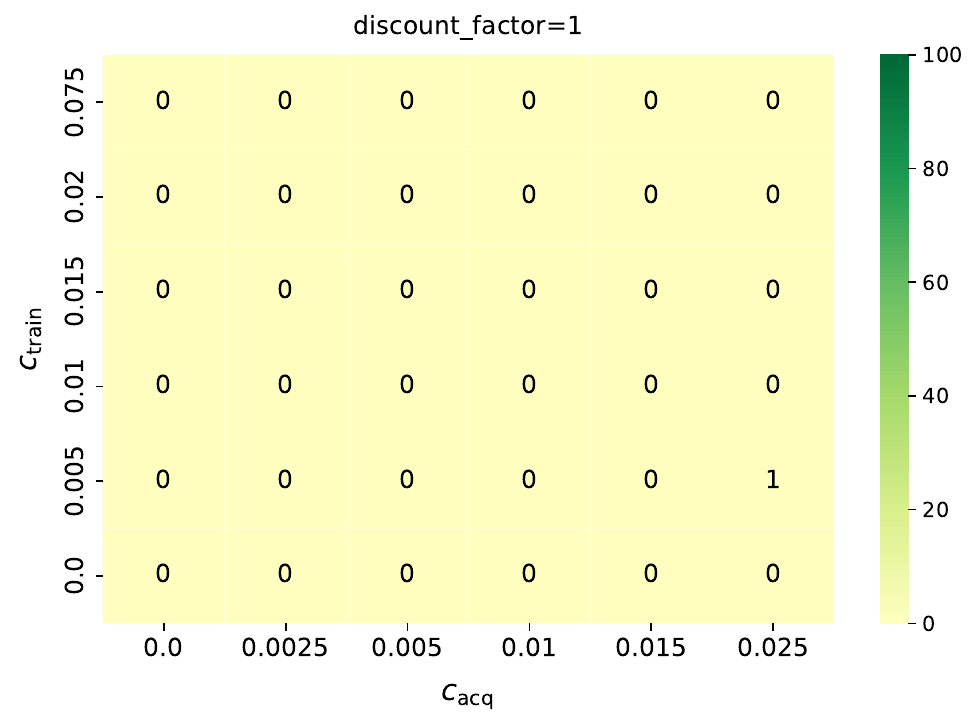}
    \caption{LR: $\beta=1$}
    \label{fig:placeholder}
  \end{subfigure}
  \begin{subfigure}[b]{0.24\textwidth}
  \hspace{7pt}
    \includegraphics[trim={0 0 0 0.75cm},clip, width=0.9\linewidth]{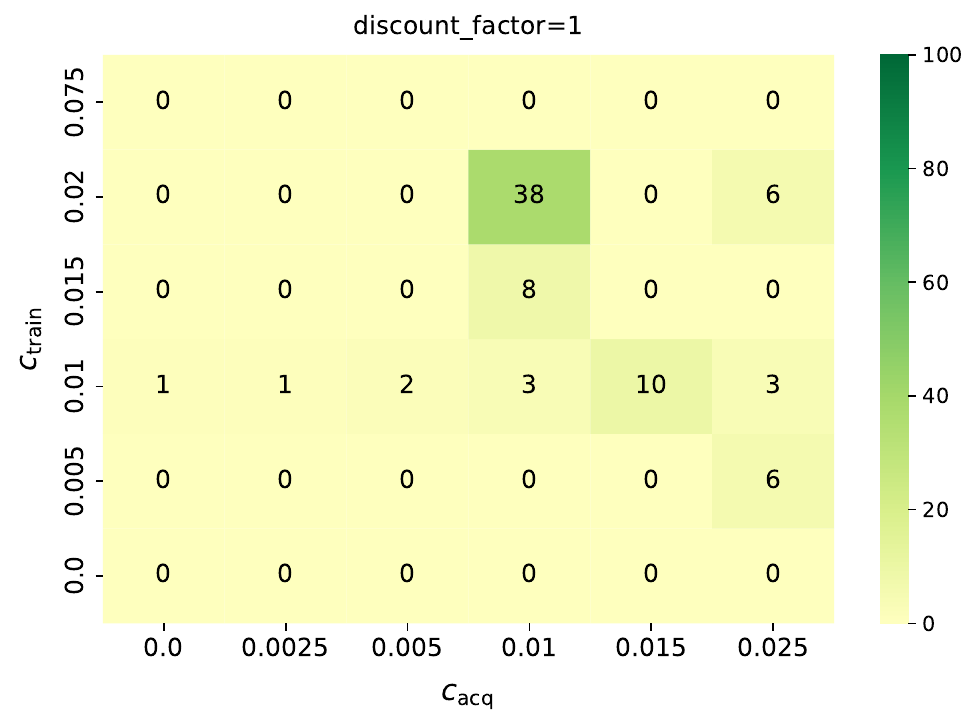}
    \caption{LightGBM: $\beta=1$}
    \label{fig:placeholder}
  \end{subfigure} 
  \begin{subfigure}[b]{0.24\textwidth}
  \hspace{7pt}
    \includegraphics[trim={0 0 0 0.75cm},clip, width=0.9\linewidth]{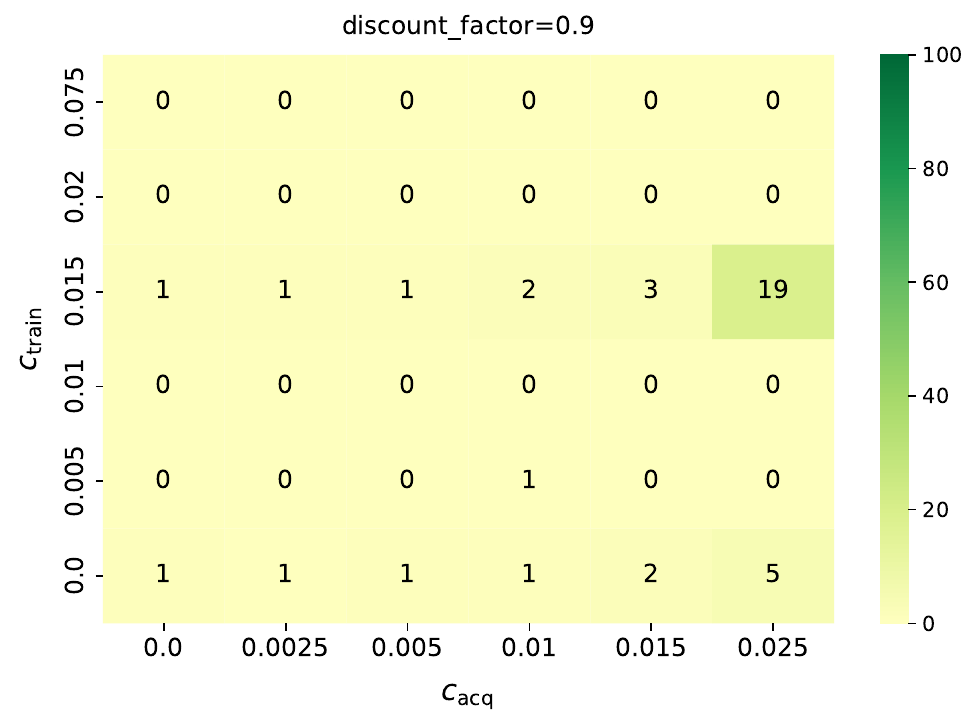}
    \caption{LR: $\beta=0.9$}
    \label{fig:placeholder}
  \end{subfigure}
  \begin{subfigure}[b]{0.24\textwidth}
  \hspace{7pt}
    \includegraphics[trim={0 0 0 0.75cm},clip, width=0.9\linewidth]{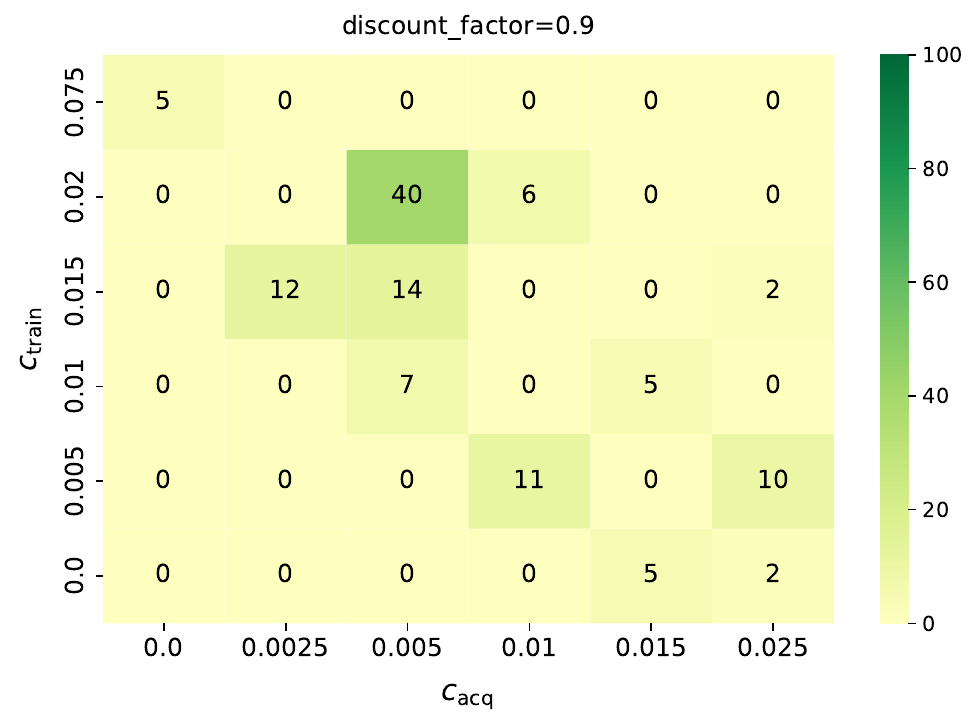}
    \caption{LightGBM: $\beta=0.9$}
    \label{fig:placeholder}
  \end{subfigure}
  \vspace{0.2cm}
  \caption{LSEc's sensitivity to $\gamma$ - Close range $\Gamma$}
  \label{fig:LSEc_sensitivity_close}

  \footnotesize \parbox{0.95\linewidth}{\textit{Note.} Illustration of LSEc’s sensitivity to the confidence parameter $\gamma$ across 30 sample paths over a grid of per-sample acquisition costs \(c_{\text{acq}}\), training costs \(c_{\text{train}}\), and discount factor \(\beta\), with \(c_s=0\) and $\tilde{\gamma}=0.1$, using LR (first column) and LightGBM (second column). Cell shading indicates maximum percentage performance gain $P$, as defined in Equation \ref{eq:P}, with \(\Gamma=\{0.0875,0.095,0.0975,0.1,0.1025,0.105,0.1125\}\).}
\end{figure}

\paragraph{LSEc results.}
In contrast, LSEc is markedly more robust to the choice of \(\gamma\). As shown in Figure~\ref{fig:LSEc_sensitivity_wide}, \(\gamma=0.1\) consistently delivers close-to-best sample-path--averaged performance across experiments, particularly under LR. Accordingly, we set \(\gamma=0.1\) for the core scenarios and other experiments. Local sensitivity analysis using a tighter grid and \(\beta \in \{1,0.9\}\), reported in Figure~\ref{fig:LSEc_sensitivity_close}, shows limited performance variation in the vicinity of \(\gamma=0.1\). Overall, LSEc performance is substantially less sensitive to \(\gamma\) than GSE, and \(\gamma=0.1\) remains close to optimal across experiments.

\subsection{Robustness Checks: Numerical Results Beyond the Core Scenarios} \label{sec:add_exp} \label{ssec:scenario_selection}
The two core scenarios analyzed in Section \ref{sec:result_core_scenario} provide a baseline understanding of how the algorithms behave when the optimal stopping time occurs early or late in the sequential process. We now turn to a set of additional experiments with two objectives. First, we study how algorithmic behaviors vary under alternative economic structures. Second, we assess the robustness of our findings to changes in the experimental design, including batch sizes, the role of time ordering, and the total sample size.
Table~\ref{exp_params} reports the parameters for all scenarios.
\begin{table}[h!]
\centering
\resizebox{0.8\textwidth}{!}{
\begin{tabular}{lcc|ccc|cccc}
\hline
 & \multicolumn{2}{c}{\textbf{Core scenario}} & \multicolumn{3}{c}{\textbf{Economic structure}} & \multicolumn{4}{c}{\textbf{Robustness}} \\ \cline{2-10 }
\textbf{Parameter} & \textbf{Early (E.1)} & \textbf{Late (E.2)} & \textbf{No costs (E.3)} & \textbf{High $c_{\text{acq}}$ (E.4)} & \textbf{$c_{S} > 0$ (E.5)} & \textbf{Small batch (E.6)} & \textbf{Large batch/cost (E.7)} & \textbf{Timeless (E.8)} & \textbf{Small size (E.9)}\\
\hline
Batch type & (\texttt{geometric}, 2) & -& - & - & - & (\texttt{geometric}, \textbf{1.5})  & (\texttt{geometric}, \textbf{2.5}) & - & - \\
Time series & Yes & - & - & - & - & - & - & No & - \\
$N_0$ & 189,250 & - & - & - & - & - & - & - & - \\
$N_{t_e}$ & 163,400 & - & - & -& -  & - & - & - & 81,700 \\
$N_1$ & 250 & - & - & - & - & - & - & - & - \\ \hline
Train-test ratio & 1 & - & - & - & - & - & - & - & -\\
Sampling ratio & 0.5 & - & - & - & - & - & - & - & -\\ 
\hline
$c_{\text{acq}}$ & 0.0025 & -  & \textbf{0}  & \textbf{0.025} & - & - & \textbf{0.025}  & -  & - \\
$\beta$ & 0.95 & - & - & - & - & - & - & -  & -\\
$C_s$ & 0 & - & - & - & 250 & - & - & - & - \\
$c_{\text{train}}$ & 0.075 & 0.005 & \textbf{0} & 0.075/0.005 & 0.075/0.005 & - & 0.075/0.005 & -  & - \\
\hline
$\gamma$ (GSE) & 1.92 & - & - & - & - & - & - & - & -\\
$\gamma$ (LSEc) & 0.1 & -  & - & - & - & - & - & - & -\\  
\hline
Metric & AUC & - & - & - & - & - & - & - & - \\
\hline
Model & LR & LightGBM & Both & Both & Both & Both & Both & Both & Both \\
\hline
Model re-estimation & - & No & No & No & No & Yes & Yes & Yes & Yes \\
\hline
\end{tabular}}
\caption{Parameters of all experiments} \label{exp_params}
    
  \footnotesize \parbox{0.95\linewidth}{\textit{Note.} The table reports the parameters for all experiments, grouped into core scenarios, economic structure, and robustness. Parameters cover epoch design, costs, algorithms, models, and performance. A dash indicates that the parameter value coincides with experiment E.1. When a cell lists two values separated by a slash, the experiment is run using LR with the first value and LightGBM with the second. Experiments under economic structure and robustness are reported in Appendix~\ref{sec:add_exp}.}
\end{table}

\begin{figure}[ht!]
  \centering
  \begin{subfigure}[b]{0.3\textwidth}
  \hspace{-14pt}
    \includegraphics[width=1\textwidth]{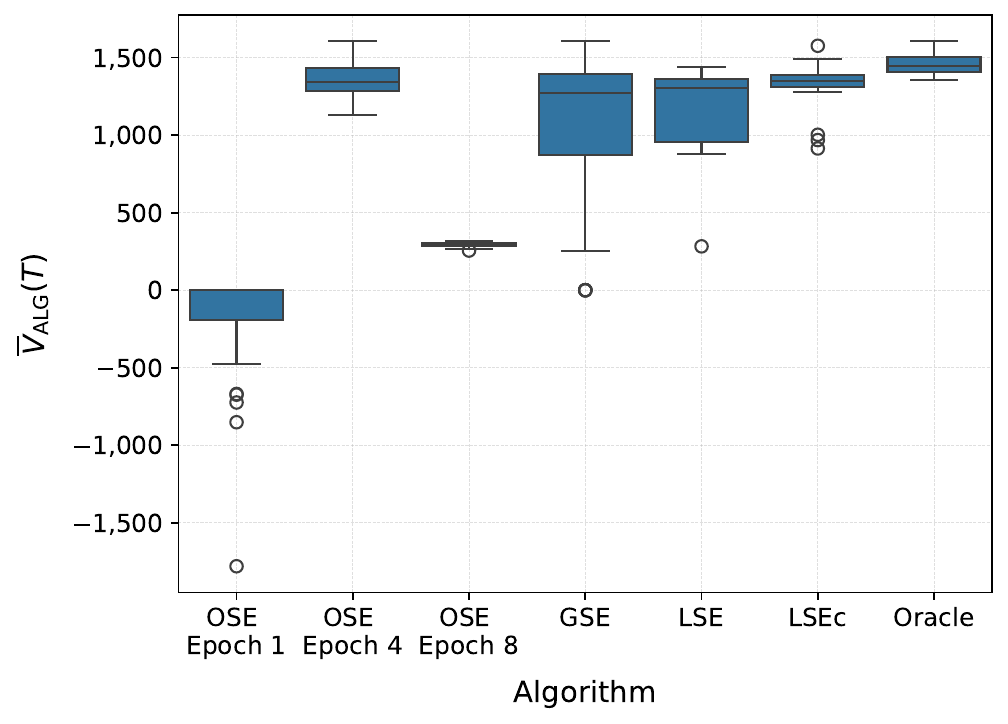}
    \caption{E.3: No costs - LR}
    \label{fig:no_costs_LR}
  \end{subfigure}
  \hfill
  \begin{subfigure}[b]{0.3\textwidth}
  \hspace{-14pt}
    \includegraphics[width=1\textwidth]{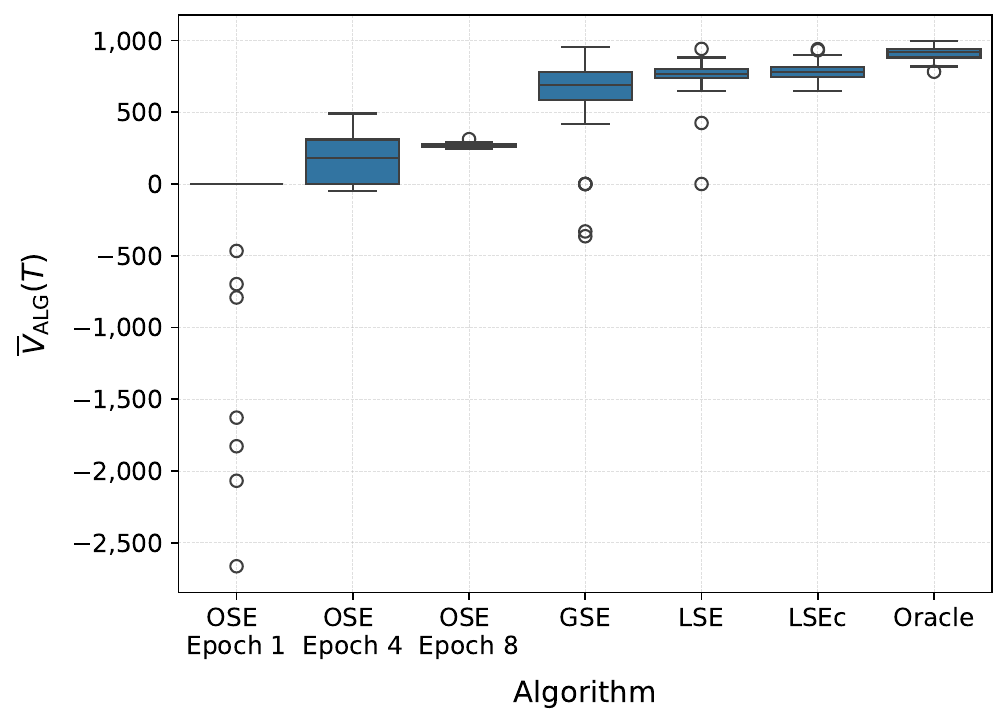}
    \caption{E.3: No costs - LightGBM}
    \label{fig:sub1}
  \end{subfigure}
  \hfill
  \begin{subfigure}[b]{0.3\textwidth}
  \hspace{-14pt}
    \includegraphics[width=1\textwidth]{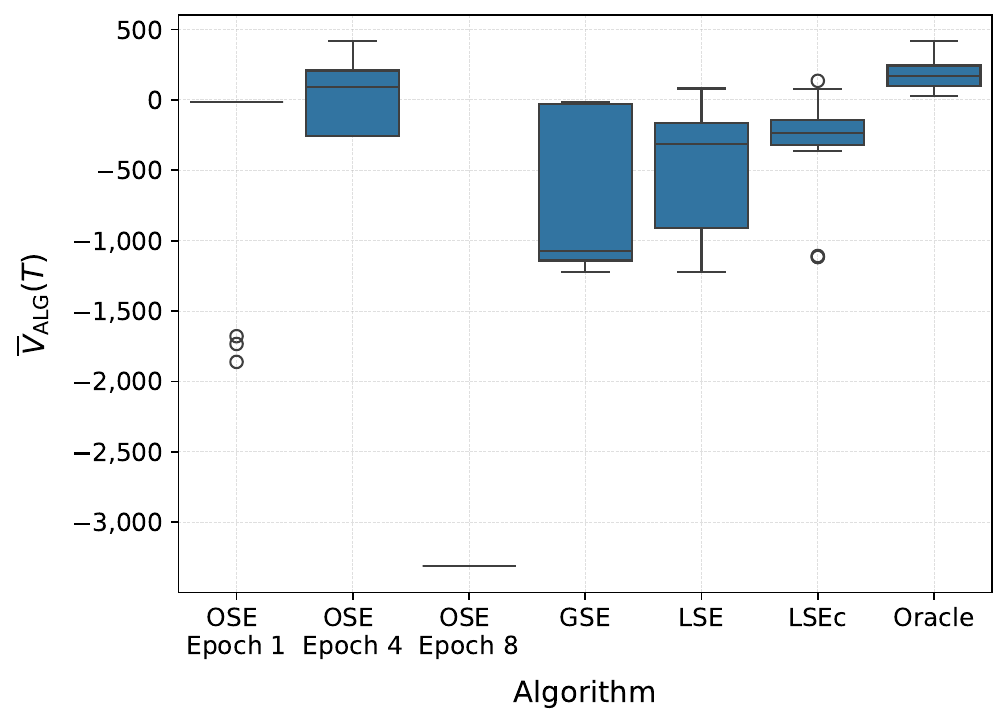}
    \caption{E.4: High $c_{\text{acq}}$ - LR}
    \label{fig:sub1}
  \end{subfigure}
  \hfill
  \begin{subfigure}[b]{0.3\textwidth}
  \hspace{-14pt}
    \includegraphics[width=1\textwidth]{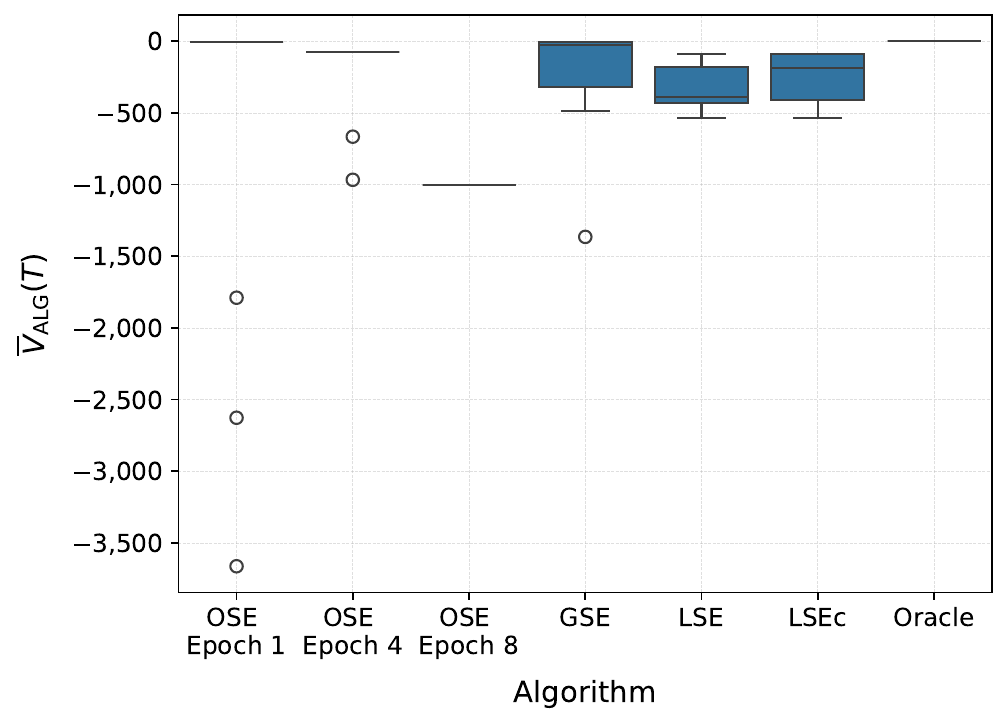}
    \caption{E.4: High $c_{\text{acq}}$ - LightGBM}
    \label{fig:sub1}
  \end{subfigure}
  \hfill
  \begin{subfigure}[b]{0.3\textwidth}
  \hspace{-14pt}
    \includegraphics[width=1\textwidth]{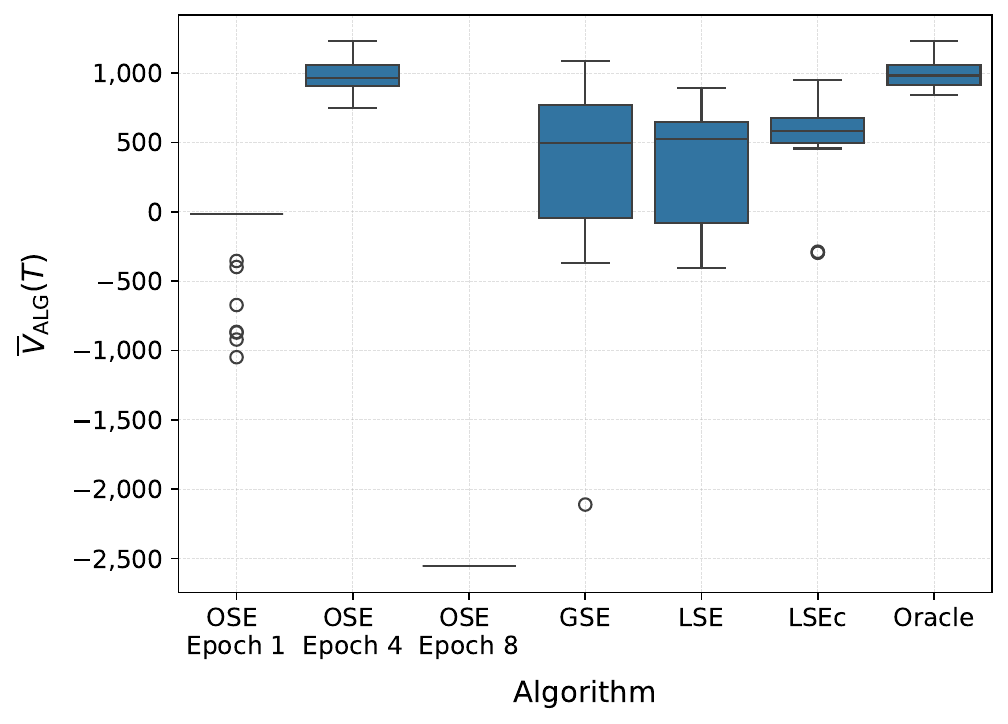}
    \caption{E.5: $c_S>0$ - LR}
    \label{fig:sub1}
  \end{subfigure}
  \hfill
  \begin{subfigure}[b]{0.3\textwidth}
  \hspace{-14pt}
    \includegraphics[width=1\textwidth]{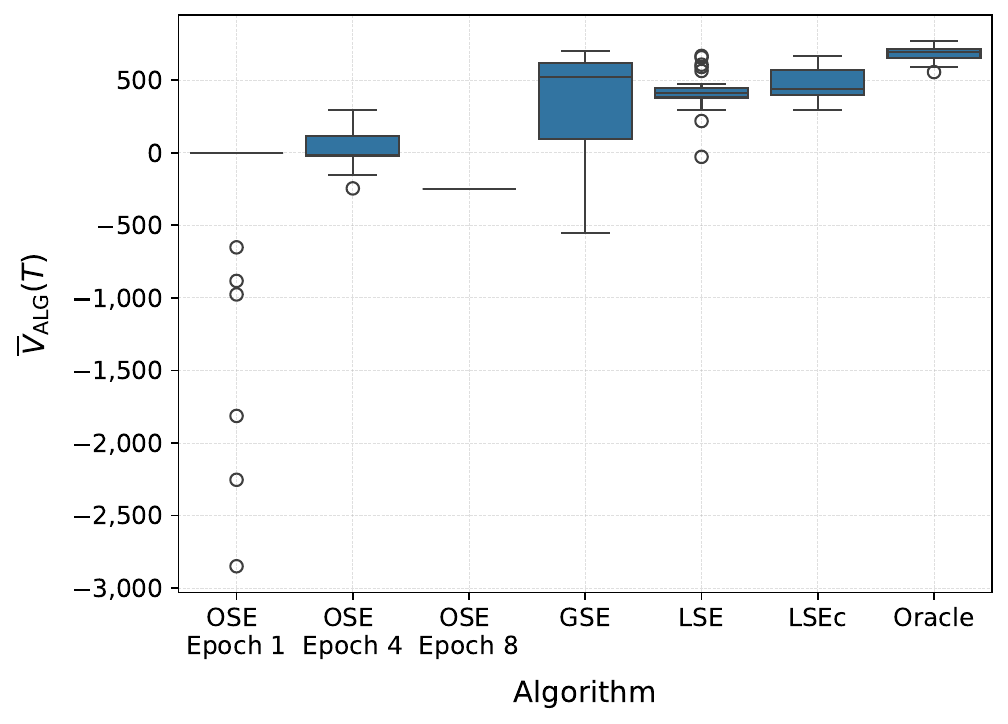}
    \caption{E.5: $c_S>0$ - LightGBM}
  \end{subfigure} 
  \\
  \vspace{0.2cm}
  \caption{Economic structure} \label{fig:cost_structure}

  \footnotesize \parbox{0.95\linewidth}{\textit{Note.} This figure displays the performance of our algorithms across the 30 sample paths for experiments E.3 to E.5.}
\end{figure}

\paragraph{Economic Structure} We examine how algorithmic behavior changes under alternative economic structures—(i) no costs, (ii) high acquisition costs, and (iii) positive switching costs—each evaluated for both the LR and the LightGBM. The full specification of these experiments, denoted E.3, E.4, and E.5, is reported in Table~\ref{exp_params}. We display the results of these experiments in Figure \ref{fig:cost_structure}.

\begin{figure}[h]\centering 
\begin{minipage}{0.90\textwidth}
  \centering
  \begin{subfigure}[b]{0.45\textwidth}
  \hspace{-13pt}
    \includegraphics[width=1\textwidth]{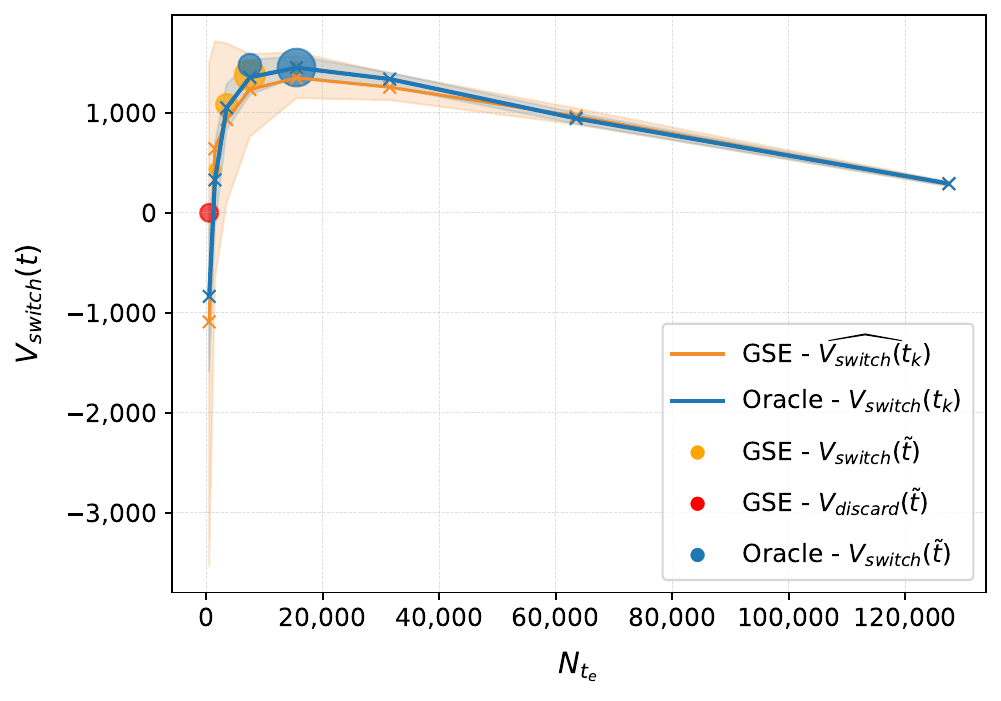}
    \caption{LR}
    \label{fig:E_9_LR} 
  \end{subfigure} 
  \hfill
  \begin{subfigure}[b]{0.45\textwidth}
  \hspace{-13pt}
    \includegraphics[width=1\textwidth]{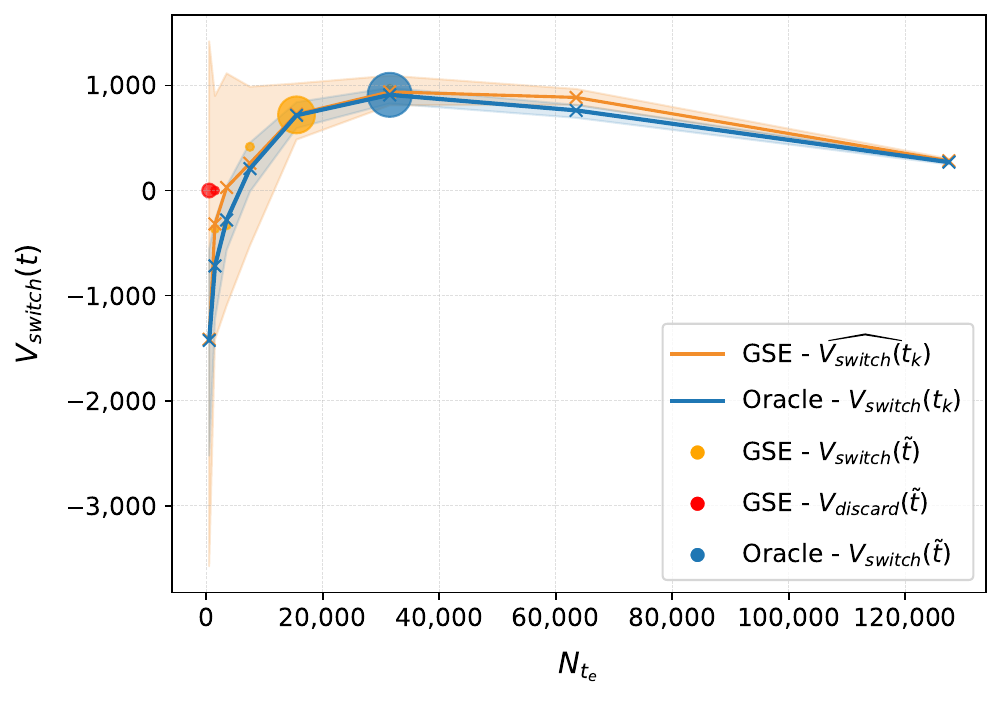}
    \caption{LightGBM}
    \label{fig:E_9_LR} 
  \end{subfigure} 
  \end{minipage}
  \caption{Economic structure - E.3: No costs. The plot conventions are the same as in Figure \ref{fig:E_1_OSE}.}
  \label{fig:robustness_E_3_costs}

\end{figure}

Under the no-costs scenario, several results stand out. First, the OSE algorithm becomes profitable when acquiring the entire dataset at once. Although profitable, it remains dominated by the other algorithms, highlighting that, even in the absence of costs, sequential decision rules can deliver higher value than a one-shot strategy buying the entire dataset at once. Second, the GSE algorithm exhibits substantially lower performance's volatility in both the early- and late-switch scenarios when costs are set to zero. While removing costs does not affect statistical uncertainty, it fundamentally alters the shape of the utility curves. In particular, eliminating acquisition costs shifts the utility curves upward, and eliminating training costs makes them flatter across epochs (see Figure \ref{fig:robustness_E_3_costs}). As a result, although the GSE algorithm remains sensitive to early-stage noise, the value differences across epochs are smaller, which reduces dispersion in stopping decisions and, consequently, lowers performance volatility. Third, the LSEc algorithm delivers the highest performance, with values that remain very close to the oracle’s. Its performance is notably closer to the oracle in E.3 than in E.1 and E.2, reflecting the fact that setting the training cost to zero removes an important source of divergence between LSEc and the oracle's behavior.

Under the high-acquisition-cost scenario, we examine whether the algorithms make appropriate switching or discarding decisions in settings where acquisition costs substantially reduce economic value and can even render switching unprofitable. Recall that acquisition costs affect the decision to switch or discard but do not influence the optimal stopping point. We distinguish between two cases, depending on the challenger model. When the challenger is a LR, the oracle’s optimal decision is to switch for all sample paths, albeit with a value close to zero. This case allows us to assess whether the algorithms can correctly switch despite limited economic upside. In contrast, when the challenger is a LightGBM, the oracle’s optimal decision is to discard the challenger for all sample paths. This case enables us to evaluate whether the algorithms appropriately avoid switching when doing so would be unprofitable. In both cases, the LSEc algorithm switches or discards sufficiently early to avoid substantial losses, particularly relative to the GSE and LSE algorithms, with this advantage being most pronounced when the challenger is a LR.\footnote{LSEc always switches in the LR case and always discards the challenger in the LightGBM case. Although LSEc switches at a time close to the oracle on average (4.87 versus 4.03), switching yields a lower and negative value for LSEc. This gap arises because even small delays in switching can be costly when gaps decline rapidly and because LSEc incurs cumulative training costs prior to switching, whereas the oracle pays the training cost only at the switching epoch. Given the high training costs in the LR case, the latter effect is likely the dominant driver.}

Under the final scenario, we analyze how introducing a switching cost affects algorithm performance. Recall that, like the acquisition cost, the switching cost does not affect the optimal stopping time but instead influences whether the algorithm chooses to switch or to discard the challenger. Consequently, as long as the switching cost does not change the optimal decision, the value of any algorithm that switches should decrease mechanically by the amount of the switching cost. This is exactly what we observe for all our algorithms. The only exception arises in the LR case, where introducing the switching cost leads the GSE algorithm to discard the challenger in one additional sample path relative to the zero–switching-cost setting.

\begin{figure}[h!]
  \centering
  \begin{subfigure}[b]{0.24\textwidth}
  \hspace{-14pt}
    \includegraphics[width=1\textwidth]{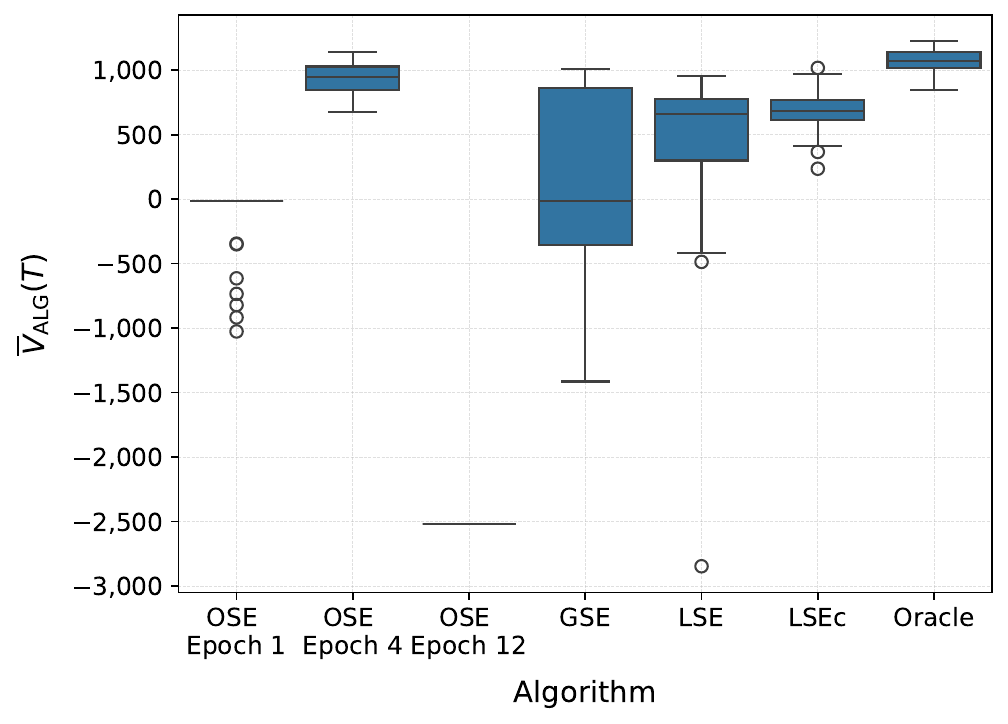}
    \caption{E.6: Small batch}
    \label{fig:E_6_LR} 
  \end{subfigure} 
  \hfill 
  \begin{subfigure}[b]{0.24\textwidth}
  \hspace{-14pt}
    \includegraphics[width=1\textwidth]{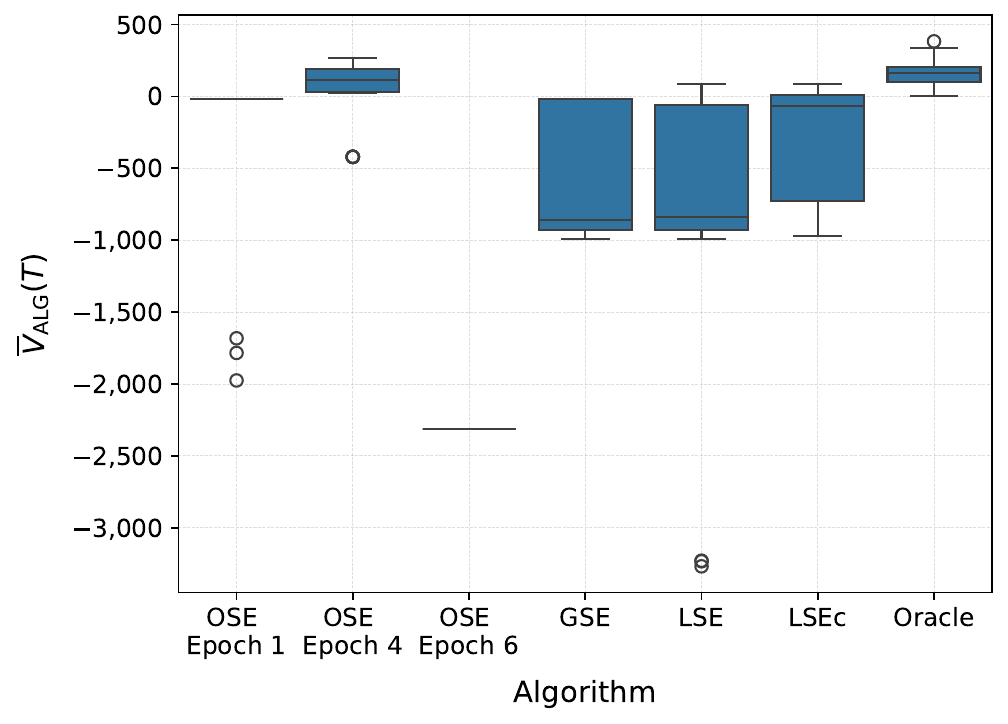}
    \caption{E.7: High $c_{\text{acq}}$ \& batch}
    \label{fig:E_7_LR}
  \end{subfigure} 
  \begin{subfigure}[b]{0.24\textwidth}
  \hspace{-14pt}
    \includegraphics[width=1\textwidth]{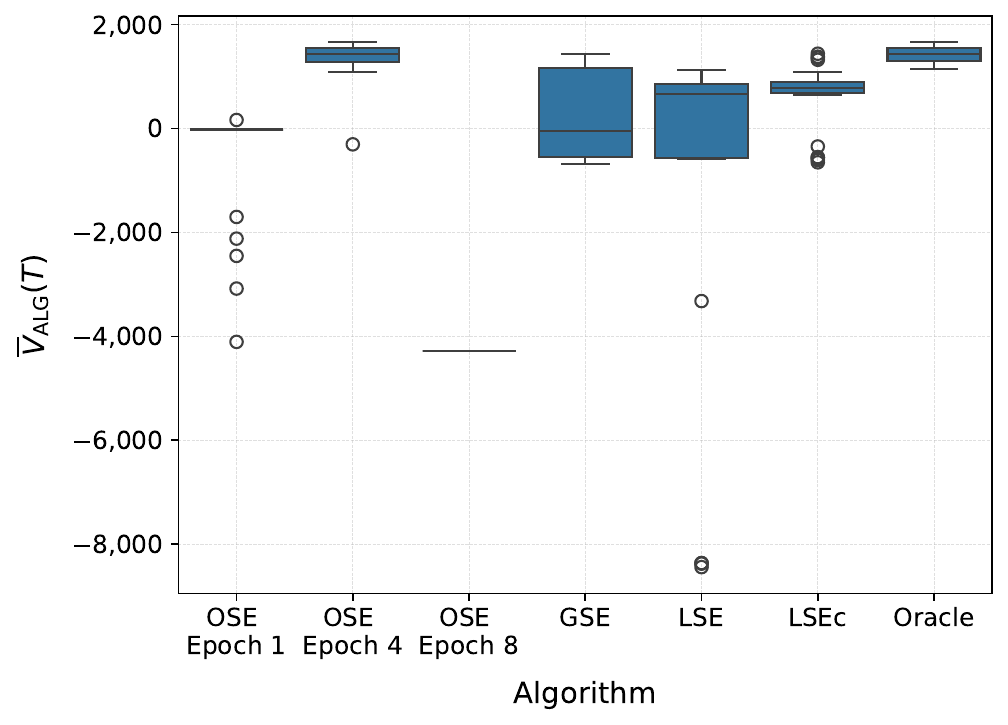}
    \caption{E.8: Timeless}
    \label{fig:E_8_LR}
  \end{subfigure}
  \hfill 
  \begin{subfigure}[b]{0.24\textwidth}
  \hspace{-14pt}
    \includegraphics[width=1\textwidth]{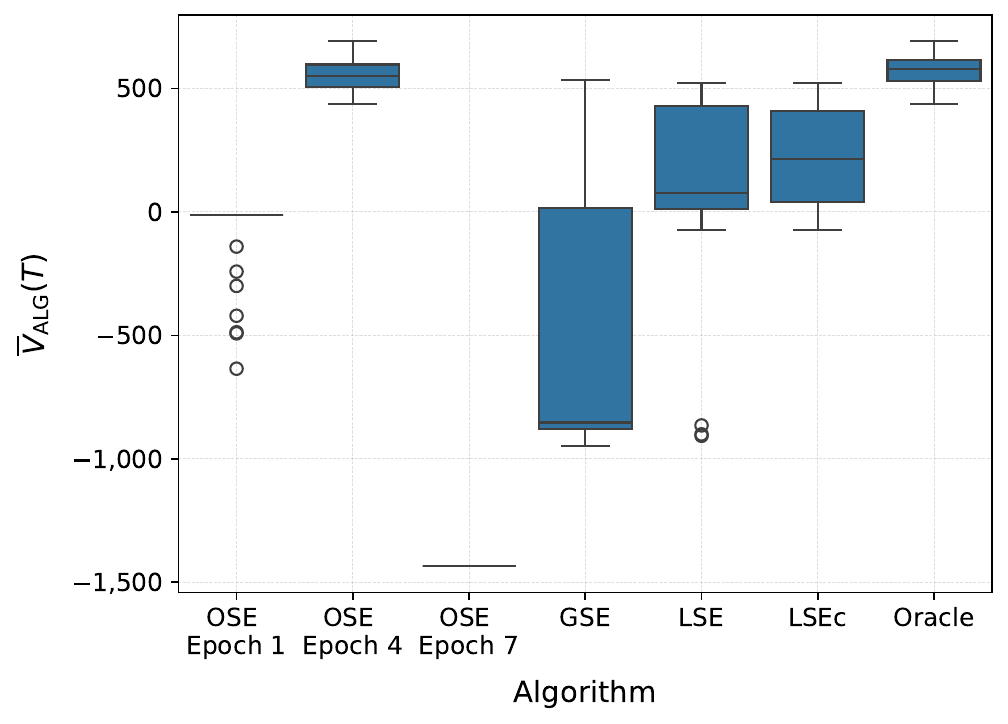}
    \caption{E.9: Small size}
    \label{fig:E_9_LR}
  \end{subfigure}
  \begin{subfigure}[b]{0.24\textwidth}
  \hspace{-14pt}
    \includegraphics[width=1\textwidth]{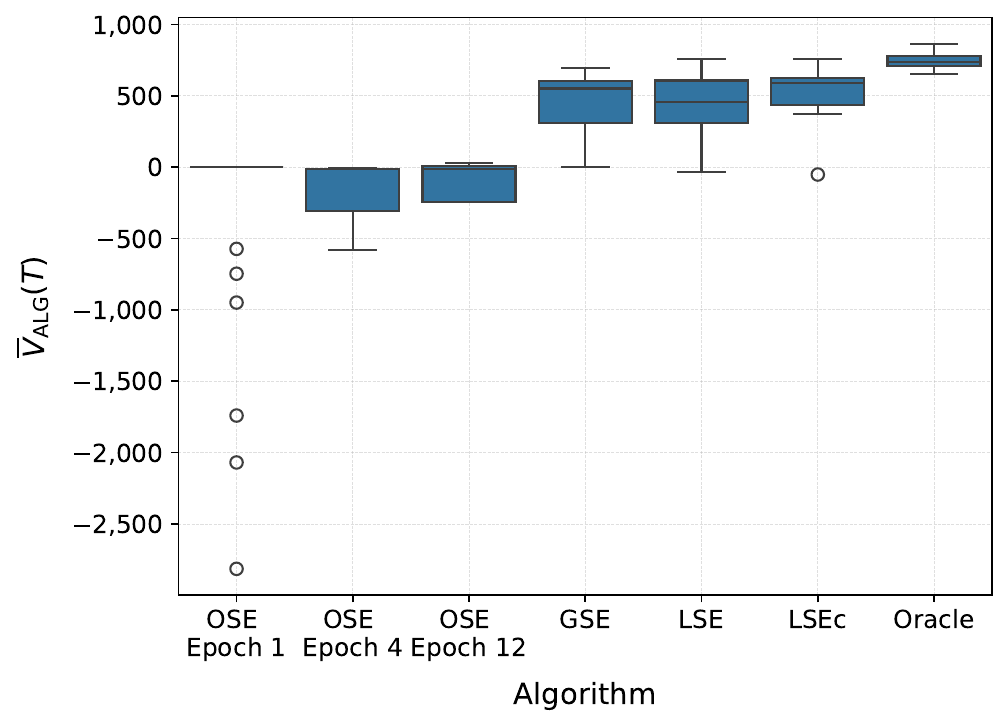}
    \caption{E.6: Small batch}
    \label{fig:E_6_LightGBM} 
  \end{subfigure} 
  \hfill 
  \begin{subfigure}[b]{0.24\textwidth}
  \hspace{-14pt}
    \includegraphics[width=1\textwidth]{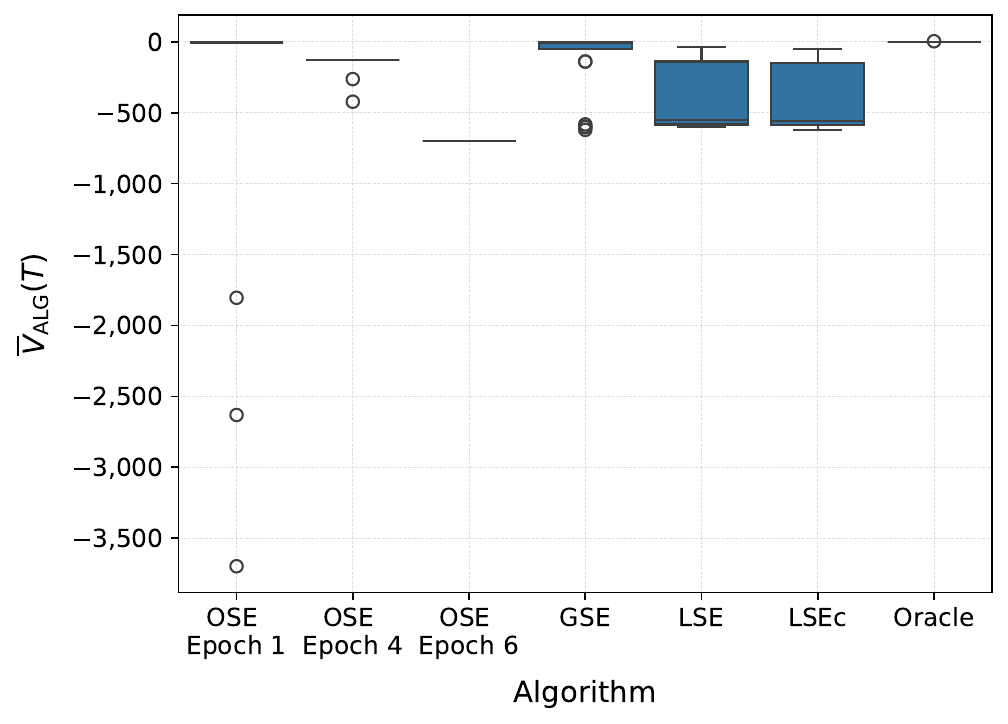}
    \caption{E.7: High $c_{\text{acq}}$ \& batch}
    \label{fig:E_7_LightGBM}
  \end{subfigure} 
  \begin{subfigure}[b]{0.24\textwidth}
  \hspace{-14pt}
    \includegraphics[width=1\textwidth]{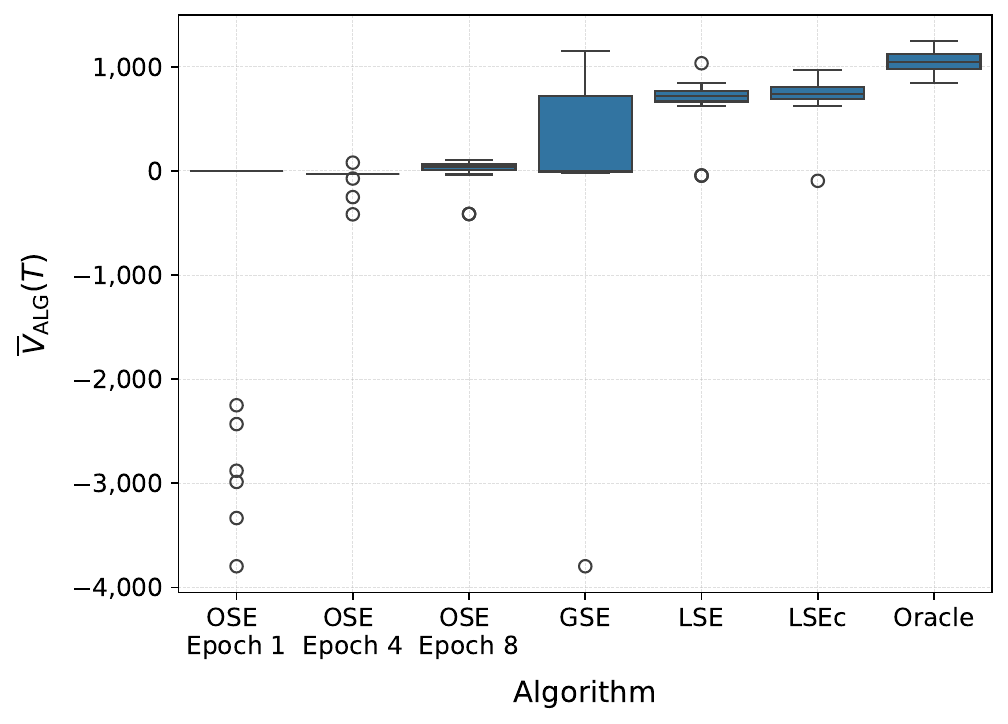}
    \caption{E.8: Timeless}
    \label{fig:E_8_LightGBM}
  \end{subfigure}
  \hfill 
  \begin{subfigure}[b]{0.24\textwidth}
  \hspace{-14pt}
    \includegraphics[width=1\textwidth]{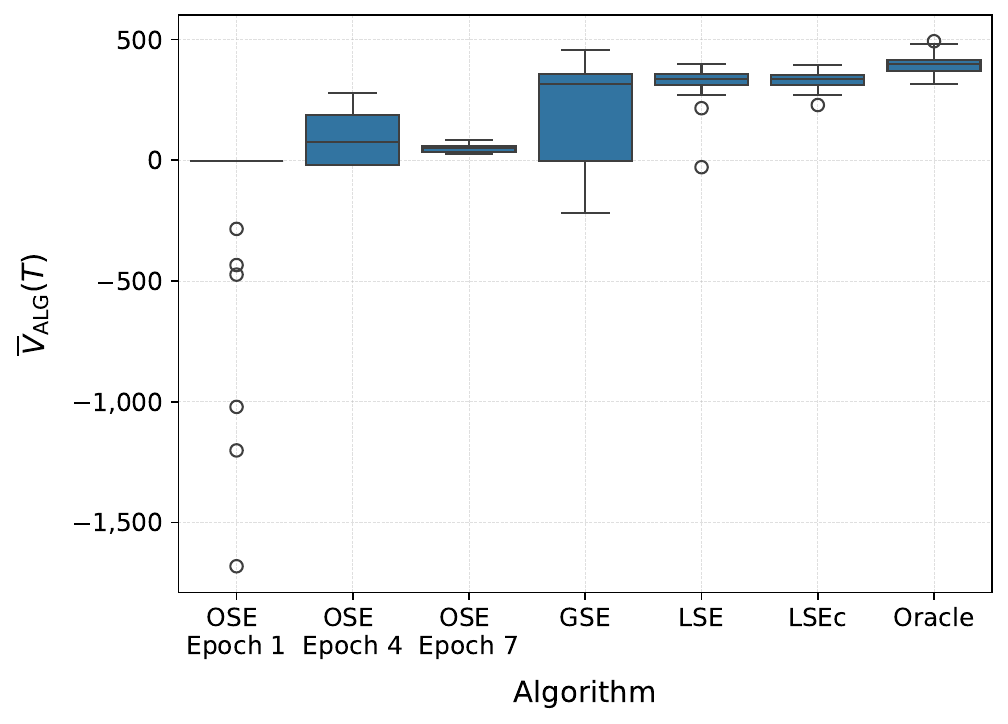}
    \caption{E.9: Small size}
    \label{fig:sub1}
  \end{subfigure}
    \caption{Robustness - LR (top line) - LightGBM (bottom line) }
  \label{fig:robustness_LightGBM}
   %
    
\end{figure}

\paragraph{Robustness} We finally assess the robustness of our findings by considering four additional experiments by varying batch sizes, discarding the time ordering, and the total sample size. Complete details about these experiments, denoted E.6, E.7, E.8 and E.9, are provided in Table \ref{exp_params}. Figure \ref{fig:robustness_LightGBM} display the performance of our algorithms across the 30 sample paths for these experiments. 

\begin{figure}[h!]
  \centering
  \begin{subfigure}[b]{0.18\textwidth}
    \includegraphics[width=1\textwidth]{Figures/LR_Utility_curve_Sequential_High_gap.pdf}
    \caption{E.1: GSE, LR}
  \end{subfigure} 
  \hfill
  \begin{subfigure}[b]{0.18\textwidth}
    \includegraphics[width=1\textwidth]{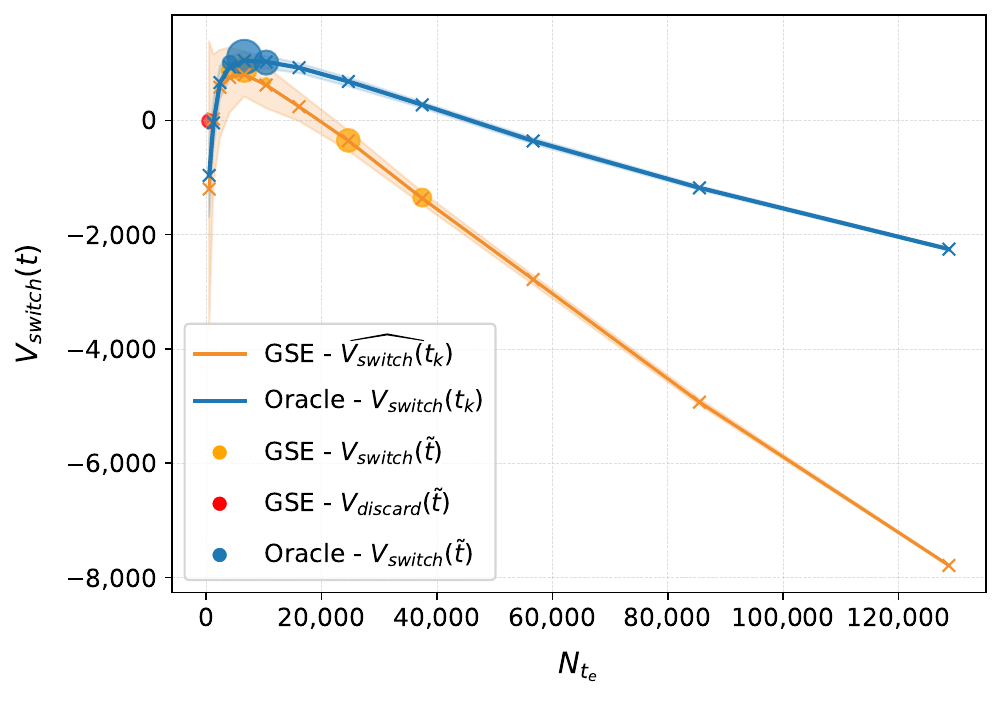}
    \caption{E.6: : GSE, LR}
  \end{subfigure}
  \hfill
  \begin{subfigure}[b]{0.18\textwidth}
    \includegraphics[width=1\textwidth]{Figures/LR_Utility_curve_Slope_based_High_gap.pdf}
    \caption{E.1: LSE, LR}
  \end{subfigure} 
  \hfill
  \begin{subfigure}[b]{0.18\textwidth}
    \includegraphics[width=1\textwidth]{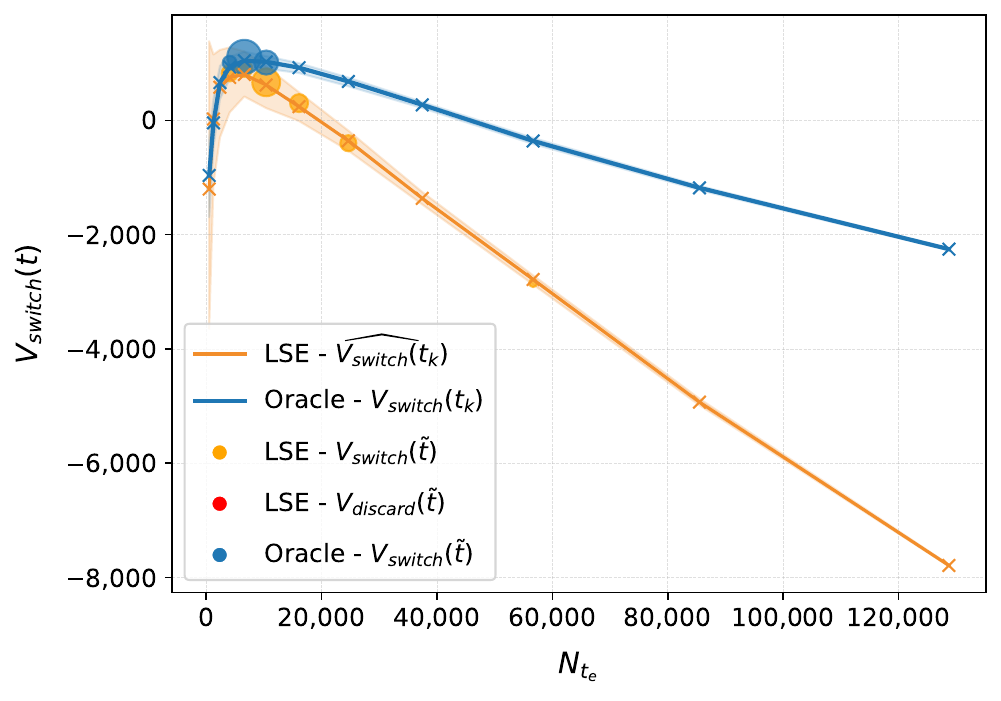}
    \caption{E.6: LSE, LR}
  \end{subfigure}
  \label{fig:robustness_E_6_utility}
  \hfill
  \begin{subfigure}[b]{0.18\textwidth}
    \includegraphics[width=1\textwidth]{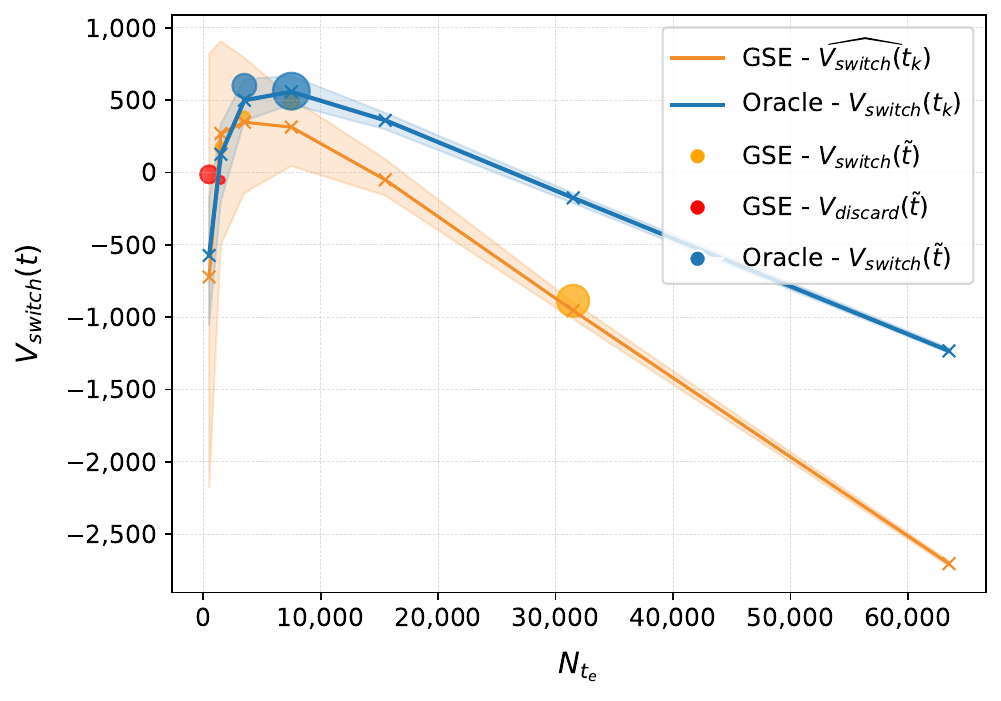}
    \caption{E.9: GSE, LR}
    \label{fig:E_9_LR_utility}
  \end{subfigure} 
  \caption{Robustness - Small batch. The plot conventions are the same as in Figure \ref{fig:E_1_OSE}.}
  \label{fig:EC_12}
\end{figure}

First, we reduce the batch-size growth across epochs from a factor of 2 to 1.5. As shown in Figures \ref{fig:E_6_LR} and \ref{fig:E_6_LightGBM}, this change has little effect on the performance of the oracle, OSE, and LSEc, with LSEc remaining close to oracle performance. In contrast, the impact on GSE and LSE is mixed: GSE performs worse with LR but better with LightGBM, while LSE improves with LightGBM and is largely unaffected with LR. These patterns reflect the impact of reduced batch-size growth, which increases decision frequency but also raises performance variability from repeated model training, leading to greater dispersion in stopping times (see Figure \ref{fig:EC_12}). In particular, for GSE, it leads to earlier and more dispersed stopping with LightGBM and later stopping with LR, improving and worsening performance, respectively. For LSE, smaller batches allow earlier decision-making after the optimal stopping time, improving performance under LightGBM. 

Second, we increase the batch-size growth across epochs from 2 to 2.5 and simultaneously raise the per-sample acquisition cost by a factor of 10. This experiment closely mirrors E.4, differing only in the larger batch-size growth, which allows us to isolate and assess the effect of increased batch size by comparing E.4 and E.7. As shown in Figure \ref{fig:E_7_LR} and \ref{fig:E_7_LightGBM}, this change has a small impact on the oracle. Although LSEc’s performance deteriorates, it remains the best-performing algorithm under LR, and its ranking is unchanged under LightGBM.

Third, we repeat experiments E.1 and E.2 after randomly shuffling the data to break the original time ordering, while still conducting the sequential evaluation based on this permuted sequence for 30 different sample paths. Figures \ref{fig:E_8_LR} and \ref{fig:E_8_LightGBM} show the results of experiment E.8 for the LR and the LightGBM, respectively. Overall, the oracle’s performance increases and the ranking of the algorithms remains largely unchanged, except that GSE drops to last place when excluding OSE. Importantly, LSEc performs near the oracle under both LR and LightGBM.

Finally, in the last experiment, denoted E.9, the total sample size available to the change is divided by two. As a consequence, the oracle’s performance diminishes, since a smaller total sample size limits the accumulation of per-sample gains from switching. This change has virtually no effect on the LightGBM results. By contrast, with the LR, the GSE algorithm’s performance deteriorates sharply and becomes largely unprofitable, whereas all other algorithms remain profitable, albeit with increased volatility (see Figure \ref{fig:E_9_LR}). This result highlights a key weakness of the GSE algorithm: when faced with substantial uncertainty about whether switching or discarding is optimal, the algorithm continues acquiring additional data until the relative cost of discarding exceeds that of switching, rather than committing earlier to a decision (see Fig. \ref{fig:E_9_LR_utility}). 

\end{document}